\documentclass{article}

\usepackage{microtype}
\usepackage{graphicx}
\usepackage{subcaption}
\usepackage{booktabs}
\usepackage{hyperref}

\usepackage[preprint]{icml2026}

\usepackage{amsmath}
\usepackage{amssymb}
\usepackage{mathtools}
\usepackage{amsthm}
\usepackage{bm}
\usepackage{enumitem}
\usepackage{xcolor}
\usepackage{multirow}

\theoremstyle{plain}
\newtheorem{theorem}{Theorem}
\newtheorem{proposition}{Proposition}
\newtheorem{lemma}{Lemma}
\newtheorem{corollary}{Corollary}
\theoremstyle{definition}
\newtheorem{definition}{Definition}
\newtheorem{example}{Example}
\theoremstyle{remark}
\newtheorem{remark}{Remark}

\newcommand{\R}{\mathbb{R}}
\newcommand{\E}{\mathbb{E}}
\newcommand{\Pp}{\mathbb{P}}
\newcommand{\one}{\mathbf{1}}
\newcommand{\Id}{\mathrm{I}}
\newcommand{\Phic}{\Phi^{\mathrm c}}
\newcommand{\cI}{\mathcal{I}}

\newcommand{\cS}{\mathcal{S}}
\newcommand{\cV}{\mathcal{V}}
\newcommand{\cL}{\mathcal{L}}
\newcommand{\cN}{\mathcal{N}}

\newcommand{\cU}{\mathcal{U}}

\newcommand{\ehat}{\hat e}
\newcommand{\chih}{\hat\chi}
\newcommand{\te}{\theta_{\mathrm e}}
\newcommand{\tho}{\theta_{\mathrm o}}
\newcommand{\Plin}{P_{\mathrm{lin}}}
\newcommand{\Ptwo}{P_{\mathrm{two}}}
\newcommand{\Pmc}{P_{\mathrm{mc}}}
\newcommand{\lr}{\mathrm{lr}}
\newcommand{\Sig}{\bar\sigma}
\DeclareMathOperator{\Cov}{Cov}
\DeclareMathOperator{\Var}{Var}

\DeclareMathOperator*{\argmax}{arg\,max}

\icmltitlerunning{The Race Is Decided Early! A Kernel Law Predicts Lock-In in Majority-Vote Test-Time RL}

\begin{document}

\twocolumn[
  \icmltitle{The Race Is Decided Early!\\ A Kernel Law Predicts Lock-In in Majority-Vote Test-Time RL}

  \icmlsetsymbol{equal}{*}
  \icmlsetsymbol{corres}{$\dagger$}
	
  \begin{icmlauthorlist}
  	\icmlauthor{Jaeho Hwang}{a}
  	\icmlauthor{Sehong Park}{a}
  	\icmlauthor{Daejin Choi}{b,corres}
  	\icmlauthor{Uiwon Hwang}{c,corres}
  \end{icmlauthorlist}
  
  \icmlaffiliation{a}{Human-Centered Artificial Intelligence Research Institute, Ewha Womans University}
  \icmlaffiliation{b}{Department of Artificial Intelligence, Ewha Womans University}
  \icmlaffiliation{c}{Department of Computer Science and Engineering, Ewha Womans University}
  
	\icmlcorrespondingauthor{Daejin Choi}{djchoi@ewha.ac.kr}
	\icmlcorrespondingauthor{Uiwon Hwang}{uiwon.hwang@ewha.ac.kr}
    
  \vskip 0.3in
]

\printAffiliationsAndNotice{\textsuperscript{$\dagger$}Co-corresponding authors }

\begin{abstract}
Test-time reinforcement learning with majority-vote rewards trains a language model on unlabeled questions by rewarding the samples that agree with its most frequent answer. During training, each question locks into one answer, on hard questions often a wrong one, but which answer it locks into is known only after training. In this paper, we show that this outcome is decided early, in a short race driven by sampling noise between the answers that are close at the start, and we derive the probability of each outcome. With a separate logit for each answer, the runner-up wins with a normal tail probability of the initial gap, scaled by an explicit noise constant. In a network, however, the answer logits depend on shared parameters, and through the kernel of these parameters the updates of the trailing answers also change the gap of the two leaders. This spillover shifts the boundary between the starts won by each leader, and a closed form computes the shifted probability from a kernel probe and the start logits. In LoRA experiments with three models from two families, predictions made before training with the kernel frozen at its start value forecast which answer each question locks into better than a logistic model refitted on the outcomes, although they use no fitted constants. The closed form matches these predictions on the runs in which a check before training finds that the two leaders stay above the other answers, and it explains why a law that ignores the spillover needs a strongly inflated step.
\end{abstract}

\section{Introduction}\label{sec:intro}

Test-time reinforcement learning with majority-vote rewards improves a language model on questions without labels \citep{zuo2025ttrl}. For each question the model samples a group of answers, takes the most frequent answer as a pseudo-label and rewards the samples that agree with it, which reinforces whatever the model already prefers. As training proceeds, the policy of each question concentrates on one answer, and on hard questions this answer is often a wrong one that training then cannot undo \citep{lin2026extinction}. Which answer a question locks into therefore decides what the method achieves on that question, but so far it is known only after training.

\begin{figure*}[t]
\centering
\includegraphics[width=\textwidth]{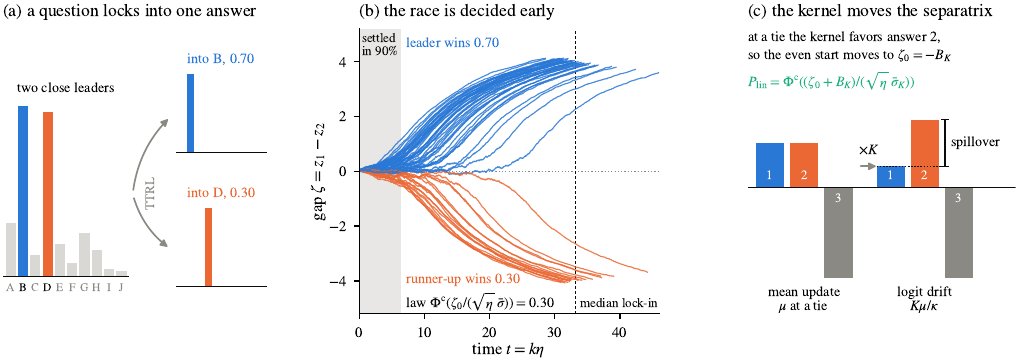}
\vspace{-14pt}
\caption{Overview. (a) An illustrative question with ten options and two close leaders. Majority-vote TTRL locks it into one answer, here B or D with the probabilities of the race in (b). (b) The gap between the two leaders in $60$ runs of the tabular chain from the start in (a), with $G=8$ and $\eta=0.02$, colored by the winner. The shading ends when $90$ percent of $4{,}000$ runs have seen their last change of the leader, long before the median lock-in. The runner-up wins in $0.30$ of the runs, as the lock-in law of Theorem~\ref{thm:lockin} predicts. (c) For the kernel of Figure~\ref{fig:race} at a tie with trailing mass $0.2$, the mean update raises both leaders equally, but the logit drift through the kernel raises answer 2 three times as much as answer 1. This spillover moves the separatrix away from the tie, and the closed form $\Plin$ computes the shift before training from a kernel probe and the start logits.}
\label{fig:overview}
\end{figure*}

Previous studies observe this lock-in and detect it during training from flip rates \citep{lin2026extinction}, explain through the gradients of a batch why majority-vote rewards track verifiable rewards \citep{wilson2026wrongvotes} or describe GRPO with noisy rewards as a deterministic flow \citep{rad2026rate}. However, none of them gives the probability with which each answer wins a question, and to our knowledge no analysis predicts it before training. This probability is determined by the start of training, because the outcome is decided early in a short race between two close answers, by the noise of a few sampled groups, and once one answer leads by a margin the reward widens the margin. Computing it is nevertheless difficult for two reasons. First, the race is a nonlinear stochastic process, whose law is not obvious even when every answer has its own logit. Second, in a network the answer logits depend on shared parameters, and the update of one answer also changes the logits of the others through the kernel of these parameters. In our experiments, a law that summarizes this kernel by a single scale, which we call the scalar law, mispredicts the outcomes so badly that its step has to be inflated by a factor of $12$ to $40$.

We study the race on a fixed set of answers that are single tokens, where the policy of a question is exactly the softmax of its answer logits. This paper answers three questions. Which answer does a question lock into, and what sets the probability of each outcome? How does the kernel of a network change this probability, and why does the scalar law fail? Can the probability be computed before training and predict the outcomes of actual training without fitted constants, also when questions share parameters? Figure~\ref{fig:overview} summarizes the answers.

\begin{figure*}[t]
\centering
\includegraphics[width=\textwidth]{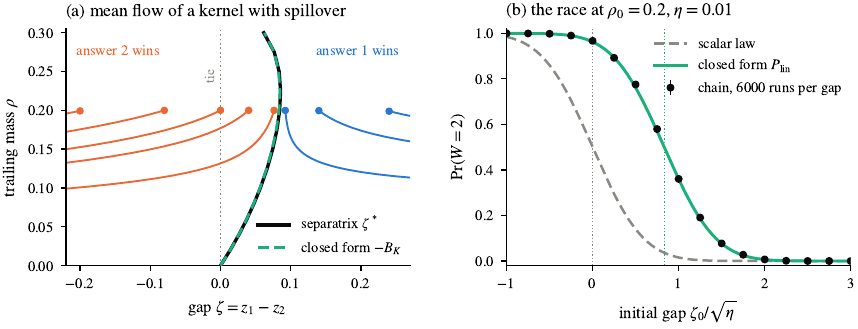}
\vspace{-14pt}
\caption{The race is decided at the separatrix. (a) The mean flow of a three-answer kernel with spillover ($G=8$, mean baseline) in the plane of the gap $\zeta$ and the trailing mass $\rho$. Orbits start at $\rho_0=0.2$ and are colored by their winner. Orbits that start between the tie and the separatrix are won by the runner-up although answer 1 leads. The zero of the closed form is indistinguishable from the separatrix $\zeta^*$ found by bisection on the flow. (b) The probability that the runner-up wins from $\rho_0=0.2$ at the step $\eta=0.01$, against the initial gap in units of $\sqrt\eta$. The dots are $6000$ runs of the frozen-kernel chain per gap with two standard errors. The closed form $\Plin$ (Definition~\ref{def:lin}) follows the chain, and the scalar law, which ignores the spillover, puts the even race at the tie.}
\label{fig:race}
\end{figure*}

\begin{enumerate}[leftmargin=*,itemsep=2pt,topsep=2pt]
\item \textbf{A lock-in law} (Section~\ref{sec:lockin}). Let $\eta$ be the effective step, the learning rate times the kernel scale along the gap. When two answers lead with a small initial gap $\beta\sqrt\eta$, the runner-up wins with probability $\Phic(\beta/\Sig)$, the chance that a normal noise with standard deviation $\Sig$ exceeds $\beta$, where $\Sig$ accumulates the noise of the votes along one deterministic path. The limit is uniform in the gap, and ties among several answers follow a Thurstone law, in which the answer with the largest start plus independent normal noise wins. Previous analyses describe a deterministic flow or detect lock-in during training, while this law gives the probability of each outcome from the start.
\item \textbf{Spillover and the separatrix law} (Section~\ref{sec:kernel}). Through the kernel, the updates of the trailing answers spill over into the gap of the two leaders. This spillover vanishes exactly when the gap direction is an eigenvector of the centered kernel, as for a tabular policy, and otherwise it moves the separatrix of the noise-free mean flow, the boundary between the starts won by each leader. The runner-up then wins with a normal tail probability of the distance to the moved separatrix (Figure~\ref{fig:race}). A closed form $\Plin$ computes this probability from a kernel probe and the start logits by accumulating the spillover along one deterministic path, and it places the moved separatrix with an error of second order in the mass of the trailing answers and of third order in the anisotropy of the kernel. The scalar law sees the kernel only through $\eta$ and omits this shift, which is why it needs an inflated step. Which answer wins thus depends on how strongly the kernel couples each leader to the trailing answers.
\item \textbf{Prediction before training} (Section~\ref{sec:exp}). We train LoRA on multiple-choice questions of MMLU-Pro with three models from two families. Before any run was trained, we computed and fixed the predictions of the frozen-kernel chain, a simulation of training with the kernel fixed at its start value. Without fitted constants, these predictions forecast which answer each question locks into better than a logistic model refitted on the outcomes, also when up to $64$ questions share the parameters. On the runs in scope, where a check before training finds that the two leaders stay on top, $\Plin$, whose code was fixed before any comparison, matches the chain without its Monte Carlo. Fitted to the chain, $\Plin$ needs almost no inflation of the step, while the scalar law needs a factor of about $15$.
\end{enumerate}


\section{Related Work}\label{sec:related}

\textbf{Majority-vote test-time RL.} TTRL trains on unlabeled questions with the majority answer of the model's own samples as the reward \citep{zuo2025ttrl}. \citet{lin2026extinction} observe an early window in which the correct answer survives only as a minority, after which a wrong majority is nearly irreversible, and they detect this window during training from flip rates. \citet{wilson2026wrongvotes} explain why majority-vote rewards track verifiable rewards through the gradients of a batch, in which inconsistent errors of different questions cancel. \citet{rad2026rate} describe GRPO with noisy rewards as a replicator flow in which one index separates learning from its failure, and \citet{corderoencinar2025certified} give statistical guarantees for majority voting and interpret TTRL as a tilt toward the mode. \citet{ding2026collective} show that questions sharing parameters can collapse together and rank runs by their risk of collapse from the initial kernel, but they do not give the probability with which each answer wins a question. 
Methods that steer training away from wrong majorities are discussed in Appendix~\ref{app:related}.

\textbf{Kernel views of fine-tuning.} Prompt-based fine-tuning of language models stays close to the kernel regime \citep{malladi2023kernel}, LoRA with a proximity penalty behaves like learning with a positive definite neural tangent kernel \citep{rahimiafzal2025linearization}, and the empirical kernel explains how RL post-training raises confidence and lowers diversity \citep{tomihari2026learning}. Unlike these works, we use the kernel as an input to a self-referential stochastic process whose distribution of outcomes we predict question by question, in which the anisotropy of the kernel, not only its scale, decides the outcome.

\textbf{Symmetry breaking and exit laws.} The probit form of a race that starts near a tie is classical and appears for explosive balls-in-bins processes \citep{mitzenmacher2004scaling}, for the exit of a small-noise diffusion near a hyperbolic equilibrium \citep{almadamonter2011normal} and, heuristically, for chiral symmetry breaking \citep{kondepudi1983chiral}. Lock-in by early fluctuations is a classical theme for urns and stochastic approximation \citep{arthur1989competing,pemantle2007survey}, and reinforced processes that interact through a network synchronize \citep{aletti2017synchronization}. The lock-in probability of \citet{borkar2002lockin} is the probability of convergence to an attractor from inside its domain, a different notion from ours. 
Beyond these classical results, our analysis gives the constants of majority votes for a discrete chain with trailing answers and shows how the spillover of a kernel moves the separatrix at an edge with a center direction.
\section{Majority-Vote Test-Time RL as a Race}\label{sec:setting}

\subsection{The Majority-Vote Update}\label{sec:update}

A question has $n$ answers with logits $z\in\R^n$ and policy $p=\mathrm{softmax}(z)$. One step samples $G\ge3$ answers from $p$ and counts them, $N=(N_1,\dots,N_n)$. Let $M=\max_aN_a$ and $f=M/G$, and let the leader $L$ be drawn uniformly from the answers with $N_a=M$. The majority reward is $1$ for the samples equal to $L$ and $0$ otherwise, and the policy gradient of the group with the baseline $f$ has the logit components
\begin{equation}\label{eq:update}
\xi_a=\omega(f)\,\frac{N_a}{G}\big(\one\{a=L\}-f\big),
\end{equation}
where $\omega$ is a positive function. The mean baseline has $\omega\equiv1$, and dividing the advantage by the standard deviation of the group rewards, as in GRPO \citep{shao2024deepseekmath}, gives $\omega(f)\propto(f(1-f))^{-1/2}$. The update sums to zero and is bounded. Its mean $\mu(p)=\E_p\xi$ and its covariance are polynomials in $p$ that commute with permutations of the answers. With tabular logits, one SGD step with the step $\eta$ is $z\leftarrow z+\eta\xi$.

Fix a threshold $1-\delta\in(\frac12,1)$. The winner $W$ is the first answer whose probability reaches $1-\delta$, and we say that the question locks into $W$. In our experiments, we set $1-\delta=0.95$, and in the sets of questions that we kept training for $300$ steps, none of the $1{,}792$ decided questions was ever overtaken afterwards.

\subsection{Logits That Move Through a Kernel}\label{sec:kernelchain}

In a network with parameters $\theta$, the answer logits $z(\theta)$ are the logits of the $n$ answer tokens at the answer position. An SGD step on the policy-gradient surrogate with learning rate $\lr$ moves $\theta$ by $\lr\,J^\top\xi$ with $J=\partial z/\partial\theta$, and hence $z$ by $\lr\,K\xi+O(\lr^2)$ with the kernel $K=JJ^\top$ \citep{jacot2018ntk,malladi2023kernel}. Freezing the kernel at its value before training gives the frozen-kernel chain at step $k$:
\begin{equation}\label{eq:chain}
z_{k+1}=z_k+\lr\,K\xi_k .
\end{equation}
Since $\xi$ sums to zero and the policy ignores a common shift of the logits, we may replace $K$ by $HKH$ with the centering matrix $H=\Id-\one\one^\top/n$. A fixed preconditioner $D$, such as the preconditioner of AdamW computed from the expected squared gradient at the start, gives $K=JDJ^\top$ and is absorbed in the kernel. Let $1$ and $2$ be the two leading answers, $e=e_1-e_2$, $\zeta=z_1-z_2$ their gap and
\begin{equation}\label{eq:kappa}
\kappa=\tfrac12e^\top Ke,\qquad\eta=\lr\,\kappa .
\end{equation}
The number $\eta$ is the effective step of the gap. When $K=\kappa H$, the chain \eqref{eq:chain} is the tabular chain with the step $\eta$. We write $t=k\eta$ for the time and $z'=K\mu(z)/\kappa$ for the mean flow of \eqref{eq:chain}.

\subsection{Winners, Ties and Windows}\label{sec:window}

We start near a tie. Let $z^0\in\mathbb{R}^{n}$ be a tied reference logit vector with $z^0_1=z^0_2$ and $p_1(z^0)>\max_{a\ge3}p_a(z^0)$, and start the chain from $z^0+\frac12\zeta_0e$. The tied path $\bar z$ of $K$ solves $\bar z'=(\Id-\frac12ee^\top)K\mu(\bar z)/\kappa$ from $z^0$, so it keeps the two leaders tied. Along a tied path, let $d=\xi_1-\xi_2$, let $r$ be the derivative of $\E d$ in $\zeta$ at a fixed sum $z_1+z_2$ and fixed trailing logits, and let $v=\Var(d)$. We put
\begin{equation}\label{eq:sigma}
\cI(t)=\int_0^tr(\tau)\,d\tau,\qquad \Sig^2=\int_0^\infty v(t)\,e^{-2\cI(t)}\,dt .
\end{equation}
As long as the two leaders stay above every other answer along the tied path, the rate $r$ is bounded below by a positive constant (Lemma~\ref{lem:tied}). This always holds for the tabular kernel, and for a general kernel it is the scope condition of Section~\ref{sec:scope}. Thus, a gap of order $\sqrt\eta$ grows like $e^{\cI(t)}$, while every step adds noise of variance of order $\eta$ per unit of time. The outcome is settled in a window of time of order one, long before the decision at a time of order $\log(1/\eta)$, and the natural scale of the initial gap is $\sqrt\eta$. We write $\zeta_0=\beta\sqrt\eta$.

\section{The Lock-In Law}\label{sec:lockin}

\subsection{Two Close Leaders}\label{sec:two}

We first take a separate logit for each answer, $K=\kappa H$, so that the chain is $z_{k+1}=z_k+\eta\xi_k$.

\begin{theorem}[Lock-in law]\label{thm:lockin}
Let $\omega$ be positive, $G\ge3$, and $n\ge2$. Let $z^0$ be tied with $p_1(z^0)>\max_{a\ge3}p_a(z^0)$, and start the tabular chain from $z^0+\frac12\beta\sqrt\eta\,e$. Then $0<\Sig<\infty$, and
\begin{equation}\label{eq:lockin}
\sup_{\beta\in\R}\big|\Pp(W=2)-\Phic(\beta/\Sig)\big|\to0,\,\,\,\,\,\Pp(W\ge3)\to0,
\end{equation}
as $\eta\to0$. For $n=2$ the error is $\eta\,\varphi_1(\beta)+O(\eta^2)$ uniformly in $\beta$, with an explicit function $\varphi_1$.
\end{theorem}

Here $\Phic(x)=\Pp(Y>x)$ for a standard normal $Y$, so the runner-up wins with the probability that $\Sig Y$ exceeds the initial gap in units of $\sqrt\eta$. The constant $\Sig$ needs one integration of the tied path with exact multinomial moments of the votes. The trailing answers never win in the limit, because the most likely answer gains log-odds on every other answer in expectation (Lemma~\ref{lem:leader}). Simulations with eight answers show an error of order $\eta$ as well (Appendix~\ref{app:numerics}). The proofs are in Appendix~\ref{app:lockin}.

\subsection{Ties Among Several Answers}\label{sec:thurstone}

\begin{corollary}[Thurstone law]\label{cor:thurstone}
Let $m\ge2$ leaders be tied at $z^0$ above all other answers, and start the tabular chain from $z^0+\sqrt\eta\,\beta$ with $\beta\in\R^n$. Let $\Sig$ be computed as in \eqref{eq:sigma} along the tied path of the $m$ leaders, and let $Y_1,\dots,Y_m$ be independent standard normal variables. Then, for $i\le m$,
\begin{equation}\label{eq:thurstone}
\begin{aligned}
\Pp(W=i)\to\Pp\Big(&\beta_i+\tfrac{\Sig}{\sqrt2}Y_i>\beta_j+\tfrac{\Sig}{\sqrt2}Y_j\\
&\text{for all } j\le m \text{ with } j\ne i\Big),
\end{aligned}
\end{equation}
and $\Pp(W>m)\to0$.
\end{corollary}

The winner is a probit choice among the tied answers in the sense of \citet{thurstone1927law}. With Gumbel noise the winner would be drawn from the start policy, but the noise of the race is normal. The probit form itself is classical \citep{mitzenmacher2004scaling,almadamonter2011normal}, and Theorem~\ref{thm:lockin} gives its constants for majority votes, holds uniformly in the gap for the discrete chain and shows that the trailing answers, which enter only through the tied path, never win. 

\section{The Kernel Moves the Separatrix}\label{sec:kernel}

\subsection{What Moves the Gap}\label{sec:gapstep}

For a general kernel put $\ehat=Ke/\kappa$, $\tilde e_a=e_a-\frac12(e_1+e_2)$ and
\begin{equation}\label{eq:chi}
\chi_a=\ehat^{\,\top}\tilde e_a\qquad(a\ge3).
\end{equation}
We call $\chi$ the spillover weights. Write $\xi_\perp=(\xi_a)_{a\ge3}$ and $\mu_\perp=(\mu_a)_{a\ge3}$.

\begin{lemma}[The gap step]\label{lem:gapstep}
One step of \eqref{eq:chain} changes the gap by
\begin{equation}\label{eq:gapstep}
\zeta_{k+1}-\zeta_k=\eta\Big(d_k+\sum_{a\ge3}\chi_a\xi_{k,a}\Big),
\end{equation}
and, for a centered kernel, $\chi=0$ if and only if $Ke=\kappa e$. At a tied point $\E d=0$, the spillover $\chi\cdot\xi_\perp$ has the mean $\psi_K=\chi\cdot\mu_\perp$, the gap has the rate $r$, and the variance of the gap step divided by $\eta^2$ is $v_K=v+\chi^\top\Cov(\xi_\perp)\chi$.
\end{lemma}

The scalar law keeps only $\eta d_k$, which is exact if and only if $e$ is an eigenvector of the centered kernel. This holds for the tabular kernel but not in a network. At a tie the drift of $d$ vanishes, while the spillover has a mean of order one per unit of time. Near the edge every trailing answer loses votes and has $\mu_a<0$, and the leader whose kernel couples more strongly to the trailing answers is dragged down with them. Since the noise of the gap is only of order $\sqrt\eta$ per unit of time, a spillover of order one decides the race unless the start sits at a compensating distance from the tie.

\subsection{A Weak Anisotropy Shifts the Probit}\label{sec:weak}

When the anisotropy is of the order of the noise, the spillover acts as a forcing of the race.

\begin{theorem}[A weak anisotropy]\label{thm:weak}
Let $A$ be symmetric with $A\one=0$ and put $\chih_a=\tilde e_a^\top Ae$. Run $z_{k+1}=z_k+\eta(H+\sqrt\eta A)\xi_k$ from $z^0+\frac12\beta\sqrt\eta\,e$ with $z^0$ as in Theorem~\ref{thm:lockin}. Let $\bar z$ be the tabular tied path and
\begin{equation}\label{eq:betaA}
\beta_A=\int_0^\infty e^{-\cI(t)}\,\chih\cdot\mu_\perp(\bar z(t))\,dt .
\end{equation}
Then $\Pp(W=2)\to\Phic((\beta+\beta_A)/\Sig)$ and $\Pp(W\ge3)\to0$ as $\eta\to0$.
\end{theorem}

The spillover shifts the probit by $\beta_A/\Sig$. For $m$ tied leaders, the Thurstone law of Corollary~\ref{cor:thurstone} is shifted by a vector (Theorem~\ref{thm:block}). With an anisotropy of order one, the shift would be of order $1/\sqrt\eta$ in units of the noise, so a fixed kernel decides the race deterministically except near one special start.

\subsection{The Race Is Decided at the Separatrix}\label{sec:separatrix}

For a fixed kernel and a start near the edge or under a weak anisotropy, the mean flow from $z^0+\frac12\zeta_0e$ decides for answer 1 if $\zeta_0$ exceeds a critical gap $\zeta^*$ and for answer 2 if $\zeta_0<\zeta^*$ (Appendix~\ref{app:kernel}). The point $z^*=z^0+\frac12\zeta^*e$ lies on the separatrix between the two basins, and its orbit converges to the edge $p^\circ=(\frac12,\frac12,0,\dots,0)$, where the two leaders share all the mass. The edge is not hyperbolic, since it has one unstable direction, the gap, and a center manifold on which the trailing mass decays only like $1/t$.

\begin{theorem}[The separatrix law]\label{thm:separatrix}
Let $K$ be positive semidefinite with $K\one=0$ and $\kappa>0$. Let the mean orbit from a finite $z^*$ converge to $p^\circ$ while every probability stays below $1-\delta$. Then the chain \eqref{eq:chain} from $z^*+\frac12\beta\sqrt\eta\,e$ satisfies
\begin{equation}\label{eq:seplaw}
\Pp(W=2)\to\Phic\Big(\frac{\beta\,\ell(0)e}{2\sigma^*}\Big),\qquad\Pp(W\ge3)\to0,
\end{equation}
where $\ell$ solves the adjoint equation $\ell'=-\ell K\partial_z\mu/\kappa$ along the orbit of $z^*$ and, near the edge, is a positive multiple of the gradient of a normal coordinate of the separatrix discounted by its growth (Appendix~\ref{app:seplaw}), and $(\sigma^*)^2=\int_0^\infty\ell K\Cov(\xi)K\ell^\top dt/\kappa^2$ is the noise that it collects. The law does not change when $\ell$ is multiplied by a positive number.
\end{theorem}

The outcome follows a probit law in the distance to the separatrix in units of $\sqrt\eta$, measured by the adjoint. Near the edge, every line of starts that differ only in the gap contains exactly one such point $z^*$ near the tie if the Schur matrix $S=Q-\frac12\chi\chi^\top$ with $Q_{ab}=\tilde e_a^\top K\tilde e_b/\kappa$ is strictly copositive, for instance if $K$ is positive definite on $\one^\perp$ (Proposition~\ref{prop:critical}). Far from the edge such a point need not exist, and a trailing answer can then win (Examples~\ref{ex:nocritical} and~\ref{ex:overtake}). The scope check of Section~\ref{sec:scope} detects these starts.

\subsection{A Closed Form From a Kernel Probe}\label{sec:closed}

\begin{definition}[The closed form]\label{def:lin}
Let $z^0$ be as in Theorem~\ref{thm:lockin}, and suppose that the two leaders stay above every other answer along the tied path $\bar z$ of $K$ from $z^0$. Along $\bar z$, let $\psi_K=\chi\cdot\mu_\perp(\bar z)$, let $r$ and $v_K$ be as in Lemma~\ref{lem:gapstep}, and let $\cI(t)=\int_0^tr$. Put
\begin{equation}\label{eq:BK}
B_K=\int_0^\infty\psi_K\,e^{-\cI}\,dt,\qquad \Sig_K^2=\int_0^\infty v_K\,e^{-2\cI}\,dt,
\end{equation}
\begin{equation}\label{eq:plin}
\Plin=\Phic\Big(\frac{\zeta_0+B_K}{\sqrt\eta\,\Sig_K}\Big).
\end{equation}
\end{definition}

$\Plin$ uses only the start logits, the kernel, the step and $G$. It needs one integration of the tied path with vote moments that are exact for $G\le8$, about a quarter of a second per run at $G=8$ with ten answers, and no Monte Carlo of the chain. Its zero $-B_K$ approximates the separatrix with the errors of the next proposition, and near the edge its scale $\Sig_K$ agrees with the scale of Theorem~\ref{thm:separatrix} up to an error of the order of the trailing mass (Corollary~\ref{cor:scale}).

\begin{proposition}[The zero of the closed form]\label{prop:zero}
Let $z^0$ be as in Theorem~\ref{thm:lockin} and let $\zeta^*$ be the critical gap of the mean flow of $K$ from $z^0$.
\begin{enumerate}[label=(\roman*),leftmargin=*,itemsep=1pt,topsep=2pt]
\item If $S$ is strictly copositive and $\rho_0=\sum_{a\ge3}p_a(z^0)$ is small, then $\zeta^*$ is unique near $0$ and
\begin{equation}\label{eq:loop}
\zeta^*=-B_K\Big(1+\frac{\gamma^\circ}{2r^\circ}\sum_{a\ge3}\chi_a^2\,p_a(z^0)\Big)+O(\rho_0^3),
\end{equation}
where $r^\circ$ is the rate at $p^\circ$ and $\gamma^\circ>0$ is the rate at which a rare answer loses mass there.
\item For $K_s=\kappa(H+sA)$ with $A$ as in Theorem~\ref{thm:weak}, one has $\zeta^*(s)=-B_{K_s}+O(s^3)$.
\item With $s=\sqrt\eta$, $B_{K_s}=\sqrt\eta\,\beta_A+O(\eta)$ and $\Sig_{K_s}=\Sig+O(\sqrt\eta)$, so $\Plin$ tends to the limit of Theorem~\ref{thm:weak}.
\end{enumerate}
\end{proposition}

Near the edge, the only second-order error of $-B_K$ comes from a feedback loop, in which a displaced gap moves the trailing answers through the kernel, and the moved answers change the spillover. Under a weak anisotropy, the swap of the two leaders splits $A$ into an even part, which only moves the tied path, and an odd part, which carries the whole spillover. The separatrix is odd in the odd part, so the error of $-B_K$ has no term of second order. Because this error does not shrink with the step for a fixed kernel, $\Plin$ is accurate as long as it is small against the noise scale $\sqrt\eta\,\Sig_K$. Appendix~\ref{app:kernel} gives the proofs.

\subsection{Why a Scalar Law Fails}\label{sec:scalar}

The scalar law $\Ptwo=\Phic(\zeta_0/(\sqrt\eta\,\Sig))$ uses $\kappa$ and the tabular tied path.

\begin{proposition}[A probit with an omitted shift]\label{prop:omitted}
Let the outcome of a run follow $\Pp(W=2\mid\zeta_0,B)=\Phic((\zeta_0+B)/\sigma)$ with $\sigma=\sqrt\eta\,\Sig$, and let $B$ be normal with mean $\mu_B$ and variance $\sigma_B^2$, independent of $\zeta_0$. Then
\begin{equation}\label{eq:omitted}
\Pp(W=2\mid\zeta_0)=\Phic\Big(\frac{\zeta_0+\mu_B}{\sqrt{\sigma^2+\sigma_B^2}}\Big),
\end{equation}
so $\Ptwo$ fits the outcomes only with the step multiplied by $\lambda=1+\sigma_B^2/(\eta\Sig^2)$ and with an intercept.
\end{proposition}

At a fixed kernel $B_K$ does not shrink with the step, so $\lambda-1$ grows like $1/\eta$ and the scalar law fails more at small steps. Section~\ref{sec:lambda} shows that the multipliers computed from the kernels alone reproduce those that the outcomes imply.

\section{Predicting Lock-In Before Training}\label{sec:exp}

\subsection{Questions, Models and Kernel Probes}\label{sec:testbed}

For each model we train on $500$ questions of the MMLU-Pro test set \citep{wang2024mmlupro} with ten options each, in a chat template that ends before the answer letter. The policy is the softmax of the ten letter logits. The questions were chosen before training among those with at most $400$ tokens, letter mass at least $0.8$ and a largest probability below $0.95$, stratified by the gap of the two leaders and by the largest probability. For Qwen2.5-1.5B, the leader is wrong and the runner-up is right in $113$ of them. The models are Qwen2.5-1.5B-Instruct, Llama-3.2-1B-Instruct and Qwen2.5-7B-Instruct \citep{qwen25report,grattafiori2024llama3} with LoRA \citep{hu2022lora} of rank $32$ on all attention and MLP projections, initialized with $B=0$, in float32. A run trains one question with the majority reward, one on-policy step per group and no clipping, and it stops at the first step at which a letter reaches probability $0.95$. Before training we computed the exact kernel $K=JJ^\top$ of the ten letter logits with respect to the LoRA parameters for every question and seed.

For Qwen2.5-1.5B we ran a reference setting with $G=8$ and the learning rate that puts the median of $\eta=\lr\,\kappa$ at $0.1$, and five unseen settings with a third or three times this rate or with $G\in\{4,16,32\}$, three seeds each ($4{,}765$ runs). Llama-3.2-1B repeats the reference setting and two unseen settings ($2{,}631$ runs), and Qwen2.5-7B has one setting ($1{,}000$ runs). We also trained Qwen2.5-1.5B with AdamW at $\beta_1\in\{0,0.9\}$ and with SGD and the standard-deviation baseline ($4{,}500$ runs), where for AdamW the kernel includes the second moment frozen at the start. Finally, $448$ sets of $2$ to $64$ questions were trained with shared parameters. Appendix~\ref{app:exp} gives the details.

\begin{figure*}[t]
\centering
\includegraphics[width=\textwidth]{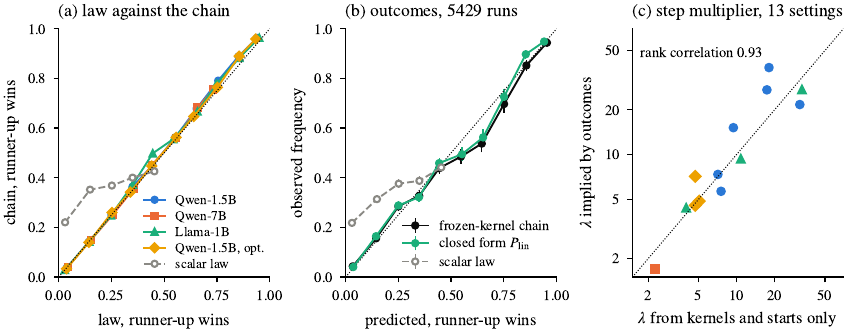}
\vspace{-14pt}
\caption{Prediction before training on the evaluated runs in scope. (a) The probability that the runner-up wins under the closed form and under the scalar law against the chain, in bins of the law. The scalar law can never favor the runner-up, while the spillover often does. (b) Observed frequency with which the runner-up wins, among runs won by one of the two leaders and pooled over the four groups of Table~\ref{tab:match}, in bins of each prediction, with one standard error. (c) The step multiplier of the scalar law that the outcomes imply, against the one that $\Plin$ implies from the kernels and the start logits alone, for the thirteen settings with colors as in (a).}
\label{fig:exp}
\end{figure*}

\subsection{The Frozen-Kernel Chain Predicts Lock-In}\label{sec:registered}

Before any run was trained, we computed for every run the frozen-kernel chain $\Pmc$, a Monte Carlo of \eqref{eq:chain} with the probed kernel and $2{,}000$ samples, and kept these predictions fixed in all comparisons below (Appendix~\ref{app:registration}).
Table~\ref{tab:registered} compares $\Pmc$ with a logistic regression on the gap, the largest probability and the entropy of the start, fitted to the outcomes of the same runs by cross-fitting. Without any constant fitted to the outcomes, the chain is better by $0.057$ to $0.084$ in Brier score in the five unseen settings, in a second model family, at a larger size, under AdamW and with the standard-deviation baseline. Its probabilities rank the outcomes well, but their calibration slope is $0.84$ $[0.75,0.94]$ in the unseen settings of Qwen-1.5B, $0.80$ for Llama-1B and $1.56$ for Qwen-7B, and we therefore do not claim calibration. The scalar law $\Ptwo$, which we had also fixed before training, fails with a calibration slope of $0.08$ and needs step multipliers of $12$ to $40$. Section~\ref{sec:kernel} traces this failure to the spillover, and the rest of this section tests that explanation.

\begin{table}[t]
\caption{Prediction of lock-in by the frozen-kernel chain, computed before training. Brier score for the event that the leader wins and its difference to a logistic model refitted on the outcomes, on the evaluated runs, those with $\eta\in[0.01,0.3]$ whose predicted probability that the leader loses lies in $[0.01,0.99]$. The last two rows compare the coupled chain with the isolated chain on members of shared sets. Intervals are $95\%$ bootstrap intervals over questions or sets. Every difference passes a one-sided noninferiority test with a margin of $0.003$ fixed before training, with $p<10^{-6}$ after Holm correction.}
\label{tab:registered}
\centering\small\setlength{\tabcolsep}{3.2pt}
\begin{tabular}{@{}lrrr@{}}
\toprule
model and setting & runs & Brier & difference\\
\midrule
Qwen-1.5B, unseen & 2{,}002 & 0.149 & $-0.075$\\
Llama-1B, unseen & 917 & 0.143 & $-0.084$\\
Qwen-7B & 418 & 0.117 & $-0.066$\\
AdamW, $\beta_1=0.9$ & 600 & 0.159 & $-0.057$\\
SGD, std baseline & 602 & 0.129 & $-0.075$\\
\midrule
shared, 16 and 64, SGD & 1{,}392 & 0.083 & $-0.086$\\
shared, 64, AdamW & 896 & 0.086 & $-0.104$\\
\bottomrule
\end{tabular}

\footnotesize Intervals of the differences exclude zero, for example $[-0.087,-0.048]$ for Qwen-7B and $[-0.117,-0.053]$ for the shared SGD sets.
\end{table}


\subsection{The Closed Form Matches the Chain}\label{sec:match}

We computed $\Plin$ for every run from the probed kernel and the start logits, with its code fixed before we compared it with the chain or the outcomes, although its predictions, unlike those of the chain, were not computed before training. Figure~\ref{fig:exp}a and Table~\ref{tab:match} compare it with the chain on the runs in scope (Section~\ref{sec:scope}).

\begin{table}[t]
\caption{The closed form against the chain and the outcomes, on the evaluated runs in scope. Probit slope of the chain on $\Plin$ for the probability that the runner-up wins, on all runs in scope with the chain in $(0.02,0.98)$. Brier scores for the event that the leader wins. The differences between $\Plin$ and the chain have $95\%$ intervals that contain zero. The row ``opt.'' pools the three optimizer settings.}
\label{tab:match}
\centering\small\setlength{\tabcolsep}{4pt}
\begin{tabular}{@{}lrrrrr@{}}
\toprule
& & probit & \multicolumn{3}{c}{Brier score}\\
\cmidrule(l){4-6}
group & runs & slope & chain & $\Plin$ & scalar\\
\midrule
Qwen-1.5B & 2{,}256 & 1.07 & .1443 & .1462 & .2434\\
Qwen-7B & 416 & 1.03 & .1160 & .1177 & .1723\\
Llama-1B & 1{,}164 & 1.04 & .1388 & .1408 & .2687\\
Qwen-1.5B, opt. & 1{,}674 & 1.06 & .1337 & .1357 & .2181\\
\bottomrule
\end{tabular}
\end{table}

On the runs in scope, the probit of the chain's probability that the runner-up wins regresses on that of $\Plin$ with slopes of $1.03$ to $1.07$, and once the wins of a third letter are set aside, the chain implies a step multiplier of $0.90$ to $1.03$ for $\Plin$ in all thirteen settings. On the outcomes, the Brier score of $\Plin$ is about $0.002$ above that of the chain in every group, with an interval that contains zero, while the scalar law is worse by $0.055$ to $0.128$ (Figure~\ref{fig:exp}b). The closed form thus keeps the predictive power of the chain without its Monte Carlo and without a fitted constant. In the reference setting of Qwen-1.5B, the chain implies on the two-letter runs a step multiplier of $15.2$ for the scalar law and $1.11$ for $\Plin$, and the outcomes imply $18.5$ and $1.56$.

The remaining difference between $\Plin$ and the chain is consistent with the cubic error of Proposition~\ref{prop:zero}, and moving the zero of $\Plin$ by this cubic term with the coefficient of synthetic kernels, without a fit, lowers the spread of the probit difference by $26$ to $39$ percent, except for Llama-1B (Appendix~\ref{app:cubic}).

\subsection{Races Outside the Two-Leader Law}\label{sec:scope}

A run is in scope when the tied path of its kernel keeps the two leaders above every other answer, which can be checked from the kernel and the start alone. Example~\ref{ex:overtake} shows why the check is needed. Along the tied path of this positive semidefinite kernel without spillover, a trailing answer overtakes the two leaders, no start near the tie lies on a separatrix of the two leaders, and the trailing answer wins $90$ percent of the runs of the chain from the tie. The check flags $0.5$ to $14$ percent of the evaluated runs, and these runs carry $92$ to $94$ percent of the excess Brier score of $\Plin$ over the chain, except for Qwen-7B, where only two runs are flagged. In the flagged runs with three or more close letters a third letter wins $31$ to $61$ percent of the time, close to the prediction of the chain, a frequency that no law of two leaders can produce. We therefore use $\Plin$ for the runs in scope and the chain for the flagged runs.

\subsection{The Step Multiplier Explained}\label{sec:lambda}

For each of the thirteen settings, we computed from $\Plin$ alone, without the chain or the outcomes, the step multiplier that it implies for $\Ptwo$ on the runs in scope with two close letters. It accounts for $57$ to $97$ percent of the excess $\lambda-1$ that the chain implies, and its rank correlation with the multiplier that the outcomes imply is $0.93$ across the settings, the same as for the chain itself (Figure~\ref{fig:exp}c). It grows as the learning rate shrinks and as $G$ grows, in the same order as the multiplier that the outcomes imply (Table~\ref{tab:lambda} in Appendix~\ref{app:lambda}). Restricting the evaluation to predicted probabilities between $0.01$ and $0.99$ selects runs near the separatrix, which correlates $\zeta_0$ with $B_K$ and raises this multiplier by up to a factor of $3.1$ against all runs in scope, an effect that Proposition~\ref{prop:omitted} leaves out. For Qwen-7B the median shift $|B_K|/(\sqrt\eta\,\Sig_K)$ is $0.22$, against $0.73$ for Qwen-1.5B and $1.12$ for Llama-1B, which is why the outcomes implied a multiplier of only $2.25$ for its scalar law on all its evaluated runs, against $23$ for Qwen-1.5B and $19$ for Llama-1B.

\subsection{Questions That Share Parameters}\label{sec:shared}

Real test-time training updates many questions with shared parameters. Let the questions $s\in\cS$ have kernel blocks $K_{ss'}=K_{s's}^\top$ and update all of them at every step, $z_{s,k+1}=z_{s,k}+\lr\sum_{s'}K_{ss'}\xi_{s',k}$, with independent votes given the state.

\begin{lemma}[The gap step with shared parameters]\label{lem:sharedstep}
Let $\kappa_s$, $\eta_s$ and $\chi_s$ be the quantities of \eqref{eq:kappa} and Lemma~\ref{lem:gapstep} for the block $K_{ss}$. One step changes the gap of question $s$ by
\begin{equation}\label{eq:sharedstep}
\eta_s\Big(d_{s,k}+\chi_s\cdot\xi_{s,k,\perp}+\sum_{s'\ne s}c_{ss'}^\top\xi_{s',k}\Big),\qquad c_{ss'}=\frac{K_{s's}e_s}{\kappa_s}.
\end{equation}
\end{lemma}

The mean of the cross term is a second spillover. If the anisotropy of every block and the cross blocks are of order $\sqrt\eta$, the questions lock in independently in the limit, and each follows Theorem~\ref{thm:weak} with a shift that collects its own spillover and the mean updates of the other questions along their tied paths (Theorem~\ref{thm:coupled}).

The coupled chain, a Monte Carlo of the shared frozen-kernel chain that we also computed before training, predicts the winner of each member better than the isolated chain (last two rows of Table~\ref{tab:registered}). In sets of $64$ the leader wins in $0.68$ of the members, against $0.67$ predicted and $0.83$ without sharing, and the shifts of the letter frequencies correlate at $0.97$ with the prediction. The closed form with the shared shift follows the change that sharing causes in the probability that the leader loses, with a Pearson correlation of $0.87$ and a slope of $0.96$ at size $64$, and it closes $89$ to $100$ percent of the Brier gap between the isolated law and the coupled chain at sizes $16$ and $64$. It is more confident than the coupled chain, with probit slopes that fall from $0.67$ at size $2$ to $0.33$ at size $64$, while the isolated law of the same members follows the isolated chain with slopes of $1.05$ to $1.06$. The residual is a pure scale and grows with the cross forcing, as Proposition~\ref{prop:omitted} predicts for an omitted random shift. Members that decide early keep pushing the others toward their own random winners, and the closed form leaves this push out.

\section{Limitations and Conclusion}\label{sec:conclusion}

\textbf{Limitations.} We study answers that are single tokens from a fixed set of ten letters, the adapters stay close to their linearization until the race is settled, and the objective has no clipping, no KL term and no exploration term. The probabilities rank and discriminate well, but their calibration depends on the model and the setting. Decisions come earlier than the frozen kernel predicts because the kernel grows late in training, and momentum lies outside the model. Races in which a third answer joins the two leaders are flagged in advance and need the chain. When questions share parameters, the closed form captures the mean shift but is more confident than the coupled chain, because the gaps of different questions are coupled and their winners become correlated at practical step sizes (Appendix~\ref{app:slow}). The error of order $\eta$ is proved for two answers and only observed in simulations with more answers. Free-form answers, long reasoning, clipped objectives and larger models are left open.

\textbf{Conclusion.} Majority-vote test-time RL decides each question in a short stochastic race between the answers that are close at the start, and this race is decided at the separatrix of the mean flow. In a network, the answer-logit kernel moves the separatrix through the updates of the other answers, a spillover that the scalar law omits. A closed form computed before training from a kernel probe and the start logits places the moved separatrix, matches the frozen-kernel chain on the runs in scope and explains why the scalar law needs a large step multiplier. The chain itself predicts the outcomes of actual training without fitted constants, also when questions share parameters, and whether a question lies in the scope of the two-leader law can be checked before training. Because both need only a kernel probe at the start, they can flag before training the questions whose outcome is still open, which could guide where to spend exploration or labels.

\newpage
\section*{Impact Statement}
This paper studies how majority-vote test-time reinforcement learning decides which answer a model adopts for each question. The results can estimate before training how likely each question is to settle on an answer other than its current leader, which makes label-free training more predictable and easier to audit. We do not foresee negative societal consequences that are specific to this work beyond those of reinforcement learning for language models in general.

\bibliography{refs}
\bibliographystyle{icml2026}

\newpage
\appendix
\onecolumn
\numberwithin{equation}{section}
\numberwithin{theorem}{section}
\numberwithin{proposition}{section}
\numberwithin{lemma}{section}
\numberwithin{corollary}{section}
\numberwithin{definition}{section}
\numberwithin{example}{section}
\numberwithin{remark}{section}

\section{Further Related Work}\label{app:related}

\textbf{Self-consistency and label-free training.} Self-consistency selects the most frequent answer among sampled reasoning paths \citep{wang2023selfconsistency}, and models improve when they are fine-tuned on their own majority-voted answers \citep{huang2023selfimprove}, trained with preferences between consistent and inconsistent answers \citep{prasad2025scpo} or rewarded by themselves \citep{yuan2024selfrewarding}. Label-free reinforcement learning also uses the confidence of the model \citep{zhao2025intuitor,prabhudesai2025rent}, the entropy of its outputs \citep{agarwal2025em,zhang2025empo} or its own consistency \citep{shafayat2025srt}, and its limits under scaling have been measured \citep{he2026unsupervised}. A further variant combines majority selection with a novelty term \citep{zhou2025evolrl}. All these rewards reinforce what the model already prefers. Our law addresses the majority reward on a fixed set of answers, where this feedback takes the form of a race between the leading answers.

\textbf{Methods against wrong majorities.} Several methods change the pseudo-label or the reward so that a wrong majority is less likely to win. Tools verify sampled attempts before they vote \citep{liao2026tool}, the reward is estimated from the whole distribution of answers \citep{du2026dare}, consensus is defined through an equilibrium of a replicator dynamics \citep{mehrish2026core}, uncertain consensus is replaced by selective and complementary rewards \citep{yan2026scrl}, a verifier coevolves with the generator \citep{pan2026coverrl}, and ambiguous samples are filtered with fixed advantages \citep{yu2026spurious}. Training can also be restricted to a small subspace of bias parameters \citep{vakada2026labelfree}. These methods are complementary to ours. A prediction before training tells which questions are still open, and the same chain can evaluate a modified reward before it is used.

\textbf{Dynamics of policy gradient.} For softmax policies, exact policy gradient converges to an optimal action \citep{mei2020softmax}, while stochastic policy gradient can settle on a suboptimal action depending on the baseline \citep{chung2021beyond}. Self-improvement through the model's own judgments sharpens the distribution toward its mode \citep{huang2025sharpening}, and the learning dynamics of fine-tuning can be explained through the empirical kernel of the logits \citep{ren2025learning}. Our setting is the stochastic and self-referential case. The reward is computed from the samples themselves, and the outcome is random even when the step is small.

\textbf{Linearized training and preconditioning.} Wide networks evolve as linear models in their parameters \citep{jacot2018ntk,lee2019wide}, a behavior that also arises when the scale of the model makes training lazy \citep{chizat2019lazy}. LoRA adapters initialized with $B=0$ start close to this linear behavior, and the kernel of our experiments changes little until the race is settled and grows later (Appendix~\ref{app:growth}). Adaptive methods have stochastic and mean-field descriptions \citep{malladi2022sdes,barakat2021adam}. We freeze the preconditioner of AdamW \citep{kingma2015adam,loshchilov2019decoupled} at its expected value at the start, which turns AdamW without momentum into a frozen kernel.

\textbf{Stochastic approximation and urns.} The frozen-kernel chain is a stochastic approximation of its mean flow with a constant step \citep{benaim1999dynamics,kushner2003stochastic}. Stochastic approximation avoids unstable points \citep{pemantle1990nonconvergence}, and nonlinear urns have been analyzed in the same framework \citep{laruelle2019nonlinear}. For us the question is not whether the chain leaves the tie but on which side, which needs the window of the noise and not only the flow.

\textbf{Exit problems.} The exit of a small-noise diffusion from a neighborhood of a repelling or hyperbolic equilibrium is a classical problem \citep{kifer1981exit,eizenberg1984exit,freidlin2012random}, and normal forms give the Gaussian law of the exit side \citep{almadamonter2011normal}. The edge of two leaders differs from these settings. It has a center direction along which the trailing mass decays only algebraically, and the orbit that matters converges to the edge without reaching it. We use center manifolds \citep{kelley1967stable,carr1981applications,vanderbauwhede1989centre} and a discrete exit theorem along the critical orbit (Theorem~\ref{thm:localexit}), with the martingale central limit theorem \citep{hall1980martingale}.

\section{Proofs for the Lock-In Law}\label{app:lockin}

Throughout this appendix the chain is the tabular chain $z_{k+1}=z_k+\eta\xi_k$ of Section~\ref{sec:lockin}. We keep the notation of Section~\ref{sec:setting}. Generic constants $C,C',c$ may change from line to line, and they depend only on $n$, $G$, $\omega$, $\delta$ and the start $z^0$ unless we say otherwise. Numbered constants $c_1,c_2,\dots$ and small positive constants $\varepsilon_1,\varepsilon_2,\dots$ keep their values within an appendix. We write $\mathcal F_k$ for the past up to step $k$ and $\E_k$ for the conditional expectation given $\mathcal F_k$.

\subsection{Updates, Vote Moments and the Advantage of the Leader}\label{app:votes}

We use three properties of the update \eqref{eq:update}.
\begin{enumerate}[label=(P\arabic*),leftmargin=*,itemsep=1pt]
\item $\sum_a\xi_a=0$, and $|\xi_a|\le C_\xi$ with $C_\xi=\max\{\omega(f)f(1-f)\mid f=\frac1G,\dots,\frac{G-1}G\}$.
\item If $z$ is replaced by its image under a permutation of the answers, the law of $\xi$ is replaced by its image under the same permutation.
\item The mean $\mu(p)=\E_p\xi$ and the covariance $\Cov_p(\xi)$ are polynomials in $p$, so they are smooth functions of $z$ with bounded derivatives of every order.
\end{enumerate}
For (P1), $N_L=M=fG$ gives $\sum_a\xi_a=\omega(f)(f-f\sum_aN_a/G)=0$. For $a\ne L$ one has $N_a\le G-M$, so $|\xi_a|=\omega(f)fN_a/G\le\omega(f)f(1-f)$, and $\xi_L=\omega(f)f(1-f)$. If $f=1$ then $\xi=0$. Property (P2) holds because the sampling rule and the tie break do not depend on the labels. For (P3), every moment is a finite sum over count vectors,
\begin{equation}\label{eq:moments}
\E_p\,h(\xi)=\sum_{N_1+\dots+N_n=G}\frac{G!}{\prod_aN_a!}\prod_ap_a^{N_a}\,\frac1{|\cL(N)|}\sum_{L\in\cL(N)}h\big(\xi(N,L)\big),
\end{equation}
where $\cL(N)$ is the set of answers with the largest count. Formula \eqref{eq:moments} with $h$ the coordinates and their products is how we compute all constants of this paper.

The next lemma is the tool behind every statement about trailing answers. It says that in expectation the most likely answer gains log-odds on every other answer.

\begin{lemma}[The advantage of the leader]\label{lem:leader}
Let $n\ge2$, $G\ge3$ and $\omega$ positive. If $p_i=\max_ap_a$ and $p_j\le p_i$, then
\begin{equation}\label{eq:leader}
\mu_i(p)-\mu_j(p)\ \ge\ c_G\,p_ip_j\big(p_i^{G-2}-p_j^{G-2}\big),\qquad c_G=\frac{2(G-1)}{G}\,\omega\Big(\frac{G-1}{G}\Big).
\end{equation}
\end{lemma}

\begin{proof}
For a count vector $N$ define the transfers
\begin{equation}\label{eq:transfer}
j_{ab}(N)=\frac{\omega(M/G)\,M}{G^2\,|\cL(N)|}\Big(N_b\,\one\{N_a=M\}-N_a\,\one\{N_b=M\}\Big),\qquad J_{ab}(p)=\E_p\,j_{ab}(N),
\end{equation}
with $j_{ab}=0$ if $M=G$. Summing the bracket over $b\ne a$ gives $G\one\{N_a=M\}-|\cL(N)|N_a$. Since $\E[\xi_a\mid N]=\omega(f)\big(f\one\{N_a=M\}/|\cL(N)|-fN_a/G\big)$, this shows
\begin{equation}\label{eq:transfer2}
J_{ab}=-J_{ba},\qquad \mu_a=\sum_{b\ne a}J_{ab}.
\end{equation}

\emph{Step 1. Pairs.} Let $p_a\ge p_b$, and pair every count vector $N$ with the vector $N^{ab}$ in which $N_a$ and $N_b$ are exchanged. Both have the same multinomial coefficient, the same $M$ and the same $|\cL(N)|$. A nonzero transfer has one of $N_a,N_b$ equal to $M$ and the other equal to some $l<M$. If $N_a=M$ and $N_b=l$, then $j_{ab}(N)=-j_{ab}(N^{ab})=\omega(M/G)Ml/(G^2|\cL(N)|)$, so the pair contributes a nonnegative factor times
\[
p_a^Mp_b^l-p_a^lp_b^M\ \ge\ 0 .
\]
Hence $J_{ab}\ge0$. The pair with the counts $G-1$ and $1$ on $a$ and $b$ and zero elsewhere has $M=G-1$, $|\cL(N)|=1$, $l=1$ and the coefficient $G$, so it contributes exactly
\begin{equation}\label{eq:pairbound}
\frac{G-1}{G}\,\omega\Big(\frac{G-1}{G}\Big)\big(p_a^{G-1}p_b-p_ap_b^{G-1}\big).
\end{equation}
All other pairs contribute nonnegative amounts, so \eqref{eq:pairbound} is a lower bound for $J_{ab}$.

\emph{Step 2. A third answer.} We show that
\begin{equation}\label{eq:third}
J_{ik}-J_{jk}\ \ge\ 0\qquad\text{if } p_i\ge p_j \text{ and } p_i\ge p_k .
\end{equation}
Fix all counts outside $\{i,j,k\}$ and sum over the assignments of a multiset of three counts to $(i,j,k)$. All assignments have the same multinomial coefficient, the same $M$ and the same $|\cL(N)|$, and we drop this common nonnegative factor. Write $x=p_i$, $y=p_j$ and $z=p_k$, so that $x\ge y$ and $x\ge z$. Up to the common factor, an assignment $(n_i,n_j,n_k)$ contributes
\[
x^{n_i}y^{n_j}z^{n_k}\Big(n_k\big(\one\{n_i=M\}-\one\{n_j=M\}\big)-(n_i-n_j)\,\one\{n_k=M\}\Big).
\]
If no count of the triple equals $M$, every term vanishes, and if all three equal $M$, the bracket vanishes. If exactly two counts equal $M$ and the third equals $l<M$, the three assignments sum to
\[
l\,z^M\big(x^My^l-x^ly^M\big)\ \ge\ 0 .
\]
If exactly one count equals $M$ and the other two are $l_1\ge l_2$ with $l_1<M$, the six assignments sum to
\begin{equation}\label{eq:Tmab}
S=l_2\,z^{l_2}\big(x^My^{l_1}-x^{l_1}y^M\big)+l_1\,z^{l_1}\big(x^My^{l_2}-x^{l_2}y^M\big)-(l_1-l_2)\,z^M\big(x^{l_1}y^{l_2}-x^{l_2}y^{l_1}\big).
\end{equation}
If $l_1=l_2$, each distinct assignment is counted twice, which does not change the sign. With $s_1=l_1-l_2\ge0$ and $s_2=M-l_1>0$, a direct expansion gives the identity
\begin{equation}\label{eq:positive-orbit}
S=x^{l_2}y^{l_2}z^{l_2}\Big\{y^{s_1}\big[l_2\,x^{s_1}+l_1\,z^{s_1}\big]\big(x^{s_2}-y^{s_2}\big)+z^{s_1}\big[l_2\,x^{s_2}+s_1\big(x^{s_2}-z^{s_2}\big)\big]\big(x^{s_1}-y^{s_1}\big)\Big\}.
\end{equation}
Every factor in braces is nonnegative because $x\ge y$ and $x\ge z$. This proves \eqref{eq:third}.

\emph{Step 3. Conclusion.} By \eqref{eq:transfer2},
\[
\mu_i-\mu_j=\sum_{b\ne i}J_{ib}-\sum_{b\ne j}J_{jb}=2J_{ij}+\sum_{k\ne i,j}\big(J_{ik}-J_{jk}\big)\ \ge\ 2J_{ij},
\]
by \eqref{eq:third}. Inserting the lower bound \eqref{eq:pairbound} with $a=i$ and $b=j$ gives \eqref{eq:leader}.
\end{proof}

We use \eqref{eq:leader} in two linear forms. Let $1$ be a most likely answer, $b\ne1$ and $\zeta_b=z_1-z_b\ge0$, so that $p_b=p_1e^{-\zeta_b}$. Since $1-e^{-(G-2)\zeta}\ge(G-2)\zeta e^{-(G-2)\zeta}$,
\begin{equation}\label{eq:leaderlin}
\E_p[\xi_1-\xi_b]\ \ge\ c_G(G-2)\,p_1^G\,\zeta_b\,e^{-(G-1)\zeta_b},\qquad \E_p[\xi_1-\xi_b]\ \ge\ c_G\,p_1^{G-1}\big(1-e^{-(G-2)\zeta_b}\big)\,p_b .
\end{equation}
Since $p_1\ge1/n$, both bounds are uniform on the simplex.

\begin{lemma}[Tied paths]\label{lem:tied}
Let $2\le m\le n$, let $z^0$ satisfy $z^0_1=\dots=z^0_m$ and $p_1(z^0)>\max_{a>m}p_a(z^0)$, and let $\bar z$ solve $\bar z'=\mu(\bar z)$ from $z^0$. Let $q(t)$ be the probability of answer $1$ along $\bar z$ and $\rho_m=\sum_{a>m}p_a$.
\begin{enumerate}[label=(\roman*),leftmargin=*,itemsep=1pt]
\item $\bar z_1=\dots=\bar z_m$ for all $t\ge0$.
\item For every $a>m$, $\bar z_1-\bar z_a$ is strictly increasing and tends to infinity. Hence $\rho_m$ decreases to $0$ and $q$ increases to $1/m$.
\item At every $z$ with $z_1=\dots=z_m$ and $p_1(z)\ge p_a(z)$ for all $a$, the rate satisfies $r(z)\ge c_G(G-2)\,p_1(z)^G$. In particular
\begin{equation}\label{eq:rlow}
r(t)\ \ge\ r_{\rm low}:=c_G(G-2)\,q(0)^G\ >\ 0\qquad\text{along }\bar z .
\end{equation}
\end{enumerate}
\end{lemma}

\begin{proof}
(i) By (P2), the set $\{z_1=\dots=z_m\}$ is invariant under the flow.

(ii) While $\bar z_1>\bar z_a$ for all $a>m$, the answers $1,\dots,m$ are most likely answers, and Lemma~\ref{lem:leader} with $b=a$ gives $(\bar z_1-\bar z_a)'>0$. Let $t_+$ be the supremum of the times up to which this holds. If $t_+<\infty$, then at $t_+$ every margin is at least its positive initial value, which contradicts the definition of $t_+$. So $t_+=\infty$. Fix $\varepsilon_1\in(0,1)$ with $\max_{a>m}p_a(z^0)\le(1-\varepsilon_1)p_1(z^0)$. Then $e^{-(\bar z_1-\bar z_a)}\le1-\varepsilon_1$ for all $t$, and $q(t)\ge q(0)$ because $q=1/(m+\sum_{a>m}e^{-(\bar z_1-\bar z_a)})$. The second bound of \eqref{eq:leaderlin} gives
\[
\big(e^{\bar z_1-\bar z_a}\big)'=e^{\bar z_1-\bar z_a}\,\E[\xi_1-\xi_a]\ \ge\ c_Gq(0)^{G}\big(1-(1-\varepsilon_1)^{G-2}\big)\ =:\ c_1>0 ,
\]
so $e^{\bar z_1-\bar z_a}\ge e^{\bar z_1(0)-\bar z_a(0)}+c_1t$. The claims on $\rho_m$ and $q$ follow from $mq+\rho_m=1$ and $\rho_m/(mq)=\frac1m\sum_{a>m}e^{-(\bar z_1-\bar z_a)}$.

(iii) Let $\zeta>0$ be small. At $z+\frac12\zeta e$, answer $1$ is the unique most likely answer, and \eqref{eq:leader} with $i=1$, $j=2$ gives
\[
\E d\ \ge\ c_G\,p_1p_2\big(p_1^{G-2}-p_2^{G-2}\big)=c_G(G-2)\,p_1(z)^{G-1}\big(p_1-p_2\big)+o(\zeta).
\]
The derivative of $p_1-p_2$ in $\zeta$ at $\zeta=0$ is $p_1(z)$, and $\E d=0$ at $\zeta=0$ by (P2). Dividing by $\zeta$ and letting $\zeta\to0$ gives $r(z)\ge c_G(G-2)p_1(z)^G$. Along $\bar z$, $q(t)\ge q(0)$ by (ii).
\end{proof}

Let $E_m$ be the space of vectors whose first $m$ entries sum to zero and whose other entries vanish, and let $H_m$ be the orthogonal projection onto $E_m$. Its orthogonal complement $E_m^\perp$ is the space of vectors with equal first $m$ entries.

\begin{lemma}[Symmetry at a tie]\label{lem:schur}
Let $z\in E_m^\perp$. Then $H_m\mu(z)=0$. The Jacobian $\partial_z\mu(z)$ maps $E_m$ to $E_m$ and $E_m^\perp$ to $E_m^\perp$, and it acts on $E_m$ as $r(z)$ times the identity. Hence $H_m\partial_z\mu(z)=r(z)H_m$ on all of $\R^n$. The covariance satisfies $H_m\Cov_z(\xi)H_m=\frac12v(z)H_m$ and $H_m\Cov_z(\xi)(\Id-H_m)=0$.
\end{lemma}

\begin{proof}
The symmetric group of the first $m$ coordinates acts on $\R^n$. Its fixed vectors form $E_m^\perp$, and $E_m$ is its standard representation, which is absolutely irreducible. By (P2), $\mu$ commutes with the action, and $z$ is fixed by it, so $\mu(z)$ is fixed, which is the first claim. The Jacobian at $z$ commutes with the action. By Schur's lemma it maps the trivial part $E_m^\perp$ to itself and $E_m$ to itself, and it is a multiple of the identity on $E_m$. Evaluating it on $\frac12e$ identifies the multiple with $r(z)$. The covariance is an invariant symmetric bilinear form, and the same argument shows that it is block diagonal with a multiple of the identity on $E_m$. Since $\Var(d)=v(z)$ and $|e|^2=2$, the multiple is $\frac12v(z)$.
\end{proof}

\subsection{Two Leaders and the Thurstone Law}\label{app:thurstone}

We prove the law for $m\ge2$ tied leaders, which contains Theorem~\ref{thm:lockin} and Corollary~\ref{cor:thurstone}.

\begin{theorem}[Tied leaders]\label{thm:multi}
Let $G\ge3$, $\omega$ positive and $2\le m\le n$. Let $z^0$ satisfy $z^0_1=\dots=z^0_m$ and $p_1(z^0)>\max_{a>m}p_a(z^0)$, let $\bar z$ be the tied path of Lemma~\ref{lem:tied}, and let $\Sig^2=\int_0^\infty v\,e^{-2\cI}dt$ along it. Then $0<\Sig<\infty$. Start the chain from $z^0+\sqrt\eta\,\beta$ with $\beta\in\R^n$, and let $Y_1,\dots,Y_m$ be independent standard normal variables. As $\eta\to0$, for every $i\le m$,
\begin{equation}\label{eq:multi}
\Pp(W=i)\to\Pp\Big(\beta_i+\tfrac{\Sig}{\sqrt2}Y_i>\beta_j+\tfrac{\Sig}{\sqrt2}Y_j\ \text{ for every } j\le m \text{ with } j\ne i\Big),
\end{equation}
and $\Pp(W>m)\to0$.
\end{theorem}

The proof rests on two lemmas. The first describes the window, and the second shows that a lead of $c_2\sqrt\eta$ is not lost.

\begin{lemma}[A finite window]\label{lem:window}
In the setting of Theorem~\ref{thm:multi}, fix $T>0$, let $k_T=\lfloor T/\eta\rfloor$ and $\sigma_T^2=\int_0^Tv\,e^{-2\cI}dt$. As $\eta\to0$, $\sup_{k\le k_T}|z_k-\bar z(k\eta)|\to0$ in probability, and
\begin{equation}\label{eq:windowlimit}
\frac{H_mz_{k_T}}{\sqrt\eta}\ \Longrightarrow\ e^{\cI(T)}\Big(H_m\beta+\frac{\sigma_T}{\sqrt2}\,H_mY\Big),
\end{equation}
where $Y$ is a standard normal vector in $\R^n$.
\end{lemma}

\begin{proof}
Write the chain as
\begin{equation}\label{eq:mart}
z_{k+1}=z_k+\eta\mu(z_k)+\eta\tilde\xi_k,\qquad \tilde\xi_k=\xi_k-\mu(z_k),
\end{equation}
where $\tilde\xi_k$ are martingale differences bounded by $2C_\xi\sqrt n$. Since $\mu$ is Lipschitz and $|z_0-z^0|\le C\sqrt\eta$, Doob's inequality for $\eta\sum_{i<k}\tilde\xi_i$ and Gronwall's lemma give
\begin{equation}\label{eq:track}
\E\sup_{k\le k_T}\big|z_k-\bar z(k\eta)\big|^2\ \le\ C_T\,\eta ,
\end{equation}
which proves the first claim. Let $y_k=H_mz_k$ and $z^{\rm s}_k=(\Id-H_m)z_k\in E_m^\perp$. Since $H_m\bar z=0$, \eqref{eq:track} gives $\E|y_k|^2\le C_T\eta$. By Lemma~\ref{lem:schur} and a Taylor expansion at $z^{\rm s}_k$,
\begin{equation}\label{eq:taylorwindow}
H_m\mu(z_k)=r(z^{\rm s}_k)\,y_k+O\big(|y_k|^2\big),\qquad \big|r(z^{\rm s}_k)-r(k\eta)\big|\le C\,\big|z_k-\bar z(k\eta)\big| .
\end{equation}
Compare $y_k$ with the linear recursion
\begin{equation}\label{eq:linrec}
\tilde y_{k+1}=\big(1+\eta r(k\eta)\big)\tilde y_k+\eta\,H_m\tilde\xi_k,\qquad \tilde y_0=y_0=\sqrt\eta\,H_m\beta .
\end{equation}
The difference $u_k=y_k-\tilde y_k$ satisfies $u_{k+1}=(1+\eta r(k\eta))u_k+\eta\big[(r(z^{\rm s}_k)-r(k\eta))y_k+O(|y_k|^2)\big]$. By \eqref{eq:track}, the Cauchy-Schwarz inequality and Gronwall's lemma,
\begin{equation}\label{eq:lindiff}
\E|u_k|\ \le\ C_T\,\eta\sum_{i<k}\E\Big(\big|z_i-\bar z(i\eta)\big|\,|y_i|+|y_i|^2\Big)\ \le\ C_T'\,\eta\ =\ o(\sqrt\eta)\qquad(k\le k_T).
\end{equation}
By variation of constants, $\tilde y_{k_T}/\sqrt\eta$ is $\prod_{i<k_T}(1+\eta r(i\eta))\,H_m\beta$ plus a martingale with increments of size $O(\sqrt\eta)$ and the conditional covariance
\[
\sum_{i<k_T}\eta\prod_{i<j<k_T}\big(1+\eta r(j\eta)\big)^2\,H_m\Cov_{z_i}(\xi)H_m .
\]
Since $\Cov_{z_i}(\xi)\to\Cov_{\bar z(i\eta)}(\xi)$ in $L^1$ by \eqref{eq:track}, Lemma~\ref{lem:schur} shows that this covariance converges in $L^1$ to $\int_0^Te^{2(\cI(T)-\cI(t))}\frac12v(t)\,dt\,H_m=\frac12e^{2\cI(T)}\sigma_T^2\,H_m$. The martingale central limit theorem \citep{hall1980martingale} gives \eqref{eq:windowlimit}.
\end{proof}

\begin{lemma}[No return]\label{lem:noreturn}
For every $\varepsilon_2>0$ there are $c_2>0$ and $\varepsilon_3>0$ with the following property. Let $\eta\le\varepsilon_3$, and let the chain start at a point $z$ with an answer $i$ such that $z_i-z_j\ge c_2\sqrt\eta$ for every $j\ne i$. Then, with probability at least $1-\varepsilon_2$, answer $i$ is the first answer to reach probability $1-\delta$.
\end{lemma}

\begin{proof}
Write $\zeta_j=z_i-z_j$ for $j\ne i$. While all $\zeta_j$ are positive, $i$ is the unique most likely answer, $p_i\ge1/n$, and \eqref{eq:leaderlin} applies with $1$ replaced by $i$. Fix $\varepsilon_4\in(0,1]$ small, to be chosen below. Then
\begin{align}
\E[\xi_i-\xi_j]&\ \ge\ c_3\,\zeta_j &&\text{if }0<\zeta_j\le\varepsilon_4,\quad c_3=c_G(G-2)n^{-G}e^{-(G-1)},\label{eq:B1}\\
\E[\xi_i-\xi_j]&\ \ge\ 0,\quad \E[\xi_i-\xi_j]\ \ge\ c_4\,p_j &&\text{if }\zeta_j\ge\tfrac12\varepsilon_4,\quad c_4=c_Gn^{1-G}\big(1-e^{-(G-2)\varepsilon_4/2}\big).\label{eq:B2}
\end{align}
Moreover $\E(\xi_i-\xi_j)^2\le c_5:=4C_\xi^2$, and we put $c_6=2c_5$.

\emph{Step 1. Leaving the window.} Let $k_{\rm low}$ be the first step at which some $\zeta_j$ is below $\frac12c_2\sqrt\eta$, and $k^{\rm up}_j$ the first step at which $\zeta_j>\varepsilon_4$. Fix $j$ and put $U=\zeta_j/\sqrt\eta$. While $U\in[\frac12c_2,\varepsilon_4/\sqrt\eta]$, the drift of $U$ is at least $\eta c_3U$ by \eqref{eq:B1}, its second moment is at most $\eta c_5$, and $|U_{k+1}-U_k|\le2C_\xi\sqrt\eta$. Let
\begin{equation}\label{eq:Psi}
F(U)=\int_U^\infty e^{-c_3s^2/c_6}\,ds .
\end{equation}
Then $F'(U)=-e^{-c_3U^2/c_6}$ and $F''(U)=(2c_3U/c_6)e^{-c_3U^2/c_6}$, and a Taylor expansion to second order with a third-order remainder gives
\begin{equation}\label{eq:Psidrift}
\E_k\big[F(U_{k+1})\big]-F(U_k)\ \le\ \eta\,e^{-c_3U_k^2/c_6}\Big[-c_3U_k\Big(1-\frac{c_5}{c_6}\Big)+C\sqrt\eta\,\big(1+U_k^2\big)\Big].
\end{equation}
Since $\sqrt\eta\,U_k\le\varepsilon_4$, the bracket is at most $-\frac12c_3U_k+C\varepsilon_4 U_k+C\sqrt\eta$, which is negative if $\varepsilon_4$ and $\eta$ are small and $c_2\ge2$. So $F(U)$, stopped at $k_{\rm low}\wedge k^{\rm up}_j$, is a supermartingale. Hence the probability that $\zeta_j$ falls below $\frac12c_2\sqrt\eta$ before it exceeds $\varepsilon_4$ is at most $F(c_2)/F(\frac12c_2)$. A union bound over $j$ gives
\begin{equation}\label{eq:leavewindow}
\Pp\big(k_{\rm low}<\max_jk^{\rm up}_j\big)\ \le\ (n-1)\,\frac{F(c_2)}{F(\frac12c_2)}\ \longrightarrow\ 0\qquad(c_2\to\infty).
\end{equation}

\emph{Step 2. The time to leave.} While $U\in[\frac12c_2,\varepsilon_4/\sqrt\eta]$, the drift of $\log\zeta_j$ is at least $\eta(c_3-C/U^2)\ge\frac12\eta c_3$ for large $c_2$, and its increments are at most $C\sqrt\eta/c_2$. By the Azuma-Hoeffding inequality, with probability tending to one, every $\zeta_j$ exceeds $\varepsilon_4$ within $k_1=\lceil4\log(1/\eta)/(\eta c_3)\rceil$ steps.

\emph{Step 3. No return from $\varepsilon_4$.} While $i$ is the unique most likely answer, every $\zeta_j$ has a nonnegative drift by \eqref{eq:B2}, and its increments are at most $2C_\xi\eta$. By the maximal Azuma-Hoeffding inequality, the probability that some $\zeta_j$ falls from above $\varepsilon_4$ to below $\frac12\varepsilon_4$ within $k_2$ steps is at most
\begin{equation}\label{eq:noreturn}
2(n-1)\exp\Big(-\frac{\varepsilon_4^2}{128\,C_\xi^2\,k_2\,\eta^2}\Big),
\end{equation}
which tends to zero for $k_2=O(\log(1/\eta)/\eta)$.

\emph{Step 4. The decision.} Let $x=\sum_{j\ne i}e^{-\zeta_j}=(1-p_i)/p_i$ and suppose that all $\zeta_j\ge\frac12\varepsilon_4$. Since $|\eta(\xi_i-\xi_j)|\le1$ for small $\eta$, the inequality $e^{-x}\le1-x+x^2$ for $|x|\le1$ gives
\[
\E_k\,x_{k+1}\ \le\ \sum_{j\ne i}e^{-\zeta_j}\Big(1-\eta\,\E_k[\xi_i-\xi_j]+C\eta^2\Big).
\]
By \eqref{eq:B2} and the Cauchy-Schwarz inequality,
\[
\sum_{j\ne i}e^{-\zeta_j}\,\E_k[\xi_i-\xi_j]\ \ge\ c_4\sum_{j\ne i}e^{-\zeta_j}p_j=c_4\,p_i\sum_{j\ne i}e^{-2\zeta_j}\ \ge\ \frac{c_4\,x^2}{n(n-1)} .
\]
While $p_i<1-\delta$, one has $x>\delta$. Therefore
\begin{equation}\label{eq:Upsilon}
\E_k\,x_{k+1}\ \le\ x_k\big(1-\eta c_7\big),\qquad c_7=\frac{c_4\,\delta}{n(n-1)}-C\eta\ >\ 0 .
\end{equation}
Since $x\le n-1$ at the start of this phase, Markov's inequality shows that the probability that $p_i<1-\delta$ throughout $k_3$ further steps is at most $(n-1)e^{-c_7\eta k_3}/\delta$, which is small for $k_3=C/\eta$ with a large $C$.

\emph{Conclusion.} While $i$ is the unique most likely answer, every other answer has probability below $\frac12<1-\delta$. Combining Steps 1 to 4, answer $i$ is the first answer to reach $1-\delta$ with probability at least $1-(n-1)F(c_2)/F(\frac12c_2)-o(1)$, and the claim follows by choosing $c_2$ large.
\end{proof}

\begin{proof}[Proof of Theorem~\ref{thm:multi}]
By Lemma~\ref{lem:tied}(iii), $r\ge r_{\rm low}>0$, and $v\le4C_\xi^2$, so $\Sig<\infty$ and $\cI(T)\to\infty$. At $z^0$ the count vector with $N_1=G-1$ and $N_2=1$ has positive probability and gives $d\ne0$, so $v(0)>0$ and $\Sig>0$.

Let $\varepsilon_2>0$, and let $c_2$ and $\varepsilon_3$ be given by Lemma~\ref{lem:noreturn}. Fix $T>0$. Along $\bar z$ the largest probability is $q(t)\le\frac1m\le\frac12<1-\delta$, so by Lemma~\ref{lem:window} no answer reaches $1-\delta$ before step $k_T$, with probability tending to one. Let $\zeta_{>m}=\log\big(p_1(z^0)/\max_{a>m}p_a(z^0)\big)>0$. By Lemma~\ref{lem:tied}, $\bar z_1(T)-\bar z_a(T)\ge\zeta_{>m}$ for $a>m$, so $z_{k_T,i}-z_{k_T,a}\ge\frac12\zeta_{>m}$ for all $i\le m$ and $a>m$, with probability tending to one. Let $U=H_mz_{k_T}/\sqrt\eta$, so that $z_{k_T,i}-z_{k_T,j}=\sqrt\eta(U_i-U_j)$ for $i,j\le m$. Let $\cU_i$ be the event that $U_i-U_j\ge c_2$ for every $j\le m$ with $j\ne i$. On $\cU_i$, Lemma~\ref{lem:noreturn} and the strong Markov property show that $W=i$ with conditional probability at least $1-\varepsilon_2$. Hence
\begin{equation}\label{eq:multierr}
\big|\Pp(W=i)-\Pp(\cU_i)\big|\ \le\ \varepsilon_2+\Pp\Big(\min_{j\ne l\le m}|U_j-U_l|<c_2\Big)+o(1).
\end{equation}
Let $\hat U=H_m\beta+\frac{\sigma_T}{\sqrt2}H_mY$. For $j,l\le m$, $\hat U_j-\hat U_l=\beta_j-\beta_l+\frac{\sigma_T}{\sqrt2}(Y_j-Y_l)$ has a bounded density. By Lemma~\ref{lem:window}, as $\eta\to0$, $\Pp(\cU_i)$ tends to the probability that $\hat U_i-\hat U_j\ge c_2 e^{-\cI(T)}$ for all $j\ne i$, and the last probability in \eqref{eq:multierr} tends to the probability that $\min_{j\ne l}|\hat U_j-\hat U_l|<c_2 e^{-\cI(T)}$. Now let $T\to\infty$. Then $e^{-\cI(T)}\to0$ and $\sigma_T\to\Sig$, so the first limit tends to the right side of \eqref{eq:multi} and the second tends to zero. Finally let $\varepsilon_2\to0$. The right sides of \eqref{eq:multi} sum to one over $i\le m$, hence $\Pp(W>m)\to0$.
\end{proof}

\begin{proposition}[Uniformity in the gap]\label{prop:uniform}
In the setting of Theorem~\ref{thm:lockin}, $\sup_{\beta\in\R}\big|\Pp(W=2)-\Phic(\beta/\Sig)\big|\to0$ as $\eta\to0$.
\end{proposition}

\begin{proof}
Let $\varepsilon_2>0$, and choose $c_8\ge c_2$ with $\Phic(c_8/\Sig)<\varepsilon_2$, where $c_2$ is given by Lemma~\ref{lem:noreturn} for $\varepsilon_2$. Let $\zeta_{>2}=\log(p_1(z^0)/\max_{a\ge3}p_a(z^0))$, or $\zeta_{>2}=\infty$ if $n=2$. If $\beta\ge c_8$, the start has $z_1-z_2=\beta\sqrt\eta\ge c_2\sqrt\eta$ and $z_1-z_a\ge\zeta_{>2}\ge c_2\sqrt\eta$ for $a\ge3$ and small $\eta$, so Lemma~\ref{lem:noreturn} gives $\Pp(W=2)\le\varepsilon_2$ uniformly in $\beta\ge c_8$. The case $\beta\le-c_8$ is symmetric. On $[-c_8,c_8]$, the proof of Theorem~\ref{thm:multi} applies to every sequence $\beta_\eta\to\beta$, because Lemma~\ref{lem:window} holds for the starts $z^0+\frac12\beta_\eta\sqrt\eta e$ with the same limit. Hence $\Pp(W=2)\to\Phic(\beta/\Sig)$ along every such sequence. If the convergence were not uniform on $[-c_8,c_8]$, there would be $\varepsilon_5>0$ and a sequence $\beta_\eta$ with $|\Pp(W=2)-\Phic(\beta_\eta/\Sig)|\ge\varepsilon_5$, and a convergent subsequence together with the continuity of $\Phic$ would give a contradiction.
\end{proof}

\begin{proof}[Proof of Theorem~\ref{thm:lockin} and Corollary~\ref{cor:thurstone}]
Corollary~\ref{cor:thurstone} is Theorem~\ref{thm:multi}. For Theorem~\ref{thm:lockin}, apply Theorem~\ref{thm:multi} with $m=2$ and the start $z^0+\frac12\beta\sqrt\eta e$, which is $z^0+\sqrt\eta\,\beta'$ with $\beta'=\frac12\beta e$. Then
\[
\Pp(W=2)\to\Pp\Big(\tfrac{\Sig}{\sqrt2}(Y_2-Y_1)>\beta\Big)=\Phic(\beta/\Sig),
\]
since $Y_2-Y_1$ has variance $2$. Proposition~\ref{prop:uniform} gives the uniformity, and Proposition~\ref{prop:eta2} below gives the error for $n=2$.
\end{proof}

\subsection{The Error of Order $\eta$}\label{app:eta}

For two answers the state of the chain is the gap $\zeta$, and one step changes it by $\eta d$. By (P2) and (P3), the odd moments of $d$ are odd functions of $\zeta$ and the even moments are even, so that
\begin{equation}\label{eq:taylorcoef}
\E d=r\zeta+r_3\zeta^3+O(\zeta^5),\quad \E d^2=v+v_2\zeta^2+O(\zeta^4),\quad \E d^3=v_3\zeta+O(\zeta^3),\quad \E d^4=v_4+O(\zeta^2),
\end{equation}
where $r$ and $v$ are the rate and the variance at the tie, and $\Sig^2=v/(2r)$. For functions $\varphi$ of a real variable $b$ define
\begin{equation}\label{eq:T0T1}
\mathcal T_0\varphi=r\,b\varphi'+\frac v2\varphi'',\qquad \mathcal T_1\varphi=r_3\,b^3\varphi'+\frac{v_2}2\,b^2\varphi''+\frac{v_3}6\,b\varphi'''+\frac{v_4}{24}\,\varphi'''' .
\end{equation}

\begin{proposition}[The error for two answers]\label{prop:eta2}
Let $n=2$, $G\ge3$ and $\omega$ positive. Let $\varphi_0(b)=\Phic(b/\Sig)$, and let $\varphi_1$ be the solution of $\mathcal T_0\varphi_1=-\mathcal T_1\varphi_0$ that tends to zero at $\pm\infty$. For the start $\zeta(0)=\beta\sqrt\eta$,
\begin{equation}\label{eq:eta2}
\Pp(W=2)=\Phic(\beta/\Sig)+\eta\,\varphi_1(\beta)+O(\eta^2),
\end{equation}
uniformly in $\beta\in\R$.
\end{proposition}

\begin{proof}
\emph{Step 1. The corrector.} Note that $\mathcal T_0\varphi_0=0$. The equation for $\varphi_1$ is a first-order linear equation for $\varphi_1'$. Its solutions are $\varphi_{\rm p}+c_9+c_{10}\varphi_0$ for a particular solution $\varphi_{\rm p}$, so the two conditions at $\pm\infty$ fix $\varphi_1$. Since $\mathcal T_1\varphi_0$ is a polynomial times $e^{-b^2/(2\Sig^2)}$, so are $\varphi_1'$ and every derivative of $\varphi_0$. Since $\varphi_1(-\infty)=0$, the term proportional to $\varphi_0$ in $\varphi_1(b)=-\int_b^\infty \varphi_1'$ vanishes, so $\varphi_1$ is also a polynomial times $e^{-b^2/(2\Sig^2)}$.

\emph{Step 2. One step of the approximation.} Let $P_2(\zeta)$ be the probability that answer $2$ wins from the gap $\zeta$, and put
\begin{equation}\label{eq:hatP}
\hat P_2(\zeta)=\varphi_0(\zeta/\sqrt\eta)+\eta\,\varphi_1(\zeta/\sqrt\eta).
\end{equation}
Fix $\varepsilon_6>0$ small, with $\varepsilon_6<\zeta_{\rm d}=\log((1-\delta)/\delta)$ and $\varepsilon_6\le\varepsilon_4$, where $\varepsilon_4$ is the level of Lemma~\ref{lem:noreturn}. For $|\zeta|<\varepsilon_6$ and $b=\zeta/\sqrt\eta$, expand
\[
\E\big[\hat P_2(\zeta+\eta d)\big]-\hat P_2(\zeta)=\sum_{j=1}^{6}\frac{\eta^{j/2}}{j!}\,\E[d^j]\,\big(\varphi_0^{(j)}(b)+\eta \varphi_1^{(j)}(b)\big)+\text{(remainder)},
\]
and expand the moments of $d$ in $\zeta=\sqrt\eta\,b$ with \eqref{eq:taylorcoef}. By the parity of the moments, every term is an integer power of $\eta$. The terms of order $\eta$ give $\eta\,\mathcal T_0\varphi_0=0$, and the terms of order $\eta^2$ give $\eta^2(\mathcal T_0\varphi_1+\mathcal T_1\varphi_0)=0$. All remaining terms carry a factor $\eta^3$ times a polynomial in $b$ times derivatives of $\varphi_0$ or $\varphi_1$ evaluated within $2C_\xi\sqrt\eta$ of $b$. Since $|b|\sqrt\eta\le\varepsilon_6$, such a shift changes the Gaussian factor by at most $e^{2C_\xi\varepsilon_6/\Sig^2}$. Hence the residual satisfies
\begin{equation}\label{eq:residual}
\big|\E[\hat P_2(\zeta+\eta d)]-\hat P_2(\zeta)\big|\ \le\ C\,\eta^3\,g(\zeta/\sqrt\eta),\qquad g(b)=(1+|b|)^{c_{15}}e^{-b^2/(2\Sig^2)},
\end{equation}
for some integer $c_{15}$.

\emph{Step 3. Optional stopping.} Let $k_{\rm out}$ be the first step with $|\zeta_k|\ge\varepsilon_6$. Since $\varepsilon_6<\zeta_{\rm d}$, no decision happens before $k_{\rm out}$, so $P_2(\zeta_0)=\E P_2(\zeta_{k_{\rm out}})$. Optional stopping, which applies because $\E k_{\rm out}<\infty$ (Step 5), gives
\begin{equation}\label{eq:optional}
\hat P_2(\zeta_0)=\E\hat P_2(\zeta_{k_{\rm out}})-\E\sum_{k<k_{\rm out}}\Big(\E_k\hat P_2(\zeta_{k+1})-\hat P_2(\zeta_k)\Big).
\end{equation}

\emph{Step 4. The exit.} Suppose that $\zeta_{k_{\rm out}}\ge\varepsilon_6$. By Lemma~\ref{lem:leader}, $\E d\ge c_{11}>0$ at every gap in $[\frac12\varepsilon_6,\zeta_{\rm d}]$, and the increments of the gap are at most $2C_\xi\eta$. So $e^{-c_{14}\zeta_k/\eta}$, stopped when $\zeta_k$ leaves $(\frac12\varepsilon_6,\zeta_{\rm d})$, is a supermartingale for a small $c_{14}>0$, and the probability of reaching $\frac12\varepsilon_6$ before $\zeta_{\rm d}$ is at most $e^{-c_{14}\varepsilon_6/(2\eta)}$. Answer $2$ can win only after this event, so $P_2(\zeta_{k_{\rm out}})\le e^{-C/\eta}$. The case $\zeta_{k_{\rm out}}\le-\varepsilon_6$ is symmetric. The Gaussian tails of $\varphi_0$ and $\varphi_1$ show that $\hat P_2(\zeta_{k_{\rm out}})$ is within $e^{-C/\eta}$ of the same value $\one\{\zeta_{k_{\rm out}}<0\}$. With \eqref{eq:residual} and \eqref{eq:optional},
\begin{equation}\label{eq:errorsum}
\big|P_2(\zeta_0)-\hat P_2(\zeta_0)\big|\ \le\ 2e^{-C/\eta}+C\eta^3\,\E\sum_{k<k_{\rm out}}g(b_k),\qquad b_k=\zeta_k/\sqrt\eta .
\end{equation}

\emph{Step 5. The occupation of the window.} It remains to show $\E\sum_{k<k_{\rm out}}g(b_k)\le C/\eta$. Let $c_{12}=\max(1,\Sig^2)$. Before $k_{\rm out}$, the drift of $\log(c_{12}+b_k^2)$ per step is
\begin{equation}\label{eq:logdrift}
\eta\,\frac{2rb_k^4+(2rc_{12}-v)\,b_k^2+vc_{12}}{(c_{12}+b_k^2)^2}+O\big(\eta\varepsilon_6^2+\eta^2\big),
\end{equation}
where the error term collects the contributions of $r_3$, $v_2$ and the higher moments. Since $2rc_{12}-v=2r(c_{12}-\Sig^2)\ge0$, this drift is at least $\eta c_{13}>0$ when $\varepsilon_6$ and $\eta$ are small. Since $|b_{k+1}-b_k|\le1$, starting from $|b|\le b_0$, the expected number of steps before $|b|$ exceeds $2b_0$ is at most $C\log(2+b_0)/\eta$. Let $c_{16}\ge1$ be such that $F(2b)/F(b)\le\frac12$ for $b\ge c_{16}$, with $F$ from \eqref{eq:Psi}. Step 1 of the proof of Lemma~\ref{lem:noreturn} applies because $\varepsilon_6\le\varepsilon_4$, so for $b_0\ge c_{16}$ the probability of returning from $|b|\ge2b_0$ to $|b|\le b_0$ before $k_{\rm out}$ is at most $\frac12$. Hence the expected number of steps before $k_{\rm out}$ with $|b_k|\le b_0$ is at most $C\log(2+b_0)/\eta$ for $b_0\ge c_{16}$, and for $b_0<c_{16}$ the bound for $b_0=c_{16}$ applies. Summing over the shells $j\le|b|<j+1$ gives
\begin{equation}\label{eq:occupation}
\E\sum_{k<k_{\rm out}}g(b_k)\ \le\ \frac C\eta\sum_{j\ge0}g_j\log(3+j)\ \le\ \frac C\eta,
\end{equation}
where $g_j$ is the largest value of $g$ on the shell. With $g$ replaced by $1$ on the window $|b|<\varepsilon_6/\sqrt\eta$, the same argument gives $\E k_{\rm out}\le C\log(1/\eta)/\eta$.

\emph{Conclusion.} By \eqref{eq:errorsum} and \eqref{eq:occupation}, $|P_2(\zeta_0)-\hat P_2(\zeta_0)|=O(\eta^2)$ at $\zeta_0=\beta\sqrt\eta$, which is \eqref{eq:eta2}. The constants do not depend on $\beta$, and if $|\beta|\sqrt\eta\ge\varepsilon_6$, then $k_{\rm out}=0$ and the bound is immediate. This proves the uniformity in $\beta$.
\end{proof}

For $n=2$, $G=8$ and the mean baseline, every value of $d$ is a multiple of $1/32$, so $\Pp(W=2)$ can be computed exactly on a lattice. The coefficients are $r=\frac{105}{256}$, $r_3=-\frac{105}{512}$, $v=\frac{203}{1024}$, $v_2=-\frac{77}{1024}$, $v_3=\frac{18375}{262144}$ and $v_4=\frac{89509}{2097152}$, and Table~\ref{tab:eta2} in Appendix~\ref{app:numerics} confirms \eqref{eq:eta2}. For $n\ge3$ the symmetric coordinates fluctuate by $O(\sqrt\eta)$ during the window. Their first-order effect cancels by the swap symmetry, since $\Cov(\xi_1+\xi_2,d)=0$ at a tie, and simulations with eight answers show an error of order $\eta$ (Appendix~\ref{app:numerics}). We do not prove this order here.

\section{Proofs for the Kernel}\label{app:kernel}

In this appendix $K$ is a symmetric $n\times n$ matrix with $K\one=0$ and $\kappa=\frac12e^\top Ke>0$. The frozen-kernel chain is $z_{k+1}=z_k+\lr\,K\xi_k$ with $\eta=\lr\,\kappa$, and its mean flow is
\begin{equation}\label{eq:meanflow}
z'=f_K(z)=K\mu(p(z))/\kappa
\end{equation}
in the time $t=k\eta$. Constants follow the conventions of Appendix~\ref{app:lockin}. We use $\ehat=Ke/\kappa$, $\tilde e_a=e_a-\frac12(e_1+e_2)$ and $\chi_a=\ehat^\top \tilde e_a$ for $a\ge3$, and in addition
\begin{equation}\label{eq:QS}
Q_{ab}=\frac{\tilde e_a^\top K\tilde e_b}{\kappa},\qquad S=Q-\tfrac12\chi\chi^\top\qquad(a,b\ge3).
\end{equation}
The matrix $S$ is the Schur complement of the entry $e^\top Ke/\kappa=2$ in the Gram matrix of $\tilde e_3,\dots,w_n,e$ under $K/\kappa$. A symmetric matrix $M$ is strictly copositive if $x^\top Mx>0$ for every nonzero $x\ge0$. If $K$ is positive definite on $\one^\perp$, then the Gram matrix is positive definite because $\tilde e_3,\dots,w_n,e$ form a basis of $\one^\perp$, and so is $S$. Since $Q-S=\frac12\chi\chi^\top\succeq0$, the matrix $Q$ is strictly copositive whenever $S$ is. Every $u\in\one^\perp$ satisfies
\begin{equation}\label{eq:decomp}
u=\tfrac12\big(e^\top u\big)\,e+\sum_{a\ge3}u_a\,\tilde e_a ,
\end{equation}
as one checks coordinate by coordinate with $u_1+u_2=-\sum_{a\ge3}u_a$.

\subsection{The Gap Step}\label{app:gapstep}

\begin{proof}[Proof of Lemma~\ref{lem:gapstep}]
One step changes the gap by $\lr\,e^\top K\xi_k=\eta\,\ehat^\top\xi_k$. By \eqref{eq:decomp} with $u=\xi_k$ and $\ehat^{\,\top}e=e^\top Ke/\kappa=2$,
\begin{equation}\label{eq:gapproof}
\ehat^{\,\top}\xi_k=\tfrac12d_k\,\ehat^{\,\top}e+\sum_{a\ge3}\xi_{k,a}\,\ehat^{\,\top}\tilde e_a=d_k+\sum_{a\ge3}\chi_a\xi_{k,a},
\end{equation}
which is \eqref{eq:gapstep}. If $Ke=\kappa e$, then $\chi_a=e^\top \tilde e_a=0$. Conversely, let $\chi=0$. The vector $\ehat-e$ lies in $\one^\perp$, it is orthogonal to $e$ because $\ehat^{\,\top}e=e^\top e=2$, and it is orthogonal to every $\tilde e_a$ because $e^\top \tilde e_a=0$ and $\ehat^{\,\top}\tilde e_a=\chi_a=0$. Since $e,\tilde e_3,\dots,w_n$ span $\one^\perp$, $\ehat=e$.

Now let $z$ be tied. The swap of the answers $1$ and $2$ fixes $z$, maps $d$ to $-d$, and fixes $\xi_a$ for $a\ge3$. By (P2) the law of $\xi$ at $z$ is invariant under the swap. Hence $\E d=0$, $\Cov(d,\xi_a)=0$ for $a\ge3$, and every $\mu_a$ with $a\ge3$ is an even function of $\zeta$ at a fixed sum $z_1+z_2$, so that $\partial_\zeta\mu_a=0$ at the tie. Therefore the gap step has the mean $\eta(\E d+\chi\cdot\mu_\perp)=\eta\,\psi_K$, its derivative in $\zeta$ is $\eta(r+\chi\cdot\partial_\zeta\mu_\perp)=\eta r$, and its variance is
\[
\eta^2\Var\big(d+\chi\cdot\xi_\perp\big)=\eta^2\big(v+\chi^\top\Cov(\xi_\perp)\chi\big)=\eta^2v_K ,
\]
because the cross term vanishes.
\end{proof}

\begin{remark}[Two sources of spillover]\label{rem:sources}
At a tie $\mu_1=\mu_2=\mu_\circ=-\frac12\sum_{a\ge3}\mu_a$, and $\chi_a=\big(K_{a1}-K_{a2}-\frac12(K_{11}-K_{22})\big)/\kappa$. Hence
\begin{equation}\label{eq:sources}
\kappa\,\psi_K=\sum_{a\ge3}\big(K_{a1}-K_{a2}\big)\mu_a+\big(K_{11}-K_{22}\big)\mu_\circ .
\end{equation}
Near the edge every trailing answer loses votes, so $\mu_a<0$ and $\mu_\circ>0$. The second term says that when both leaders gain, the leader with the larger self-kernel gains more. The first term says that a trailing answer that loses votes drags down the leader with the larger kernel entry $K_{a\cdot}$. On the calibration runs of Qwen-1.5B, the self-kernel difference $|K_{11}-K_{22}|/(2\kappa)$ had a median of $0.39$, and the part of $Ke$ on the trailing letters had a median norm of $0.79$ relative to its part on the leaders. Both sources are therefore large in practice.
\end{remark}

\subsection{A Weak Anisotropy}\label{app:weak}

We prove the law for $m\ge2$ tied leaders, which contains Theorem~\ref{thm:weak}. Let $2\le m\le n$ and let $z^0$ satisfy $z^0_1=\dots=z^0_m>z^0_a$ for $a>m$. Let $\bar z$ be the tabular tied path from $z^0$ (Lemma~\ref{lem:tied}), $r$ its rate, $\cI=\int_0^tr$, and $v(t)=\Var(\xi_1-\xi_2)$ along $\bar z$, which equals $\Var(\xi_i-\xi_j)$ for all $i\ne j\le m$. Let $E_m$ and $H_m$ be as in Lemma~\ref{lem:schur}. For a symmetric $A$ with $A\one=0$ consider the chain
\begin{equation}\label{eq:weakchain}
z_{k+1}=z_k+\eta\big(H+\sqrt\eta A\big)\xi_k,\qquad z_0=z^0+\sqrt\eta\,\beta,\qquad\beta\in\R^n,
\end{equation}
and define the shift and the noise of the block
\begin{equation}\label{eq:bA}
b_A=\int_0^\infty e^{-\cI(t)}\,H_mA\,\mu(\bar z(t))\,dt,\qquad \sigma_m^2=\frac12\int_0^\infty e^{-2\cI(t)}v(t)\,dt .
\end{equation}
The integrals converge by \eqref{eq:rlow}, $b_A\in E_m$, and $\sigma_m>0$.

\begin{theorem}[The block law under a weak anisotropy]\label{thm:block}
Let $G\ge3$ and $\omega$ positive. For the chain \eqref{eq:weakchain} and every $i\le m$,
\begin{equation}\label{eq:blocklaw}
\lim_{\eta\to0}\Pp(W=i)=\Pp\Big(\beta_i+(b_A)_i+\sigma_mZ_i>\beta_j+(b_A)_j+\sigma_mZ_j\ \text{ for all } j\le m,\ j\ne i\Big),
\end{equation}
where $Z_1,\dots,Z_m$ are independent standard normal variables. Moreover $\Pp(W>m)\to0$, and the probability that no answer reaches $1-\delta$ tends to zero.
\end{theorem}

\begin{proof}
\emph{Step 1. The window.} Since $H\xi=\xi$, write $\tilde\xi_k=\xi_k-\mu(z_k)$, so that
\begin{equation}\label{eq:weakdecomp}
z_{k+1}=z_k+\eta\mu(z_k)+\eta\tilde\xi_k+\eta\sqrt\eta\,A\mu(z_k)+\eta\sqrt\eta\,A\tilde\xi_k .
\end{equation}
The deterministic correction $\eta\sqrt\eta A\mu$ accumulates to $O_T(\sqrt\eta)$ on $[0,T]$, and the additional martingale $\eta\sqrt\eta\sum A\tilde\xi_k$ has a maximal second moment of order $O_T(\eta^2)$. With Doob's inequality and Gronwall's lemma, as in \eqref{eq:track},
\begin{equation}\label{eq:weaktrack}
\E\max_{k\le T/\eta}\big|z_k-\bar z(k\eta)\big|^2\ \le\ C_T\,\eta .
\end{equation}
Put $U_k=H_mz_k/\sqrt\eta$. By Lemma~\ref{lem:schur}, $H_m\mu(\bar z)=0$ and $H_m\partial_z\mu(\bar z)=r\,H_m$ on all of $\R^n$, so a Taylor expansion gives
\begin{equation}\label{eq:weakU}
U_{k+1}=\big(1+\eta r(k\eta)\big)U_k+\eta\,H_mA\mu\big(\bar z(k\eta)\big)+\sqrt\eta\,H_m\tilde\xi_k+R_k .
\end{equation}
The quadratic Taylor remainder and the replacement of $A\mu(z_k)$ by $A\mu(\bar z(k\eta))$ have an accumulated scaled $L^1$ norm of order $O_T(\sqrt\eta)$ by \eqref{eq:weaktrack}, and the scaled extra martingale has an $L^2$ norm of order $O_T(\sqrt\eta)$. So the accumulated propagated remainder tends to zero in probability on $[0,T]$. Variation of constants and the martingale central limit theorem, with the covariance identity $H_m\Cov_{\bar z}(\xi)H_m=\frac12v\,H_m$ of Lemma~\ref{lem:schur}, give
\begin{equation}\label{eq:weakwindow}
e^{-\cI(T)}U_{\lfloor T/\eta\rfloor}\Longrightarrow Y_T\sim\mathcal N\Big(H_m\beta+\int_0^Te^{-\cI}\,H_mA\mu(\bar z)\,dt,\ \Big(\frac12\int_0^Te^{-2\cI}v\,dt\Big)H_m\Big).
\end{equation}
No independence between the correction noise and the votes is needed, because the predictable covariance converges by \eqref{eq:weaktrack} and the increments are of order $\sqrt\eta$. As $T\to\infty$, $Y_T\Rightarrow Y=H_m\beta+b_A+\sigma_mH_mZ$.

\emph{Step 2. Amplification of the selected leader.} Choose $\varepsilon_1>0$ below the initial margin between the leaders and the trailing answers, and fix a small $\varepsilon_2\in(0,\varepsilon_1/4)$. If $m=n$, drop every statement about $\varepsilon_1$. At a fixed $T$, by \eqref{eq:weaktrack}, all margins between block and trailing answers exceed $\frac34\varepsilon_1$ and all differences inside the block are below $\varepsilon_2$, with probability tending to one. Let $i$ be the largest block coordinate at step $k_T=\lfloor T/\eta\rfloor$ and $\zeta_j=z_i-z_j$ for $j\ne i$, and condition on
\begin{equation}\label{eq:Rcond}
\zeta_j(k_T)\ \ge\ c_1\sqrt\eta\qquad(j\le m,\ j\ne i).
\end{equation}
While $i$ is a most likely answer, Lemma~\ref{lem:leader} with $p_i\ge1/n$ and $p_j=p_ie^{-\zeta_j}$ gives a uniform outward drift,
\begin{equation}\label{eq:outward}
\mu_i-\mu_j\ \ge\ r_{\rm out}\,\zeta_j\quad(0<\zeta_j<\varepsilon_2),\qquad r_{\rm out}=c_G\,e^{-\varepsilon_2}n^{-G}\,\frac{1-e^{-(G-2)\varepsilon_2}}{\varepsilon_2}>0,
\end{equation}
since $(1-e^{-(G-2)s})/s$ decreases in $s$. The correction in \eqref{eq:weakchain} changes every mean drift of a log-odds by at most $C\sqrt\eta$ per unit of time, and every gap martingale has a conditional step variance of at most $C\eta^2$.

For $j\le m$ let $y_j=\zeta_j(k_T)$. Until $\zeta_j$ first reaches $\varepsilon_2$, declare a bad event if it falls to $y_j/2$, and after that, if it falls to $\varepsilon_2/2$. For every trailing answer declare a bad event if its margin falls below $\varepsilon_1/2$. Stop at the first bad event or at the horizon $t_\eta=C(1+\log(1/\eta))$ in time. Before this stopping time $i$ is a most likely answer, so \eqref{eq:outward} applies. For a gap that has not reached $\varepsilon_2$, take $c_1\ge4C/r_{\rm out}$. On $y_j/2\le \zeta_j<\varepsilon_2$ its drift is at least $\frac12r_{\rm out}\zeta_j$. With $q=(1+\frac12r_{\rm out}\eta)^{-1}$, the stopped discounted martingale satisfies
\[
\E\Big|\sum_{k\ge0}\eta\,q^{k+1}\nu_{k+1}\one\{k<\tau\}\Big|^2\ \le\ C\eta^2\sum_{k\ge0}q^{2k+2}\ \le\ C'\eta ,
\]
where $\nu_{k+1}$ are the martingale differences of $\zeta_j$ divided by $\eta$. By the maximal inequality, its fluctuation exceeds $y_j/2$ with probability at most $C\eta/y_j^2\le C/c_1^2$. Otherwise variation of constants gives $\zeta_j(k_T+k)\ge(1+\frac12r_{\rm out}\eta)^ky_j/2$ until $\zeta_j$ reaches $\varepsilon_2$ or the common stop, so $\zeta_j$ reaches $\varepsilon_2$ before $t_\eta$. For a gap that has reached $\varepsilon_2$, the tabular drift is nonnegative while $i$ is a most likely answer. Its adverse correction before $t_\eta$ is at most $C\sqrt\eta t_\eta=o(1)$, and its martingale has a maximal second moment of at most $C\eta t_\eta=o(1)$. Hence a fall to $\varepsilon_2/2$ has probability $o(1)$, and the same estimate excludes a fall of a trailing margin to $\varepsilon_1/2$. A union bound over the finitely many gaps shows that, with probability at least $1-C_m/c_1^2-o(1)$, every block competitor reaches $\varepsilon_2$ and stays above $\varepsilon_2/2$, and every trailing answer stays below $i$ by at least $\varepsilon_1/2$, up to $t_\eta$. If a threshold is crossed earlier, its answer can only be $i$, because no other answer can exceed $\frac12$ while $p_i$ is larger.

\emph{Step 3. Capture.} If no decision has occurred, the policy at the end of Step 2 lies in the compact set
\[
\mathcal K_i=\big\{p\mid p_i\ge e^{\varepsilon_3}p_j\ \text{for every}\ j\ne i\big\},\qquad \varepsilon_3=\min(\varepsilon_2,\varepsilon_1)/2.
\]
Every tabular orbit from $\mathcal K_i$ converges to the vertex $i$. Indeed, every ratio $p_j/p_i$ is nonincreasing by Lemma~\ref{lem:leader}, so the probabilities converge, and a positive limit of some $p_j$ would keep $\mu_i-\mu_j$ bounded below by a positive constant, which contradicts a finite limit of $\log(p_i/p_j)$. Continuous dependence and the compactness of $\mathcal K_i$ give a common finite time by which every such orbit has $p_i>1-\delta/2$. The chain tracks these orbits uniformly on that interval, with an extra drift of size $O(\sqrt\eta)$. Hence $W=i$ with probability $1-o(1)$, uniformly over the states at the end of Step 2.

\emph{Step 4. Removal of the cutoffs.} Let $\mathrm{gap}(y)$ be the largest coordinate of $y$ in the block minus the second largest. By Steps 2 and 3, for fixed $c_1$ and $T$,
\begin{equation}\label{eq:mismatch}
\Big|\Pp(W=i)-\Pp\big(\argmax\nolimits_{j\le m}U_{k_T,j}=i\big)\Big|\ \le\ \Pp\big(e^{\cI(T)}\mathrm{gap}(Y_T)\le c_1\big)+\frac{C_m}{c_1^2}+o(1).
\end{equation}
The limit $Y$ has a nondegenerate normal law on $E_m$ because $\sigma_m>0$, so $\mathrm{gap}(Y)>0$ almost surely. Let $\eta\to0$, then $T\to\infty$ and then $c_1\to\infty$. Since $\cI(T)\to\infty$ and $Y_T\Rightarrow Y$, the right side of \eqref{eq:mismatch} vanishes, and $W$ converges in law to the block maximizer of $Y$. Subtracting the block average of $Z$ does not change the maximizer, which gives \eqref{eq:blocklaw}. Steps 2 and 3 also show that no trailing answer wins and that a decision occurs.
\end{proof}

\begin{proof}[Proof of Theorem~\ref{thm:weak}]
Take $m=2$ and the start $z^0+\frac12\beta\sqrt\eta\,e$, so that the vector in \eqref{eq:weakchain} is $\frac12\beta e$. Then $H_2=\frac12ee^\top$. At a tie $\mu$ is even under the swap, so $e^\top\mu=0$ and \eqref{eq:decomp} gives $\mu=\sum_{a\ge3}\mu_a\tilde e_a$. Hence $e^\top A\mu=\sum_{a\ge3}\mu_a\,\tilde e_a^\top Ae=\chih\cdot\mu_\perp$, and the difference of the first two coordinates of $\frac12\beta e+b_A+\sigma_2Z$ is
\[
\beta+e^\top b_A+\sigma_2(Z_1-Z_2)=\beta+\beta_A+\sigma_2(Z_1-Z_2).
\]
Finally $2\sigma_2^2=\Sig^2$ by \eqref{eq:bA}, and $\Pp(W=2)\to\Pp(\sigma_2(Z_2-Z_1)>\beta+\beta_A)=\Phic((\beta+\beta_A)/\Sig)$.
\end{proof}

\subsection{The Edge Chart and a Local Separatrix}\label{app:chart}

Besides (P1) to (P3) we use the following facts.
\begin{enumerate}[label=(V\arabic*),leftmargin=*,itemsep=1pt]
\item $\mu$ is a polynomial in $p$ with $\one^\top\mu=0$, and it commutes with every permutation of the answers.
\item For every answer $a$, $\mu_a/p_a$ is a polynomial in $p$, because the update of an answer that is not drawn vanishes.
\item Lemma~\ref{lem:leader} holds.
\item At the edge $p^\circ=(\frac12,\frac12,0,\dots,0)$ one has $\mu_a/p_a=-\gamma^\circ$ for every $a\ge3$, where $\gamma^\circ=\E[\omega(\tilde f)\tilde f]$ for $G\ge4$, with $\tilde f=\max(\tilde N_1,\tilde N_2)/G$ for the counts $(\tilde N_1,\tilde N_2)$ of $G-1$ votes between two answers of probability $\frac12$, and $\gamma^\circ=\frac13\omega(\frac23)$ for $G=3$.
\item The covariance of $\xi$ is a polynomial in $p$, and $\Var(\xi_a)\le Cp_a$.
\end{enumerate}
For (V4), answer $a$ is drawn exactly once with probability $Gp_a+O(p_a^2)$. For $G\ge4$, one of the two leaders then has at least two of the other $G-1$ votes, so $a$ does not lead and $\xi_a=-\omega(\tilde f)\tilde f/G$, which gives $\mu_a=-p_a\E[\omega(\tilde f)\tilde f]+O(p_a^2)$. For $G=3$, the other two votes agree with probability $\frac12$, and then $\xi_a=-\frac29\omega(\frac23)$. If they differ, the three counts are equal and the uniform tie break gives $\E\xi_a=0$. Hence $\mu_a=-\frac13\omega(\frac23)p_a+O(p_a^2)$. For (V5), $\xi_a=0$ unless $a$ is drawn, and $|\xi_a|\le C_\xi$.

By (V1) and (V2), $\mu(p^\circ)=0$. By (V3), the rate of two answers at the tie,
\begin{equation}\label{eq:rcirc}
r^\circ=\partial_\zeta(\mu_1-\mu_2)\big|_{p^\circ}\ \ge\ c_G(G-2)2^{-G}\ >\ 0 ,
\end{equation}
by Lemma~\ref{lem:tied}(iii). A point $z$ is tied if $z_1=z_2$. We fix the threshold $1-\delta\in(\frac12,1)$. An orbit of the mean flow decides for $i$ if $i$ is the first answer whose probability reaches $1-\delta$, and it does not decide if no probability reaches the threshold.

\begin{lemma}[The edge chart]\label{lem:chart}
Put $\zeta=z_1-z_2$, $x_a=p_a(z)$ for $a\ge3$, $\rho=\sum_ax_a$ and $y_a=z_a-\frac12(z_1+z_2)$. The map $z\mapsto(\zeta,x)$ identifies the policies with $\{(\zeta,x)\mid x_a>0,\ \rho<1\}$, with
\[
p_{1,2}=\frac{(1-\rho)\,e^{\pm\zeta/2}}{2\cosh(\zeta/2)} .
\]
In this chart $f_K$ becomes a vector field $V_K=(V^\zeta,V^x)$ that is real analytic near $(0,0)$ and depends on $K$ only through $\chi$ and $Q$, polynomially. Along its orbits, with $\bar d=\mu_1-\mu_2$ and $\mu_\perp=(\mu_a)_{a\ge3}$,
\begin{align}
\zeta'&=\bar d+\chi\cdot\mu_\perp,\label{eq:chartgap}\\
y_a'&=\tfrac12\bar d\,\chi_a+\big(Q\mu_\perp\big)_a,\label{eq:charty}\\
x_a'&=x_a\Big(y_a'-\sum_bx_by_b'-\tfrac12(p_1-p_2)\,\zeta'\Big).\label{eq:chartx}
\end{align}
Moreover $\bar d=\zeta\,g(\zeta,x)$ and $\mu_a=-x_a\gamma_a(\zeta,x)$ for $a\ge3$, with $g$ and $\gamma_a$ analytic and even in $\zeta$, $g(0,0)=r^\circ$ and $\gamma_a(0,0)=\gamma^\circ$. The value $g(0,x)$ is the rate $r$ at the tied point with trailing probabilities $x$. The point $(0,0)$ is an equilibrium for every $K$, every face $\{x_a=0\}$ is invariant, and
\begin{equation}\label{eq:DV}
DV_K(0,0)=\begin{pmatrix}r^\circ&-\gamma^\circ\chi^\top\\0&0\end{pmatrix}.
\end{equation}
\end{lemma}

\begin{proof}
The flow commutes with the shifts $z\mapsto z+c\one$, so it descends to the chart. The policy is a polynomial in $(e^{\pm\zeta/2},x)$ divided by $\cosh(\zeta/2)$, so $\mu$ and $V_K$ are analytic near $(0,0)$, also for $x$ with negative entries. By \eqref{eq:decomp}, $\mu=\frac12\bar d\,e+\sum_b\mu_b\tilde e_b$. Hence
\[
\zeta'=\ehat^{\,\top}\mu=\tfrac12\bar d\,\ehat^{\,\top}e+\sum_b\mu_b\,\chi_b=\bar d+\chi\cdot\mu_\perp ,
\]
which is \eqref{eq:chartgap}, and $y_a'=\tilde e_a^\top K\mu/\kappa$ gives \eqref{eq:charty}. For \eqref{eq:chartx}, $p_a'=p_a(z_a'-p^\top z')$. Writing $z_{1,2}'=u'\pm\frac12\zeta'$ and $z_b'=u'+y_b'$ with $u=\frac12(z_1+z_2)$ gives $z_a'-p^\top z'=y_a'-\sum_bp_by_b'-\frac12(p_1-p_2)\zeta'$. So $V_K$ is a polynomial in $(\chi,Q)$ with analytic coefficients.

The swap of the answers $1$ and $2$ acts on the chart as $(\zeta,x)\mapsto(-\zeta,x)$, and $\mu$ commutes with it by (V1). So $\bar d$ is odd and every $\mu_a$ with $a\ge3$ is even in $\zeta$. An odd analytic function of $\zeta$ is $\zeta$ times an even analytic function, so $\bar d=\zeta g$ with $g$ even and $g(0,x)=\partial_\zeta\bar d(0,x)$. By (V2), $\gamma_a=-\mu_a/x_a$ is a polynomial in $p$, so it is analytic, and it is even. The values at $(0,0)$ are \eqref{eq:rcirc} and (V4). Since $\mu(p^\circ)=0$, $V_K(0,0)=0$, and the faces are invariant by \eqref{eq:chartx}. At $(0,0)$, $\partial_\zeta V^\zeta=g(0,0)+\chi\cdot\partial_\zeta\mu_\perp=r^\circ$, $\partial_{x_a}V^\zeta=\partial_{x_a}\bar d-\gamma^\circ\chi_a=-\gamma^\circ\chi_a$ because $\bar d$ vanishes on $\{\zeta=0\}$, and $DV^x=0$ because every $V^x_a$ is $x_a$ times a function that vanishes at $(0,0)$.
\end{proof}

The next lemma is the geometric core. The edge has one unstable direction, the gap, and a center manifold. The center manifold of the flow suspended over the kernel gives a local separatrix that depends smoothly on the kernel.

\begin{lemma}[A local separatrix]\label{lem:local}
Let $\Theta\subset\R^{n_\theta}$ be open and let $\theta\mapsto K_\theta$ be a $C^k$ map, $k\ge3$, into the symmetric matrices with $K_\theta\one=0$ and $\kappa_\theta>0$. Write $V_\theta$, $\chi_\theta$, $Q_\theta$ and $S_\theta$ for the objects of $K_\theta$, and let $\theta_*\in\Theta$ have $S_{\theta_*}$ strictly copositive. There are, depending on $k$, a compact neighborhood $\Theta_0$ of $\theta_*$, numbers $\varepsilon_4>\varepsilon_5\ge\varepsilon_6>0$, constants $C,c>0$ and a $C^k$ function $h$ on $\{|x|<\varepsilon_4\}\times\Theta_0$ with the following properties for every $\theta\in\Theta_0$. Put
\[
\cN=\{(\zeta,x)\mid |\zeta|<\varepsilon_4,\ x\ge0,\ \rho<\varepsilon_5\},\qquad F_\theta=\zeta-h(x,\theta).
\]
\begin{enumerate}[label=(\alph*),leftmargin=*,itemsep=1pt]
\item $h(0,\theta)=0$, $\partial_xh(0,\theta)=(\gamma^\circ/r^\circ)\chi_\theta^\top$, $|h(x,\theta)|\le C|x|$ and $|\partial_xh|+|\partial_\theta h|\le C$. Moreover $g\ge r^\circ/2$ on $\cN$.
\item Along every orbit of $V_\theta$, while it lies in $\cN$,
\begin{equation}\label{eq:normal}
F_\theta'=\alpha_\theta F_\theta,\qquad \rho'\ \le\ -c\rho^2+C\rho\,|F_\theta|,
\end{equation}
with a continuous function $\alpha_\theta\ge r^\circ/2$ of the state.
\item If $(\zeta,x)\in\cN$ and $F_\theta(\zeta,x)=0$, the orbit stays in $\cN$, converges to $(0,0)$ and does not decide.
\item If $(\zeta,x)\in\cN$ has $|\zeta|<\varepsilon_6$ and $\rho<\varepsilon_6$, the orbit decides for answer $1$ when $F_\theta(\zeta,x)>0$ and for answer $2$ when $F_\theta(\zeta,x)<0$.
\item If $\chi_\theta=0$, then $h(x,\theta)=0$ for every $x\ge0$ with $\rho<\varepsilon_6$.
\end{enumerate}
\end{lemma}

\begin{proof}
\emph{Step 1. Suspension.} Consider the vector field $\tilde V(\zeta,x,\theta)=(V_\theta(\zeta,x),0)$ near $(0,0,\theta_*)$. It is $C^k$, because $V_\theta$ is a polynomial in $(\chi_\theta,Q_\theta)$ with analytic coefficients. The points $(0,0,\theta)$ are equilibria, so $\partial_\theta V_\theta(0,0)=0$, and by \eqref{eq:DV}
\[
D\tilde V(0,0,\theta)=\begin{pmatrix}r^\circ&-\gamma^\circ\chi_\theta^\top&0\\0&0&0\\0&0&0\end{pmatrix}.
\]
At $\theta_*$ the spectrum consists of the simple eigenvalue $r^\circ$ and the eigenvalue $0$, whose generalized eigenspace is the kernel $E^{\rm c}=\{r^\circ\zeta=\gamma^\circ\chi_{\theta_*}\cdot x\}$. There is no stable direction, and $E^{\rm c}$ is a graph over the coordinates $(x,\theta)$. By the center manifold theorem \citep{kelley1967stable,carr1981applications,vanderbauwhede1989centre} there are a neighborhood $\mathcal U$ of $(0,0,\theta_*)$, which depends on $k$, and a $C^k$ function $h(x,\theta)$ near $(0,\theta_*)$ such that the graph $\{\zeta=h(x,\theta)\}$ is locally invariant under $\tilde V$, is tangent to $E^{\rm c}$ at $(0,0,\theta_*)$, and contains every equilibrium of $\tilde V$ in $\mathcal U$. Since $\theta$ is constant along orbits, the graph of $h(\cdot,\theta)$ has the same properties for $V_\theta$, and in particular $h(0,\theta)=0$. The graph is locally invariant and contains the equilibrium $(0,0,\theta)$, so its tangent space there is invariant under $D\tilde V(0,0,\theta)$. The image of the tangent vector $(\partial_xh\,x+\partial_\theta h\,u,x,u)$ is
\[
\Big(r^\circ\big(\partial_xh\,x+\partial_\theta h\,u\big)-\gamma^\circ\chi_\theta\cdot x,\ 0,\ 0\Big),
\]
which lies in the tangent space only if $\partial_xh(0,\theta)=(\gamma^\circ/r^\circ)\chi_\theta^\top$. Let $c_2$ bound $|\partial_xh|$ and $|\partial_\theta h|$ on a compact neighborhood of $(0,\theta_*)$. We take $\varepsilon_4$ and $\Theta_0$ so small that $\{|\zeta|<(c_2+2)\varepsilon_4,|x|<\varepsilon_4\}\times\Theta_0\subset\mathcal U$. Then the graph points with $|x|<\varepsilon_4$ lie in $\mathcal U$, since $|h|\le c_2|x|$, and so do the segments that join them to the points with $|\zeta|<\varepsilon_4$. This gives the bounds of (a) with $C=c_2$. We also take $\varepsilon_4$ so small that $g\ge r^\circ/2$ on $\{|\zeta|<(c_2+2)\varepsilon_4,|x|<\varepsilon_4\}$, that $\frac12(1+\tanh(\varepsilon_4/2))<1-\delta$ and that $\varepsilon_4<1-\delta$. Then no probability reaches the threshold at a point of $\cN$.

\emph{Step 2. The normal coordinate.} The derivative of $F_\theta$ along $V_\theta$ is $DF_\theta V_\theta=V^\zeta_\theta-\partial_xh\,V^x_\theta$, and local invariance means that it vanishes at $(h(x,\theta),x)$. Hence
\begin{equation}\label{eq:alpha}
DF_\theta V_\theta=\alpha_\theta F_\theta,\qquad \alpha_\theta(\zeta,x)=\int_0^1\partial_\zeta(DF_\theta V_\theta)\big(h+sF_\theta,x\big)\,ds ,
\end{equation}
which is continuous in $(\zeta,x,\theta)$. At $(0,0)$, $\partial_\zeta(DF_\theta V_\theta)=r^\circ-\partial_xh\,\partial_\zeta V^x_\theta=r^\circ$, since $DV^x=0$ there. Shrinking $\varepsilon_4$ and $\Theta_0$ gives $\alpha_\theta\ge r^\circ/2$ on $\{|\zeta|<\varepsilon_4,|x|<\varepsilon_4\}\times\Theta_0$.

\emph{Step 3. The trailing mass.} Summing \eqref{eq:chartx} gives
\begin{equation}\label{eq:rhoprime}
\rho'=(1-\rho)\sum_ax_ay_a'-\tfrac12\rho\,(p_1-p_2)\,\zeta' .
\end{equation}
Write $\zeta=h+F$ with $F=F_\theta$. By Lemma~\ref{lem:chart}, $\frac12\bar d\chi_a=\frac12g\,(h+F)\chi_a$ with $g=r^\circ+O(\zeta^2+|x|)$ and $h=(\gamma^\circ/r^\circ)\chi_\theta\cdot x+O(|x|^2)$, and $(Q\mu_\perp)_a=-\gamma^\circ(Qx)_a+O(|x|^2+|x|\zeta^2)$. So \eqref{eq:charty} gives
\begin{equation}\label{eq:ySx}
y_a'=-\gamma^\circ\big(S_\theta x\big)_a+O\big(|x|^2+|F|\big)
\end{equation}
uniformly on $\cN\times\Theta_0$. Since $x\ge0$ and $|x|\le\rho$, $\sum_ax_ay_a'=-\gamma^\circ x^\top S_\theta x+O(\rho^3+\rho|F|)$. Also $|p_1-p_2|\le|\zeta|$ and $|\zeta'|\le C(|\zeta|+\rho)$, so $\rho|p_1-p_2||\zeta'|\le C\rho(\rho+|F|)^2\le C(\rho^3+\rho|F|)$. The minimum $c_\theta$ of $x^\top S_\theta x$ over $\{x\ge0,\ \sum_ax_a=1\}$ is continuous and positive at $\theta_*$. Shrinking $\Theta_0$, $x^\top S_\theta x\ge\frac12c_{\theta_*}\rho^2$ for $x\ge0$, and therefore
\[
\rho'\ \le\ -\tfrac12\gamma^\circ c_{\theta_*}\,\rho^2+C\rho^3+C\rho\,|F|\ \le\ -c\rho^2+C\rho\,|F|
\]
once $\varepsilon_5$ is small. We also take $C\varepsilon_5<\varepsilon_4/2$.

\emph{Step 4. Points on the graph.} If $F_\theta=0$ at a point of $\cN$, then $F_\theta\equiv0$ along the orbit while it is in $\cN$, by Step 2. Then $\rho'\le-c\rho^2$, so $\rho(t)\le\rho(0)/(1+c\rho(0)t)$, and $|\zeta|=|h|\le C\rho<\varepsilon_4/2$. The faces are invariant, so $x\ge0$ persists. Hence the orbit never reaches the boundary of $\cN$, stays in $\cN$ and converges to $(0,0)$. By Step 1 it does not decide.

\emph{Step 5. Points off the graph.} Put $\alpha_0=r^\circ/2$ and fix $\varepsilon_7\in(0,\varepsilon_4/4)$. Take a point of $\cN$ with $0<|F|<\varepsilon_7$ and $\rho<\varepsilon_8$, with $\varepsilon_8$ chosen below, and let $\tau$ be the exit time of its orbit from $\{|F|<\varepsilon_7\}\cap\cN$. On $[0,\tau)$,
\begin{equation}\label{eq:Fgrowth}
|F(t)|\ \ge\ |F(0)|\,e^{\alpha_0t},\qquad |F(t)|\ \le\ \varepsilon_7\,e^{-\alpha_0(\tau-t)},
\end{equation}
so $\tau<\infty$. By Step 3, $(\log\rho)'\le C|F|$, so \eqref{eq:Fgrowth} gives $\rho(t)\le\varepsilon_8e^{C\varepsilon_7/\alpha_0}=:\varepsilon_9$ on $[0,\tau]$. If $\varepsilon_9$ is small, then $|\zeta|\le\varepsilon_7+C\varepsilon_9<\varepsilon_4$ and $\rho\le\varepsilon_9<\varepsilon_5$ on $[0,\tau]$, and $x\ge0$ by the invariance of the faces. So the orbit does not leave $\cN$ at $\tau$. It leaves $\{|F|<\varepsilon_7\}$ there, and it does not decide before $\tau$. At $\tau$, $F=\pm\varepsilon_7$ with the sign of $F(0)$, $\zeta=\pm\varepsilon_7+O(\varepsilon_9)$ and $\rho\le\varepsilon_9$.

On the face $\{x=0\}$ the field is $\zeta'=\bar d(\zeta,0)$ by \eqref{eq:chartgap}, the same for all $\theta$, and $\bar d(\zeta,0)>0$ for $\zeta>0$ by (V3). So the face orbit from $\zeta=\varepsilon_7$ increases to $+\infty$. Its $p_1$ crosses $1-\delta$ at a finite time $t_1$ with $p_1'>0$, while $p_2<\frac12$. The field is continuous in $(\zeta,x,\theta)$ on the compact set $\{|\zeta|\le c_3,|x|\le\varepsilon_4\}\times\Theta_0$, where $c_3$ exceeds the gap of the face orbit at $t_1+1$. By continuous dependence there is $\varepsilon_{10}>0$ such that every orbit of every $V_\theta$, $\theta\in\Theta_0$, that starts with $|\zeta-\varepsilon_7|\le\varepsilon_{10}$ and $|x|\le\varepsilon_{10}$ crosses $p_1=1-\delta$ before $t_1+1$, while all other probabilities stay below $1-\delta$ up to that time. Choose $\varepsilon_8$ with $\varepsilon_9\le\varepsilon_{10}$ and $C\varepsilon_9\le\varepsilon_{10}$. Then the orbit decides for answer $1$ if $F(0)>0$. The case $F(0)<0$ follows in the same way from the face orbit from $-\varepsilon_7$, the mirror image of the first. Finally choose $\varepsilon_6\le\min(\varepsilon_8,\varepsilon_5)$ with $\varepsilon_6+C\varepsilon_6<\varepsilon_7$. A point of $\cN$ with $|\zeta|<\varepsilon_6$ and $\rho<\varepsilon_6$ has $|F|<\varepsilon_7$, which proves (d).

\emph{Step 6. A symmetric kernel.} If $\chi_\theta=0$, then $\zeta'=\zeta g$ by \eqref{eq:chartgap}, so $\{\zeta=0\}$ is invariant. On it $y_a'=(Q_\theta\mu_\perp)_a$ and $\rho'=-\gamma^\circ x^\top Q_\theta x+O(\rho^3)<0$, because $Q_\theta=S_\theta$ is strictly copositive. Shrinking $\varepsilon_6$ if necessary, the orbit of $(0,x)$ with $x\ge0$ and $\rho<\varepsilon_6$ converges to $(0,0)$ without deciding. By (d), $F_\theta(0,x)=0$, that is, $h(x,\theta)=0$.
\end{proof}

On $\cN_1=\{(\zeta,x)\in\cN\mid|\zeta|<\varepsilon_6,\ \rho<\varepsilon_6\}$ the graph is unique, because by (c) and (d) it is the set of points of $\cN_1$ whose orbits do not decide. A center manifold need not be unique outside $\cN_1$, but no statement below depends on that part.

\subsection{The Separatrix Near the Edge}\label{app:edge}

\begin{lemma}[Tied paths near the edge]\label{lem:tiededge}
Let $Q$ be strictly copositive. There are $\varepsilon_{11}>0$, $c>0$ and $C$ such that the following holds for every tied $z^0$ with $\rho_0=\rho(z^0)<\varepsilon_{11}$ and $x^0=x(z^0)$. The tied path $\bar z$ of $K$ from $z^0$ has the chart point $(0,\bar x(t))$, its trailing mass satisfies $\rho(t)\le\rho_0/(1+c\rho_0t)$, and $|\log(\bar x_a(t)/x^0_a)|\le C\rho_0t$. Let $r(t)=g(0,\bar x(t))$, $\cI(t)=\int_0^tr$, let $v$ be the variance of $d$ along the path and $v^\circ$ its value at the edge. Then for every $u\in\R^{n-2}$
\begin{align}
\int_0^\infty e^{-\cI}\,u\cdot\mu_\perp(\bar z)\,dt&=-\frac{\gamma^\circ}{r^\circ}\,u\cdot x^0+O\Big(\rho_0\sum_a|u_a|x^0_a\Big),\label{eq:edgeint1}\\
\int_0^\infty e^{-2\cI}\big(v+u^\top\Cov(\xi_\perp)u\big)\,dt&=\frac{v^\circ}{2r^\circ}+O\big(\rho_0(1+|u|^2)\big).\label{eq:edgeint2}
\end{align}
The constants depend on $K$ only through bounds on $|\chi|$ and $|Q|$ and a lower bound on the copositivity constant of $Q$.
\end{lemma}

\begin{proof}
The tied path keeps $\bar z_1=\bar z_2$ because $e^\top(\Id-\frac12ee^\top)=0$. On the tied hyperplane $\bar d=0$ and $(\Id-\frac12ee^\top)\tilde e_a=\tilde e_a$, so $y_a'=\tilde e_a^\top K\mu/\kappa=(Q\mu_\perp)_a$ by \eqref{eq:decomp}. With $\mu_b=-x_b\gamma_b(0,x)$ and $\gamma_b(0,x)=\gamma^\circ+O(|x|)$, this is
\[
y_a'=-\gamma^\circ(Qx)_a+O(\rho^2),\qquad |y_a'|\le C\rho .
\]
By \eqref{eq:chartx} with $\zeta=0$, $(\log\bar x_a)'=y_a'-\sum_b\bar x_by_b'$, so $|(\log\bar x_a)'|\le C\rho$, and
\[
\rho'=(1-\rho)\sum_a\bar x_ay_a'=-\gamma^\circ\,\bar x^\top Q\bar x+O(\rho^3)\ \le\ -c\rho^2
\]
for $\rho<\varepsilon_{11}$. This gives the bounds on $\rho$ and on $\log(\bar x_a/x^0_a)$. Now
\[
u\cdot\mu_\perp(t)=-\sum_au_a\bar x_a\gamma_a(0,\bar x)=-\gamma^\circ\,u\cdot x^0+R(t),\qquad |R(t)|\le C\sum_a|u_a|x^0_a\,\rho_0(1+t)e^{C\rho_0t},
\]
because $|\bar x_a/x^0_a-1|\le C\rho_0te^{C\rho_0t}$ and $\gamma_a(0,\bar x)-\gamma^\circ=O(\rho_0)$. Also $r(t)=r^\circ+O(\rho_0)$, so $|\cI(t)-r^\circ t|\le C\rho_0t$ and $|e^{-\cI(t)}-e^{-r^\circ t}|\le C\rho_0te^{-(r^\circ-C\rho_0)t}$. If $C\varepsilon_{11}\le r^\circ/4$, integration against $e^{-r^\circ t/2}$ gives \eqref{eq:edgeint1}, since $\int_0^\infty e^{-r^\circ t}dt=1/r^\circ$. For \eqref{eq:edgeint2}, $v(t)=v^\circ+O(\rho_0)$ and $|u^\top\Cov(\xi_\perp)u|\le|u|^2\sum_a\Var(\xi_a)\le C|u|^2\rho_0$ by (V5).
\end{proof}

Near the edge, therefore, the whole effect of the kernel beyond $\kappa$ is carried by one number per question, the spillover index $\sum_a\chi_ap_a(z^0)$. Applied to the weak anisotropy of Theorem~\ref{thm:weak}, whose tied path is the tabular one with $Q=\Id+\frac12\one\one^\top$, Lemma~\ref{lem:tiededge} with $u=\chih$ gives
\begin{equation}\label{eq:edgeform}
\beta_A=-\frac{\gamma^\circ}{r^\circ}\sum_{a\ge3}\chih_a\,p_a(z^0)+O\Big(\rho_0\sum_{a\ge3}|\chih_a|\,p_a(z^0)\Big),\qquad \Sig=\Sig^\circ+O(\rho_0),
\end{equation}
with $\Sig^\circ=(v^\circ/(2r^\circ))^{1/2}$. Table~\ref{tab:edgeconst} lists the constants. The edge form overstates the spillover at the trailing masses of our experiments, because the trailing answers lose mass along the path. We use it as a link to the tabular law and not as a predictor.

\begin{table}[ht]
\caption{Constants of two answers at a tie. The last column is the shift of the probit per unit of the spillover index at $\eta=1$. The rationals for the mean baseline are, for instance, $r^\circ=105/256$ and $v^\circ=203/1024$ at $G=8$. The standard-deviation baseline uses the unbiased standard deviation of the group rewards.}
\label{tab:edgeconst}
\centering\small
\begin{tabular}{@{}lrrrrrrr@{}}
\toprule
baseline & $G$ & $r^\circ$ & $v^\circ$ & $\Sig^\circ$ & $\gamma^\circ$ & $\gamma^\circ/r^\circ$ & $\gamma^\circ/(r^\circ\Sig^\circ)$\\
\midrule
mean & 4 & 0.1875 & 0.1641 & 0.6614 & 0.5625 & 3.000 & 4.536\\
mean & 8 & 0.4102 & 0.1982 & 0.4916 & 0.5742 & 1.400 & 2.848\\
mean & 16 & 0.6873 & 0.2216 & 0.4015 & 0.5669 & 0.825 & 2.055\\
mean & 32 & 1.0496 & 0.2351 & 0.3346 & 0.5543 & 0.528 & 1.578\\
std & 4 & 0.3750 & 0.5625 & 0.8660 & 1.0245 & 2.732 & 3.155\\
std & 8 & 0.8667 & 0.7656 & 0.6646 & 1.1237 & 1.297 & 1.951\\
std & 16 & 1.4191 & 0.8789 & 0.5565 & 1.1265 & 0.794 & 1.426\\
std & 32 & 2.1328 & 0.9385 & 0.4691 & 1.1062 & 0.519 & 1.106\\
\bottomrule
\end{tabular}
\end{table}

\begin{proposition}[The separatrix near the edge]\label{prop:sepedge}
Let $G\ge3$ and let $S$ be strictly copositive, for instance $K$ positive definite on $\one^\perp$.
\begin{enumerate}[label=(\roman*),leftmargin=*,itemsep=1pt]
\item The edge is an equilibrium of $V_K$ with the linearization \eqref{eq:DV}. It has the eigenvalue $r^\circ$ in the direction of the gap and the eigenvalue $0$ on $E^{\rm c}=\{r^\circ\zeta=\gamma^\circ\chi\cdot x\}$.
\item For each finite $k$ there are a neighborhood $\cN_1$ of the edge in $\{x\ge0\}$ and a $C^k$ function $h$ with $h(0)=0$ and $Dh(0)=(\gamma^\circ/r^\circ)\chi^\top$ such that a point of $\cN_1$ does not decide if and only if $\zeta=h(x)$, and then it flows to the edge. The function $h$ depends $C^k$ on $K$ near every kernel with $\kappa>0$ and $S$ strictly copositive.
\item There are $\varepsilon_{12},\varepsilon_{13}>0$ such that for every tied $z^0$ with $\rho_0=\rho(z^0)<\varepsilon_{12}$ there is exactly one $\zeta^*\in(-\varepsilon_{13},\varepsilon_{13})$ for which the flow from $z^0+\frac12\zeta^*e$ does not decide. The flow from $z^0+\frac12\zeta_0e$ decides for answer $1$ if $\zeta^*<\zeta_0<\varepsilon_{13}$ and for answer $2$ if $-\varepsilon_{13}<\zeta_0<\zeta^*$. With $B_{\rm edge}=-(\gamma^\circ/r^\circ)\sum_{a\ge3}\chi_ap_a(z^0)$,
\begin{equation}\label{eq:sepedge}
\zeta^*=-B_{\rm edge}+O(\rho_0^2),\qquad B_K=B_{\rm edge}+O(\rho_0^2),\qquad \Sig_K=\Sig^\circ+O(\rho_0),
\end{equation}
and hence $\zeta^*=-B_K+O(\rho_0^2)$.
\end{enumerate}
\end{proposition}

\begin{proof}
(i) is Lemma~\ref{lem:chart}. (ii) is Lemma~\ref{lem:local} with $\theta$ the kernel itself, which ranges over an open subset of the symmetric centered matrices, together with the uniqueness of the graph on $\cN_1$.

(iii) The line $z^0+\frac12\zeta_0e$ has the chart points $(\zeta_0,x(\zeta_0))$ with
\[
x_a(\zeta_0)=\frac{e^{y_a}}{2\cosh(\zeta_0/2)+\sum_be^{y_b}},\qquad y_a=z^0_a-z^0_1 .
\]
So $0\le x^0_a-x_a(\zeta_0)\le x^0_a(\cosh(\zeta_0/2)-1)\le x^0_a\zeta_0^2$ for $|\zeta_0|\le1$, and $|\partial_{\zeta_0}x_a(\zeta_0)|\le x^0_a|\zeta_0|$. Let $\hat F(\zeta_0)=F(\zeta_0,x(\zeta_0))=\zeta_0-h(x(\zeta_0))$ with $h$, $\varepsilon_6$ and $C$ from Lemma~\ref{lem:local}. Then
\begin{equation}\label{eq:gammac}
\hat F'(\zeta_0)\ \ge\ 1-C\rho_0|\zeta_0|,\qquad |\hat F(\zeta_0)-\zeta_0|\ \le\ C\rho_0 .
\end{equation}
Choose $\varepsilon_{13}\le\varepsilon_6/2$ and $\varepsilon_{12}\le\min(\varepsilon_6,\varepsilon_{11})$, with $\varepsilon_{11}$ of Lemma~\ref{lem:tiededge}, such that $C\varepsilon_{12}<\varepsilon_{13}/2$ and $C\varepsilon_{12}\varepsilon_{13}\le\frac12$. Then $\hat F(-\varepsilon_{13})<0<\hat F(\varepsilon_{13})$ and $\hat F'\ge\frac12$ on $(-\varepsilon_{13},\varepsilon_{13})$, so $\hat F$ has exactly one zero $\zeta^*$ there. By Lemma~\ref{lem:local}(c) the flow from $z^0+\frac12\zeta^*e$ does not decide, and by Lemma~\ref{lem:local}(d) every other $\zeta_0\in(-\varepsilon_{13},\varepsilon_{13})$ decides for answer $1$ if $\zeta_0>\zeta^*$ and for answer $2$ if $\zeta_0<\zeta^*$, since its chart point has $|\zeta|<\varepsilon_6$ and $\rho\le\rho_0<\varepsilon_6$.

Since $|\zeta^*|=|h(x(\zeta^*))|\le C\rho_0$, one has $x(\zeta^*)=x^0+O(\rho_0^3)$, and therefore
\[
\zeta^*=h(x^0)+O(\rho_0^3)=\frac{\gamma^\circ}{r^\circ}\chi\cdot x^0+O(\rho_0^2)=-B_{\rm edge}+O(\rho_0^2).
\]
The matrix $Q\succeq S$ is strictly copositive, so Lemma~\ref{lem:tiededge} with $u=\chi$ gives $B_K=\int_0^\infty e^{-\cI}\chi\cdot\mu_\perp(\bar z)\,dt=B_{\rm edge}+O(\rho_0^2)$, and with $u=\chi$ in \eqref{eq:edgeint2} it gives $\Sig_K^2=v^\circ/(2r^\circ)+O(\rho_0)$.
\end{proof}

\begin{theorem}[The loop term]\label{thm:second}
In the setting of Proposition~\ref{prop:sepedge}(iii),
\begin{equation}\label{eq:loopterm}
\zeta^*+B_K=\frac{\gamma^\circ}{2r^\circ}\,\zeta^*\sum_{a\ge3}\chi_a^2\,p_a(z^0)+O(\rho_0^3),
\end{equation}
or equivalently $\zeta^*=-B_K\big(1+\frac{\gamma^\circ}{2r^\circ}\sum_{a\ge3}\chi_a^2p_a(z^0)\big)+O(\rho_0^3)$.
\end{theorem}

The tied path holds the gap at zero. On the separatrix the gap stays near $\zeta^*$ during the window, and by \eqref{eq:charty} the kernel then moves each trailing log-odds at the extra rate $\frac12r^\circ\chi_a\zeta^*$. The moved answers change the spillover by $-\gamma^\circ\chi\cdot(x-\bar x)$. This loop is the only second-order difference between $\zeta^*$ and $-B_K$, because the other second-order terms of the field enter $\zeta^*$ and $-B_K$ in the same way.

\begin{proof}
\emph{Step 1. Exact representations.} Let $(\zeta(t),x(t))$ be the orbit from $z^0+\frac12\zeta^*e$. It stays in $\cN$, satisfies $\zeta=h(x)$, converges to the edge, and $\rho(t)\le\rho_0$. By \eqref{eq:chartgap}, $\zeta'=g(\zeta,x)\zeta+\psi(\zeta,x)$ with $\psi=\chi\cdot\mu_\perp$. Put $\cI_g(t)=\int_0^tg(\zeta,x)$, which is at least $r^\circ t/2$ by Lemma~\ref{lem:local}(a). Then $(e^{-\cI_g}\zeta)'=e^{-\cI_g}\psi$ and $e^{-\cI_g(t)}\zeta(t)\to0$, so
\begin{equation}\label{eq:cstarrep}
\zeta^*=\zeta(0)=-\int_0^\infty e^{-\cI_g(t)}\,\psi\big(\zeta(t),x(t)\big)\,dt .
\end{equation}
Along the tied path, $B_K=\int_0^\infty e^{-\cI(t)}\psi(0,\bar x(t))\,dt$ with $\cI(t)=\int_0^tg(0,\bar x)$. Hence
\begin{equation}\label{eq:diff}
\zeta^*+B_K=-\int_0^\infty\Big(e^{-\cI_g}\psi(\zeta,x)-e^{-\cI}\psi(0,\bar x)\Big)\,dt .
\end{equation}

\emph{Step 2. Bounds on the separatrix orbit.} Since $\zeta=h(x)$, $|\zeta|\le C\rho$ and $|\zeta^*|\le C\rho_0$. By \eqref{eq:chartx}, $|x'|\le C\rho^2$, so $|\zeta'|=|\partial_xh\,x'|\le C\rho^2$ and $\zeta(t)=\zeta^*+O(\rho_0^2t)$. The functions $\gamma_a$ and $g$ are even in $\zeta$, so
\begin{equation}\label{eq:even}
\big|\mu_\perp(\zeta,x)-\mu_\perp(0,x)\big|\ \le\ C\rho\zeta^2\ \le\ C\rho_0^3,\qquad \big|g(\zeta,x)-g(0,x)\big|\ \le\ C\zeta^2 .
\end{equation}

\emph{Step 3. The trailing answers.} Write $x_a'=x_ar_a$ and $\bar x_a'=\bar x_a\bar r_a$, with $r_a$ the bracket of \eqref{eq:chartx} on the separatrix orbit and $\bar r_a$ the bracket on the tied path. By \eqref{eq:charty} and \eqref{eq:even}, $y_a'(\zeta,x)-y_a'(0,x)=\frac12\zeta g(\zeta,x)\chi_a+O(\rho_0^3)=\frac12r^\circ\chi_a\zeta+O(\rho_0^2)$. The sums $\sum_bx_by_b'$ differ by $O(\rho_0^2+\rho_0|x-\bar x|)$, and the last term of \eqref{eq:chartx} is $O(|\zeta||\zeta'|)=O(\rho_0^3)$ on the separatrix orbit and zero on the tied path. With the Lipschitz bound for the bracket at $\zeta=0$,
\begin{equation}\label{eq:Ydiff}
r_a-\bar r_a=\tfrac12r^\circ\chi_a\,\zeta^*+O\big(\rho_0^2(1+t)\big)+O\big(|x-\bar x|\big).
\end{equation}
Let $u_a=\log x_a-\log\bar x_a$. Then $u_a(0)=\log(x_a(\zeta^*)/x^0_a)=O(\rho_0^2)$, and $|x-\bar x|\le C\rho_0|u|$ while $|u|\le1$. For $\tilde u(t)=u(t)-\frac12r^\circ \zeta^*t\,\chi$, \eqref{eq:Ydiff} gives $|\tilde u'|\le C\rho_0^2(1+t)+C\rho_0|\tilde u|$ and $|\tilde u(0)|\le C\rho_0^2$, so Gronwall's lemma gives $|\tilde u(t)|\le C\rho_0^2(1+t)^2e^{C\rho_0t}$. In particular $|u|\le1$ for $t\le\varepsilon_{14}/\rho_0$ with a small $\varepsilon_{14}>0$. Since $\bar x_a(t)=x^0_a+O(\rho_0^2t)$ by Lemma~\ref{lem:tiededge}, for $t\le\varepsilon_{14}/\rho_0$
\begin{equation}\label{eq:dx}
x_a(t)-\bar x_a(t)=\tfrac12r^\circ \zeta^*\,\chi_a\,x^0_a\,t+O\big(\rho_0^3(1+t)^2e^{C\rho_0t}\big).
\end{equation}

\emph{Step 4. Assembly.} Split the integrand of \eqref{eq:diff} as $e^{-\cI}(\psi(\zeta,x)-\psi(0,\bar x))+(e^{-\cI_g}-e^{-\cI})\psi(\zeta,x)$. First, $\psi(\zeta,x)-\psi(0,x)=O(\rho_0^3)$ by \eqref{eq:even}, and $\partial_x\psi(0,x)=-\gamma^\circ\chi^\top+O(|x|)$, so by \eqref{eq:dx}
\begin{equation}\label{eq:psidiff}
\psi(\zeta,x)-\psi(0,\bar x)=-\tfrac12\gamma^\circ r^\circ \zeta^*\,t\sum_a\chi_a^2x^0_a+O\big(\rho_0^3(1+t)^2e^{C\rho_0t}\big).
\end{equation}
Second, by \eqref{eq:even} and \eqref{eq:dx}, $|\cI_g-\cI|\le\int_0^t(C\zeta^2+C|x-\bar x|)\le C\rho_0^2(1+t)^2e^{C\rho_0t}$, and $|\psi|\le C\rho_0$, so the second part is $O(\rho_0^3(1+t)^2e^{-(r^\circ/2-C\rho_0)t})$. For $t>\varepsilon_{14}/\rho_0$ both integrands are at most $C\rho_0e^{-r^\circ t/2}$, whose integral is $O(\rho_0^3)$. Finally $\int_0^\infty te^{-\cI(t)}dt=1/(r^\circ)^2+O(\rho_0)$ by the bound $|\cI(t)-r^\circ t|\le C\rho_0t$. Inserting everything into \eqref{eq:diff} gives
\[
\zeta^*+B_K=\tfrac12\gamma^\circ r^\circ\,\zeta^*\sum_a\chi_a^2x^0_a\cdot\frac1{(r^\circ)^2}+O(\rho_0^3),
\]
which is \eqref{eq:loopterm}. The second form follows from $\zeta^*=-B_K+O(\rho_0^2)$.
\end{proof}

The correction factor exceeds one, so $-B_K$ places the separatrix slightly too close to the tie, by a relative amount of order $\rho_0\max_a\chi_a^2$. The edge offset $-B_{\rm edge}$ also has an error of order $\rho_0^2$, but its coefficient collects every second-order term of the field, and in the checks of Appendix~\ref{app:numerics} it is ten to thirty times larger than that of $-B_K$.

\subsection{The Separatrix Under a Weak Anisotropy}\label{app:aniso}

Let $A$ be symmetric with $A\one=0$ and $e^\top Ae=0$. Let $P_{12}$ be the permutation matrix of the swap of the answers $1$ and $2$, and put $A_{\rm e}=\frac12(A+P_{12}AP_{12})$ and $A_{\rm o}=\frac12(A-P_{12}AP_{12})$. For $\theta=(\te,\tho)\in\R^2$ let
\begin{equation}\label{eq:Ktheta}
K_\theta=\kappa\big(H+\te A_{\rm e}+\tho A_{\rm o}\big),\qquad K_s=K_{(s,s)}=\kappa(H+sA),\qquad f_\theta=f_{K_\theta}=\big(H+\te A_{\rm e}+\tho A_{\rm o}\big)\mu .
\end{equation}
The last identity uses $e^\top K_\theta e=2\kappa$, which holds because $e^\top A_{\rm e}e=e^\top Ae=0$ and $e^\top A_{\rm o}e=0$. For small $|\theta|$ the kernel $K_\theta$ is positive definite on $\one^\perp$. Let $\chih_a=\tilde e_a^\top Ae$. A vector $u$ is even if $P_{12}u=u$, that is $u_1=u_2$, and odd if $P_{12}u=-u$, that is $u\in\R e$. For a tied $z^0$ let $\bar z_{\te}$ be the orbit of $f_{(\te,0)}$ from $z^0$, $\bar x_{\te}$ its trailing part, $r_{\te}=g(0,\bar x_{\te})$, $\cI_{\te}=\int_0^tr_{\te}$, and
\begin{equation}\label{eq:Mte}
\beta_A(\te)=\int_0^\infty e^{-\cI_{\te}(t)}\,\chih\cdot\mu_\perp\big(\bar z_{\te}(t)\big)\,dt .
\end{equation}

\begin{lemma}[Parity]\label{lem:parity}
\begin{enumerate}[label=(\roman*),leftmargin=*,itemsep=1pt]
\item $A_{\rm e}e=0$, $A_{\rm o}e$ is even, and the spillover weights of $K_\theta$ are $\chi=\tho\chih$.
\item $f_{(\te,0)}$ is tangent to the tied hyperplane, and the tied path of $K_\theta$ from a tied $z^0$ is $\bar z_{\te}$ for every $\tho$.
\item $e^\top Df_{(\te,0)}(z)=r\,e^\top$ at every tied $z$, with $r=g(0,x(z))$. Hence the fundamental matrix $M_{\te}$ of the linearization of $f_{(\te,0)}$ along $\bar z_{\te}$ satisfies
\begin{equation}\label{eq:fundamental}
e^\top\,M_{\te}(t,t_0)=e^{\cI_{\te}(t)-\cI_{\te}(t_0)}\,e^\top .
\end{equation}
\item $B_{K_\theta}=\tho\,\beta_A(\te)$ exactly, and $v_{K_\theta}=v+\tho^2\chih^\top\Cov(\xi_\perp)\chih$ along $\bar z_{\te}$.
\item $P_{12}f_{(\te,\tho)}(z)=f_{(\te,-\tho)}(P_{12}z)$ for every $z$.
\end{enumerate}
\end{lemma}

\begin{proof}
(i) $P_{12}A_{\rm e}e=A_{\rm e}P_{12}e=-A_{\rm e}e$, so $A_{\rm e}e$ is a multiple of $e$, and $e^\top A_{\rm e}e=0$ gives $A_{\rm e}e=0$. Similarly $P_{12}A_{\rm o}e=-A_{\rm o}P_{12}e=A_{\rm o}e$. So $\ehat=e+\tho A_{\rm o}e$ and $\chi_a=\tho\,\tilde e_a^\top A_{\rm o}e=\tho\,\tilde e_a^\top Ae=\tho\chih_a$, because $\tilde e_a^\top A_{\rm e}e=0$.

(ii) At a tied $z$, $\mu$ is even by (V1). The matrices $H$ and $A_{\rm e}$ map even vectors to even vectors, and $A_{\rm o}$ maps them to odd vectors. So $f_{(\te,0)}(z)$ is even, which means $e^\top f_{(\te,0)}(z)=0$, and $(\Id-\frac12ee^\top)K_\theta\mu/\kappa=(H+\te A_{\rm e})\mu=f_{(\te,0)}(z)$, because $\Id-\frac12ee^\top$ fixes even vectors and removes odd ones. The two fields agree on the tied hyperplane and are tangent to it, so they have the same orbit from $z^0$.

(iii) $e^\top H=e^\top$ and $e^\top A_{\rm e}=(A_{\rm e}e)^\top=0$, so $e^\top f_{(\te,0)}=\bar d$ everywhere. At a tied $z$ the gradient of the odd function $\bar d$ is odd, hence a multiple of $e$, and $e^\top\nabla\bar d=2\partial_\zeta\bar d=2g(0,x)$. Then $\partial_t\big(e^\top M_{\te}(t,t_0)\big)=r_{\te}(t)\,e^\top M_{\te}(t,t_0)$, which gives \eqref{eq:fundamental}.

(iv) By (i), (ii) and Lemma~\ref{lem:gapstep}, the spillover along the tied path is $\tho\chih\cdot\mu_\perp(\bar z_{\te})$, the rate is $r_{\te}$, and $v_{K_\theta}=v+\chi^\top\Cov(\xi_\perp)\chi$.

(v) $P_{12}HP_{12}=H$, $P_{12}A_{\rm e}P_{12}=A_{\rm e}$, $P_{12}A_{\rm o}P_{12}=-A_{\rm o}$ and $\mu(p(P_{12}z))=P_{12}\mu(p(z))$.
\end{proof}

\begin{lemma}[Tied orbits]\label{lem:tiedorbits}
Let $z^0$ be tied with $p_1(z^0)>\max_{a\ge3}p_a(z^0)$.
\begin{enumerate}[label=(\roman*),leftmargin=*,itemsep=1pt]
\item The tabular tied path $\bar z_0$ converges to the edge, $p_1=p_2\ge\max_ap_a$ along it, and $r_0(t)\ge r_*=c_G(G-2)n^{-G}$.
\item There are $\varepsilon_{15}>0$, $C$, $c_4$ and $r_{**}>0$ such that for $|\te|\le \varepsilon_{15}$ the orbit $\bar z_{\te}$ converges to the edge, $r_{\te}\ge r_{**}$, and $|\bar x_{\te}(t)-\bar x_0(t)|\le C|\te|(1+t)^{c_4}$ for all $t\ge0$. Consequently $|\beta_A(\te)-\beta_A(0)|\le C|\te|$ and $\big|\int_0^\infty e^{-2\cI_{\te}}v(\bar x_{\te})\,dt-\int_0^\infty e^{-2\cI_0}v(\bar x_0)\,dt\big|\le C|\te|$.
\end{enumerate}
\end{lemma}

\begin{proof}
(i) is Lemma~\ref{lem:tied} with $m=2$.

(ii) Take $T$ with $\rho(\bar z_0(T))<\varepsilon_{11}/2$, where $\varepsilon_{11}$ is given by Lemma~\ref{lem:tiededge} for the kernels $K_{(\te,0)}$ with $|\te|\le \varepsilon_{15}$, whose matrices $Q$ are uniformly strictly copositive for small $\varepsilon_{15}$. On $[0,T]$, Gronwall's lemma gives $|\bar z_{\te}-\bar z_0|\le C_T|\te|$, so for small $\varepsilon_{15}$ the trailing mass of $\bar z_{\te}(T)$ is below $\varepsilon_{11}$ and $r_{\te}\ge r_*/2$ on $[0,T]$. For $t\ge T$ both chart points obey $x'=V^x_{\te}(x)$ with $V^x_{\te}(x)_a=x_a(y_a'-\sum_bx_by_b')$ and $y'=Q_{\te}\mu_\perp(0,x)$, where $Q_{\te}$ is the matrix $Q$ of $K_{(\te,0)}$. These fields satisfy $|V^x_{\te}(x)-V^x_0(x)|\le C|\te|\rho^2$ and $|D_xV^x_{\te}(x)|\le C\rho$, and both masses satisfy $\rho(t)\le C/(1+c(t-T))$ by Lemma~\ref{lem:tiededge}. So $u=\bar x_{\te}-\bar x_0$ obeys
\[
|u|'\ \le\ \frac{C|u|}{1+c(t-T)}+\frac{C|\te|}{(1+c(t-T))^2},
\]
and Gronwall's lemma gives $|u(t)|\le C|\te|(1+t)^{c_4}$ with $c_4=C/c$. The rates $r_{\te}=g(0,\bar x_{\te})$ are at least $r^\circ/2$ for $t\ge T$ once $\varepsilon_{11}$ is small, so $r_{**}=\min(r_*/2,r^\circ/2)$ works. Finally $|r_{\te}-r_0|\le C|u|$, $|\cI_{\te}-\cI_0|\le C|\te|(1+t)^{c_4+1}$, $|\mu_\perp(\bar x_{\te})-\mu_\perp(\bar x_0)|\le C|u|$ and $|v(\bar x_{\te})-v(\bar x_0)|\le C|u|$, and integration against $e^{-r_{**}t}$ gives the last two bounds.
\end{proof}

\begin{theorem}[The separatrix under a weak anisotropy]\label{thm:third}
Let $z^0$ be tied with $p_1(z^0)>\max_{a\ge3}p_a(z^0)$. For each finite $k\ge3$ there are $\varepsilon_{16},\varepsilon_{17}>0$ and a $C^k$ function $\zeta^*$ on $[-\varepsilon_{16},\varepsilon_{16}]^2$ with values in $(-\varepsilon_{17},\varepsilon_{17})$ with the following properties.
\begin{enumerate}[label=(\roman*),leftmargin=*,itemsep=1pt]
\item For $|\te|,|\tho|\le \varepsilon_{16}$ and $|\zeta_0|\le \varepsilon_{17}$, the flow $f_\theta$ from $z^0+\frac12\zeta_0e$ decides for answer $1$ if $\zeta_0>\zeta^*(\theta)$ and for answer $2$ if $\zeta_0<\zeta^*(\theta)$. For $\zeta_0=\zeta^*(\theta)$ it does not decide and converges to the edge.
\item $\zeta^*(\te,-\tho)=-\zeta^*(\te,\tho)$, and in particular $\zeta^*(\te,0)=0$.
\item $\partial_{\tho}\zeta^*(\te,0)=-\beta_A(\te)$.
\item $\zeta^*(\te,\tho)=-\tho\,\beta_A(\te)+O(|\tho|^3)$ uniformly in $|\te|\le \varepsilon_{16}$. In particular $\zeta^*(s,s)=-B_{K_s}+O(|s|^3)$.
\end{enumerate}
\end{theorem}

\begin{proof}
\emph{Step 1. The crossing function.} Apply Lemma~\ref{lem:local} with this $k$ to $\theta\mapsto K_\theta$ at $\theta_*=(0,0)$, where $S=Q=\Id+\frac12\one\one^\top$ is positive definite. This gives $\Theta_0\supset[-\varepsilon_{18},\varepsilon_{18}]^2$, $\varepsilon_6$, $h$ and $F_\theta$. By Lemma~\ref{lem:tiedorbits}(i) there is $T$ with $\rho(\bar z_0(T))<\varepsilon_6/2$, and along $\bar z_0$ no probability exceeds $\frac12$. Let $\varphi^\theta_t$ be the flow of $f_\theta$. By continuity there are $\varepsilon_{17}>0$ and $\varepsilon_{16}\le\min(\varepsilon_{15},\varepsilon_{18})$, with $\varepsilon_{15}$ from Lemma~\ref{lem:tiedorbits}, such that for $|\zeta_0|\le \varepsilon_{17}$ and $|\te|,|\tho|\le \varepsilon_{16}$ the chart point of $\varphi^\theta_T(z^0+\frac12\zeta_0e)$ has $|\zeta|<\varepsilon_6$ and $\rho<\varepsilon_6$, and no probability reaches $1-\delta$ on $[0,T]$. Put
\begin{equation}\label{eq:crossing}
\hat F_T(\zeta_0,\theta)=F_\theta\big(\varphi^\theta_T(z^0+\tfrac12\zeta_0e)\big),
\end{equation}
which is $C^k$. At $\theta=(\te,0)$, $\chi_\theta=0$, so $h(\cdot,(\te,0))=0$ on $\{x\ge0,\rho<\varepsilon_6\}$ by Lemma~\ref{lem:local}(e). The point $\bar x_{\te}(T)$ has positive entries, so it lies in the interior of that set and $\partial_xh(\bar x_{\te}(T),(\te,0))=0$. Hence $DF_{(\te,0)}=e^\top$ at $\bar z_{\te}(T)$, $\hat F_T(0,(\te,0))=e^\top\bar z_{\te}(T)=0$, and by \eqref{eq:fundamental}
\begin{equation}\label{eq:dcC}
\partial_{\zeta_0}\hat F_T\big(0,(\te,0)\big)=\tfrac12\,e^\top M_{\te}(T,0)\,e=e^{\cI_{\te}(T)}\ \ge\ 1 .
\end{equation}
Shrinking $\varepsilon_{17}$ and $\varepsilon_{16}$, $\partial_{\zeta_0}\hat F_T\ge\frac12$ on the box, and $\hat F_T(\varepsilon_{17},\theta)>0>\hat F_T(-\varepsilon_{17},\theta)$. So $\hat F_T(\cdot,\theta)$ has exactly one zero $\zeta^*(\theta)\in(-\varepsilon_{17},\varepsilon_{17})$, and $\zeta^*$ is $C^k$ by the implicit function theorem.

\emph{Step 2. Decisions.} If $\zeta_0>\zeta^*(\theta)$, then $F_\theta>0$ at the time $T$ at a point with $|\zeta|<\varepsilon_6$ and $\rho<\varepsilon_6$, and no decision happened before $T$. By Lemma~\ref{lem:local}(d) the flow decides for answer $1$. The case $\zeta_0<\zeta^*(\theta)$ is the same, and $\zeta_0=\zeta^*(\theta)$ falls under Lemma~\ref{lem:local}(c). This proves (i).

\emph{Step 3. Oddness.} By Lemma~\ref{lem:parity}(v), $P_{12}$ maps the orbit of $f_{(\te,\tho)}$ from $z$ to the orbit of $f_{(\te,-\tho)}$ from $P_{12}z$. Since $P_{12}z^0=z^0$ and $P_{12}e=-e$, it maps the start $z^0+\frac12\zeta_0e$ to $z^0-\frac12\zeta_0e$, and it exchanges the probabilities of the answers $1$ and $2$. So the flow of $f_{(\te,-\tho)}$ from $z^0-\frac12\zeta^*(\te,\tho)e$ does not decide, and the uniqueness in (i) gives $\zeta^*(\te,-\tho)=-\zeta^*(\te,\tho)$.

\emph{Step 4. The first variation.} Fix $|\te|\le \varepsilon_{16}$ and $T'\ge T$. The orbit $\bar z_{\te}$ stays tied and its mass decreases after $T$, so its chart point at $T'$ has $|\zeta|<\varepsilon_6$ and $\rho<\varepsilon_6$. By continuity there is a neighborhood $\mathcal U'$ of $(0,(\te,0))$ on which the chart point of $\varphi^\theta_{T'}(z^0+\frac12\zeta_0e)$ has the same property and no decision happens before $T'$. On $\mathcal U'$ the sign of $\hat F_{T'}(\zeta_0,\theta)=F_\theta(\varphi^\theta_{T'}(z^0+\frac12\zeta_0e))$ determines the decision by Lemma~\ref{lem:local}(c) and (d), as does the sign of $\zeta_0-\zeta^*(\theta)$ by (i). Hence $\hat F_{T'}(\zeta^*(\theta),\theta)=0$ for $\theta$ near $(\te,0)$. As in Step 1, $\partial_{\zeta_0}\hat F_{T'}(0,(\te,0))=e^{\cI_{\te}(T')}$. The derivative $u=\partial_{\tho}\varphi^\theta_t(z^0)$ at $\theta=(\te,0)$ solves
\[
u'=Df_{(\te,0)}(\bar z_{\te})\,u+A_{\rm o}\,\mu(\bar z_{\te}),\qquad u(0)=0 .
\]
At a tie $\mu=\sum_b\mu_b\tilde e_b$, so $e^\top A_{\rm o}\mu=\sum_b\mu_b\,\tilde e_b^\top A_{\rm o}e=\chih\cdot\mu_\perp$. With \eqref{eq:fundamental},
\begin{equation}\label{eq:dthoC}
\partial_{\tho}\,\hat F_{T'}\big(0,(\te,0)\big)=-\partial_{\tho}h\big(\bar x_{\te}(T'),(\te,0)\big)+e^{\cI_{\te}(T')}\int_0^{T'}e^{-\cI_{\te}(t)}\,\chih\cdot\mu_\perp\big(\bar z_{\te}(t)\big)\,dt .
\end{equation}
The implicit function theorem for $\hat F_{T'}$ gives
\begin{equation}\label{eq:firstvar}
\partial_{\tho}\zeta^*(\te,0)=e^{-\cI_{\te}(T')}\,\partial_{\tho}h\big(\bar x_{\te}(T'),(\te,0)\big)-\int_0^{T'}e^{-\cI_{\te}}\,\chih\cdot\mu_\perp(\bar z_{\te})\,dt .
\end{equation}
The left side does not depend on $T'$. The function $\partial_{\tho}h$ is bounded on the compact domain, and $\cI_{\te}(T')\ge r_{**}T'\to\infty$ because $\varepsilon_{16}\le \varepsilon_{15}$. Letting $T'\to\infty$ gives (iii).

\emph{Step 5. Taylor expansion.} For fixed $\te$, $\tho\mapsto \zeta^*(\te,\tho)$ is $C^3$ and odd, so its terms of order $0$ and $2$ vanish, and $\zeta^*(\te,\tho)=\tho\,\partial_{\tho}\zeta^*(\te,0)+R$ with $|R|\le\frac16|\tho|^3\sup|\partial_{\tho}^3\zeta^*|$. With (iii) this is (iv). By Lemma~\ref{lem:parity}(iv), $B_{K_s}=B_{K_{(s,s)}}=s\,\beta_A(s)$, so $\zeta^*(s,s)=-B_{K_s}+O(|s|^3)$.
\end{proof}

The assumption $e^\top Ae=0$ is no restriction. For a general symmetric $A$ with $A\one=0$, put $\kappa_A=\frac12e^\top Ae$ and $A'=A-\kappa_AH$. Then $e^\top A'e=0$ and
\begin{equation}\label{eq:rescale}
H+sA=\big(1+\kappa_As\big)\big(H+s'A'\big),\qquad s'=\frac{s}{1+\kappa_As}=s+O(s^2).
\end{equation}
The mean flow, the critical gap, $B_K$ and $\Sig_K$ depend on $K$ only through $K/\kappa$, so they do not change when $K$ is multiplied by a positive number. Moreover $\tilde e_a^\top A'e=\tilde e_a^\top Ae$, because $\tilde e_a^\top H e=\tilde e_a^\top e=0$. Hence Theorem~\ref{thm:third}(iv) applied to $A'$ gives $\zeta^*(s)=-B_{K_s}+O(|s|^3)$ for $K_s=\kappa(H+sA)$, with the same weights $\chih$. The same rescaling extends Proposition~\ref{prop:consist} below to such $A$. Indeed $B_{K_s}=s'\beta_{A'}(s')$ and $\Sig_{K_s}=\Sig+O(s)$, where $\beta_{A'}$ is computed with $A'$ and $\beta_{A'}(0)=\beta_A$, and the effective step of $K_s$ is $\eta(1+O(s))$ by \eqref{eq:kappa}.

The proof shows more than Proposition~\ref{prop:zero}(ii) states. The linear term of the separatrix in the odd part is the spillover offset computed along the tied path of the even part, to every order in the even part. The even part only moves the tied path, and $B_K$ uses exactly this tied path, which is why the error is of third order and not of second order.

\begin{proposition}[Consistency]\label{prop:consist}
In the setting of Theorem~\ref{thm:third}, $B_{K_s}=s\beta_A+O(s^2)$ and $\Sig_{K_s}=\Sig+O(s)$ as $s\to0$, where $\beta_A$ and $\Sig$ are the constants of Theorem~\ref{thm:weak}. With $s=\sqrt\eta$ and $\zeta_0=\beta\sqrt\eta$, $\Plin\to\Phic((\beta+\beta_A)/\Sig)$.
\end{proposition}

\begin{proof}
By Lemma~\ref{lem:parity}(iv), $B_{K_s}=s\,\beta_A(s)$, and $\beta_A(0)=\beta_A$ because $\bar z_0$ is the tabular tied path. Lemma~\ref{lem:tiedorbits}(ii) gives $\beta_A(s)=\beta_A+O(s)$. Also
\[
\Sig_{K_s}^2=\int_0^\infty e^{-2\cI_s}\Big(v(\bar x_s)+s^2\chih^\top\Cov(\xi_\perp)\chih\Big)dt=\Sig^2+O(s)
\]
by Lemmas~\ref{lem:parity}(iv) and~\ref{lem:tiedorbits}(ii). Then $(\beta\sqrt\eta+\sqrt\eta\,\beta_A(\sqrt\eta))/(\sqrt\eta\,\Sig_{K_{\sqrt\eta}})\to(\beta+\beta_A)/\Sig$.
\end{proof}

\begin{proof}[Proof of Proposition~\ref{prop:zero}]
Part (i) is Proposition~\ref{prop:sepedge}(iii) together with Theorem~\ref{thm:second}. Part (ii) is Theorem~\ref{thm:third}(iv). Part (iii) is Proposition~\ref{prop:consist}.
\end{proof}

\subsection{The Separatrix Law and Its Counterexample}\label{app:seplaw}

In this subsection $K$ is positive semidefinite. We first show that a separatrix orbit forces the copositivity that Lemma~\ref{lem:local} needs.

\begin{lemma}[Copositivity from a critical orbit]\label{lem:copositive}
Suppose an orbit of \eqref{eq:meanflow} starts from a finite $z$ and satisfies $p_1,p_2\to\frac12$ and $p_a\to0$ for $a\ge3$. Then there is $c_K>0$ such that
\begin{equation}\label{eq:conebound}
x^\top S\,x\ \ge\ c_K\Big(\sum_{a\ge3}x_a\Big)^2\qquad\text{for every } x\in\R^{n-2}_{\ge0}.
\end{equation}
In particular $S$ is strictly copositive. Conversely, if $S$ has a nonzero nonnegative null vector, no such orbit exists.
\end{lemma}

\begin{proof}
For $x\in\R^{n-2}$ put
\begin{equation}\label{eq:uvector}
u=\sum_{a\ge3}x_a\tilde e_a-\frac{\chi^\top x}{2}\,e .
\end{equation}
A direct expansion with $e^\top Ke/\kappa=2$ and $e^\top K\tilde e_a/\kappa=\chi_a$ gives
\begin{equation}\label{eq:uKu}
\frac{u^\top Ku}{\kappa}=x^\top Qx-(\chi^\top x)^2+\tfrac12(\chi^\top x)^2=x^\top Sx .
\end{equation}
Since $K$ is positive semidefinite, $S$ is positive semidefinite, and $x^\top Sx=0$ implies $Ku=0$. Now let $x\ge0$ be nonzero with $x^\top Sx=0$. Then $u^\top z(t)$ is constant along the mean flow, and also along every path of the chain. With $y_a=z_a-\frac12(z_1+z_2)=\log p_a-\frac12\log(p_1p_2)$,
\[
u^\top z(t)=\sum_{a\ge3}x_a\,y_a(t)-\frac{\chi^\top x}{2}\,\zeta(t).
\]
Along the orbit $y_a(t)\to-\infty$ for every $a\ge3$ and $\zeta(t)\to0$, so this expression tends to $-\infty$, which contradicts its constant finite value. Hence $x^\top Sx>0$ for every nonzero $x\ge0$, and the minimum of $x^\top Sx$ on the simplex $\{x\ge0,\sum_ax_a=1\}$ is positive. Homogeneity gives \eqref{eq:conebound}. The same argument proves the converse.
\end{proof}

The next theorem is a discrete exit law across a normally repelling graph with center directions. We state it in coordinates $X=(\zeta,x)$ with $x\in\R^{n-2}_{\ge0}$ and $\rho=\sum_ax_a$.

\begin{theorem}[Exit across a normally repelling graph]\label{thm:localexit}
Suppose a $C^3$ function $h$ and a $C^2$ vector field $b$ satisfy, on a compact neighborhood,
\begin{equation}\label{eq:exithyp}
F(X)=\zeta-h(x),\qquad DF(X)\,b(X)=\alpha(X)F(X),\qquad \alpha(X)\ge\alpha_0>0,\qquad D\rho\,b(X)\le-c\rho^2+C\rho|F(X)| .
\end{equation}
Consider a Markov chain with nonnegative $x$ and, uniformly on that neighborhood,
\begin{equation}\label{eq:exitchain}
|X_{k+1}-X_k|\le c_5\eta,\qquad \E_k[X_{k+1}-X_k]=\eta\,b(X_k)+O(\eta^2),\qquad \Cov_k(X_{k+1}-X_k)=\eta^2\cV(X_k)+O(\eta^3),
\end{equation}
where $\cV$ is continuous. Choose $\varepsilon_{19}>0$ and $\varepsilon_{20}>0$ small, with $\varepsilon_{20}<c\varepsilon_{19}/(8C)$, so that the tube $\mathcal D=\{x\ge0,\ \rho<\varepsilon_{19},\ |F|<\varepsilon_{20}\}$ lies in the neighborhood. Let $\bar X$ be the orbit of $b$ on $\{F=0\}$ that starts at a point $X^0$ with $\rho(X^0)<\varepsilon_{19}/4$, and start the chain at $X_0=X^0+\sqrt\eta\,u+o(\sqrt\eta)$ with a fixed vector $u$. Define
\begin{equation}\label{eq:exitell}
\cI_\alpha(t)=\int_0^t\alpha(\bar X(t'))\,dt',\qquad \ell(t)=e^{-\cI_\alpha(t)}DF(\bar X(t)),\qquad \sigma_{\rm loc}^2=\int_0^\infty\ell(t)\,\cV(\bar X(t))\,\ell(t)^\top\,dt .
\end{equation}
If $\sigma_{\rm loc}>0$, then the first exit from $\mathcal D$ is through $\{F=\pm\varepsilon_{20}\}$ with probability tending to one, it occurs within $O_\Pp(\eta^{-1}\log(1/\eta))$ steps, and
\begin{equation}\label{eq:exitlaw}
\lim_{\eta\to0}\Pp\big(F(X_\tau)<0\big)=\Phic\Big(\frac{DF(X^0)\,u}{\sigma_{\rm loc}}\Big).
\end{equation}
\end{theorem}

\begin{proof}
\emph{Step 1. The discounted normal coordinate.} A Taylor expansion of $F$ and \eqref{eq:exitchain} give, before the exit,
\begin{equation}\label{eq:Frec}
F_{k+1}=(1+\eta\alpha_k)F_k+\eta\,\nu_{k+1}+\eta^2R_k ,
\end{equation}
where $\alpha_k=\alpha(X_k)$, $R_k$ is predictable and bounded, and $\nu_{k+1}$ is a bounded martingale difference with $\E_k\nu_{k+1}^2=DF(X_k)\cV(X_k)DF(X_k)^\top+O(\eta)$. The centered quadratic Taylor remainder is included in $\nu_{k+1}$. Put $s_0=1$ and $s_{k+1}=s_k(1+\eta\alpha_k)$ up to the stopping time, which is $\mathcal F_k$-measurable. Then
\begin{equation}\label{eq:discount}
\frac{F_{k\wedge\tau}}{\sqrt\eta\,s_{k\wedge\tau}}=\frac{F_0}{\sqrt\eta}+\sum_{j<k\wedge\tau}\frac{\sqrt\eta\,\nu_{j+1}}{s_{j+1}}+\sum_{j<k\wedge\tau}\frac{\eta^{3/2}R_j}{s_{j+1}} .
\end{equation}
The last sum is bounded by $C\sqrt\eta$ pathwise and uniformly in $k$, because $s_j\ge(1+\eta\alpha_0)^j$.

\emph{Step 2. The Gaussian limit.} For a fixed $T$, the maximal inequality and Gronwall's lemma give $\sup_{k\eta\le T}|X_k-\bar X(k\eta)|=O_\Pp(\sqrt\eta)$. So there is no exit before $T/\eta$ with probability tending to one, and the quadratic variation of the martingale in \eqref{eq:discount} up to this time converges to $\int_0^Te^{-2\cI_\alpha}DF\,\cV\,DF^\top dt$, while its largest increment tends to zero. The martingale central limit theorem gives a normal limit with mean $DF(X^0)u$ and this variance. The conditional variance of the discounted martingale after $T/\eta$ is at most $C\eta\sum_{k\ge T/\eta}(1+\eta\alpha_0)^{-2k}\le C'e^{-2\alpha_0T}$, uniformly over stopping times. Hence, letting first $\eta\to0$ and then $T\to\infty$, the discounted stopped expression converges in law to $\mathcal N(DF(X^0)u,\sigma_{\rm loc}^2)$ along every stopping horizon that diverges.

\emph{Step 3. The center boundary.} On the strip $\varepsilon_{19}/2\le\rho\le\varepsilon_{19}$, $|F|<\varepsilon_{20}$, the drift of $\rho$ is at most $-c_6<0$ by \eqref{eq:exithyp} and the choice of $\varepsilon_{20}$, so the conditional drift of the chain is at most $-c_6\eta/2$ for small $\eta$, and $|\rho_{k+1}-\rho_k|\le c_7\eta$. For a small constant $\varepsilon_{21}>0$, a Taylor expansion of $\exp(\varepsilon_{21}(\rho_{k+1}-\rho_k)/\eta)$ gives $\E_k\exp(\varepsilon_{21}(\rho_{k+1}-\rho_k)/\eta)\le1$ on the strip. Optional stopping of this exponential supermartingale bounds the probability of an upcrossing from $\varepsilon_{19}/2+O(\eta)$ to $\varepsilon_{19}$ by $c_8e^{-c_9/\eta}$. A union bound over the at most $k_1$ possible entrance times before step $k_1$ shows that an exit through the center boundary before step $k_1=O(\eta^{-1}\log(1/\eta))$ has probability $o(1)$.

\emph{Step 4. The exit and its sign.} Take $k_\eta=\lceil c_{10}\eta^{-1}\log(1/\eta)\rceil$ with $c_{10}\alpha_0>\frac12$, and stop at the minimum of $k_\eta$ and the exit from $\mathcal D$. The discounted stopped expression still has the normal limit of Step 2. If there has been no exit, its absolute value is at most
\[
\frac{\varepsilon_{20}}{\sqrt\eta\,(1+\eta\alpha_0)^{k_\eta}}=\varepsilon_{20}\,\eta^{c_{10}\alpha_0-1/2+o(1)}\longrightarrow0 .
\]
Since the limiting normal law has no atom at zero, the probability of no exit tends to zero. On the remaining event $s_\tau>0$, so $F_\tau$ and the discounted expression have the same sign, and the normal limit gives \eqref{eq:exitlaw}.
\end{proof}

\begin{corollary}[An exact adjoint]\label{cor:adjoint}
Along $\bar X$, $\ell'=-\ell\,Db(\bar X)$ in row-vector convention, so $\ell(t)$ is the adjoint of the linearized flow. Under a smooth change to logit coordinates the covector is pulled back by the coordinate map, and for an initial logit displacement $\frac12\beta\sqrt\eta\,e$ the exit probability \eqref{eq:exitlaw} reads $\Phic\big(\beta\,\ell_z(0)e/(2\sigma_{\rm loc})\big)$ with $\sigma_{\rm loc}^2=\int_0^\infty\ell_z\cV_z\ell_z^\top dt$. A positive multiple of $\ell$ does not change this expression.
\end{corollary}

\begin{proof}
Differentiating $DF\,b=\alpha F$ at $F=0$ gives $(D^2F)\,b+DF\,Db=\alpha\,DF$ on the graph. Along $\bar X$ the derivative of $DF(\bar X(t))$ is $(D^2F)\,b$, so $\ell'=-\alpha\ell+e^{-\cI_\alpha}(\alpha DF-DF\,Db)=-\ell\,Db$. The pairing of a covector with a displacement and with a covariance is invariant under a change of coordinates, which gives the second claim.
\end{proof}

\begin{lemma}[The trailing mass vanishes at the exit]\label{lem:rhoexit}
In Theorem~\ref{thm:localexit}, suppose in addition that $\rho(\bar X(t))\to0$, that $|\rho_{k+1}-\rho_k|\le C\eta\rho_k$, and that $\E_k[\rho_{k+1}-\rho_k]=\eta\,D\rho\,b(X_k)+O(\eta^2\rho_k)$. Then $\rho(X_\tau)\to0$ in probability.
\end{lemma}

\begin{proof}
If $\rho=0$, the multiplicative bound keeps it zero. Otherwise a Taylor expansion of $\log\rho$ and \eqref{eq:exithyp} give $\E_k[\log\rho_{k+1}-\log\rho_k]\le\eta C|F_k|+O(\eta^2)$, and the centered increments of $\log\rho$ are $O(\eta)$, so their maximal sum over $O(\eta^{-1}\log(1/\eta))$ steps tends to zero in probability. Let $\tilde F_k=F_k/(\sqrt\eta\,s_k)$. By the tail estimate of Step 2 of the proof of Theorem~\ref{thm:localexit}, for every $\varepsilon_{22}>0$ there are $0<\varepsilon_{23}<c_{11}<\infty$ and a large $T$ such that $\varepsilon_{23}\le|\tilde F_k|\le c_{11}$ for every $T/\eta\le k\le\tau$ with probability at least $1-\varepsilon_{22}$. On this event $|F_\tau|\le\varepsilon_{20}+C\eta$ gives
\[
\sum_{T/\eta\le k<\tau}\eta|F_k|\ \le\ \frac{c_{11}}{\varepsilon_{23}}\,(\varepsilon_{20}+C\eta)\sum_{j\ge1}\eta(1+\eta\alpha_0)^{-j}\ \le\ \frac{c_{11}(\varepsilon_{20}+C\eta)}{\varepsilon_{23}\,\alpha_0},
\]
and therefore
\[
\rho(X_\tau)\ \le\ \rho\big(X_{\lfloor T/\eta\rfloor}\big)\exp\Big(\frac{Cc_{11}(\varepsilon_{20}+C\eta)}{\varepsilon_{23}\,\alpha_0}+o_\Pp(1)\Big).
\]
At a fixed $T$, the first factor converges to $\rho(\bar X(T))$, and $\varepsilon_{23},c_{11}$ can be kept fixed as $T$ grows. Letting $T\to\infty$ and then $\varepsilon_{22}\to0$ proves the claim.
\end{proof}

\begin{proof}[Proof of Theorem~\ref{thm:separatrix}]
\emph{Step 1. The local geometry.} By Lemma~\ref{lem:copositive}, $S$ is strictly copositive. Lemma~\ref{lem:local}, with the kernel fixed, gives a neighborhood $\cN$ of the edge, the graph $\zeta=h(x)$ and the normal coordinate $F=\zeta-h(x)$ with the estimates \eqref{eq:normal}. These are the hypotheses \eqref{eq:exithyp} in the chart $X=(\zeta,x)$, with $b=V_K$. The chain in the chart has bounded increments, conditional mean $\eta V_K+O(\eta^2)$ and conditional covariance $\eta^2\cV_x+O(\eta^3)$, by the Taylor expansion of the softmax, and it satisfies $|\rho_{k+1}-\rho_k|\le C\eta\rho_k$ with a drift remainder $O(\eta^2\rho_k)$. So Theorem~\ref{thm:localexit} and Lemma~\ref{lem:rhoexit} apply.

\emph{Step 2. The early segment.} The orbit of $z^*$ converges to the edge and lies on the graph near it. Let $t_0$ be a time at which it lies in the tube of Theorem~\ref{thm:localexit} with $\rho<\varepsilon_{19}/4$, and $k_0=\lfloor t_0/\eta\rfloor$. On $[0,t_0]$ the finite-time argument of Lemma~\ref{lem:window} applies to the deviation of the chain from the orbit. The deviation $Y_\eta=(z_{k_0}-z^*(t_0))/\sqrt\eta$ converges in law to a normal vector $Y$ whose linearized evolution is $\mathrm dy=Df_K\,y\,\mathrm dt+\mathrm dN$, with $Df_K=K\partial_z\mu/\kappa$ and a martingale $N$ of covariance $\cV_z\,\mathrm dt$, where $\cV_z=K\Cov_z(\xi)K/\kappa^2$. Extend $\ell$ backward from $t_0$ by $\ell'=-\ell\,Df_K$. Then $\mathrm d(\ell y)=\ell\,\mathrm dN$, so
\begin{equation}\label{eq:early}
\ell(t_0)Y\sim\mathcal N\Big(\tfrac12\beta\,\ell(0)e,\ \int_0^{t_0}\ell\,\cV_z\,\ell^\top dt\Big).
\end{equation}

\emph{Step 3. The concatenation.} The proof of Theorem~\ref{thm:localexit} does not use that the initial displacement is deterministic. Started at $X_{k_0}=X^0+\sqrt\eta\,Y_\eta$, where $X^0$ is the chart point of the orbit at $t_0$ and $Y_\eta$ is measurable at step $k_0$, its discounted variable at step $k_0$ equals $DF(X^0)Y_\eta+o(1)$, and the later martingale has a deterministic limit of its conditional quadratic variation. Hence the exit variable converges in law to $\ell(t_0)Y$ plus an independent normal variable with the variance $\int_{t_0}^\infty\ell\cV_z\ell^\top dt$. The two variances add to $(\sigma^*)^2$, and the mean is $\frac12\beta\ell(0)e$. So the side of the exit is negative with probability tending to $\Phic(\beta\ell(0)e/(2\sigma^*))$.

\emph{Step 4. Positivity and finiteness of the noise.} Near the edge $D_zF=e^\top+O(\rho)$, and on the two-answer face the noise of the gap is exactly $d$, because $\ehat^{\,\top}e/2=1$. So $\ell\cV_z\ell^\top\ge\frac12e^{-2\cI_\alpha}v^\circ$ for large $t$, and $\sigma^*>0$. The adjoint is damped like $e^{-\alpha_0t}$ on the tail, so $\sigma^*<\infty$.

\emph{Step 5. From the exit to the winner.} By Lemma~\ref{lem:rhoexit}, the trailing mass at the exit tends to zero in probability, so the exit points approach the two points of the two-answer face with the gap $\pm\varepsilon_{20}$. By Lemma~\ref{lem:leader}, the two-answer orbits from these points cross the threshold for the leader on their side in a finite time, and finite-time tracking identifies the sign of the exit with the winner, with an error probability that tends to zero. The orbit of $z^*$ stays below the threshold with a positive margin, so no answer decides before the exit, and no trailing answer can reach the threshold. This proves \eqref{eq:seplaw}.
\end{proof}

The proof uses the initial displacement only through $\ell(0)u$. Hence for a start $z^*+\sqrt\eta\,u$ with any fixed $u\in\R^n$,
\begin{equation}\label{eq:displacement}
\Pp(W=2)\to\Phic\big(\ell(0)u/\sigma^*\big).
\end{equation}
A displacement of the trailing logits of order $\sqrt\eta$ therefore shifts the law through $\ell(0)$ even when the gap is unchanged.

Near the edge, the hypothesis of Theorem~\ref{thm:separatrix} holds on every line of starts that differ only in the gap.

\begin{proposition}[A critical point near the edge]\label{prop:critical}
Let $K$ be positive semidefinite with $K\one=0$ and $\kappa>0$, and let $S$ be strictly copositive, for instance $K$ positive definite on $\one^\perp$. Let $\varepsilon_{12}$ and $\varepsilon_{13}$ be as in Proposition~\ref{prop:sepedge}(iii). For every tied $z^0$ with $\rho(z^0)<\varepsilon_{12}$, the line $z^0+\frac12\zeta_0e$ with $|\zeta_0|<\varepsilon_{13}$ contains exactly one point $z^*=z^0+\frac12\zeta^*e$ whose mean orbit does not decide. This orbit converges to $p^\circ$ while every probability stays below $1-\delta$, so Theorem~\ref{thm:separatrix} applies at $z^*$.
\end{proposition}

\begin{proof}
Proposition~\ref{prop:sepedge}(iii) gives $\zeta^*$ and its uniqueness on $(-\varepsilon_{13},\varepsilon_{13})$. The chart point of $z^*$ lies in $\cN$ and on the graph, so by Lemma~\ref{lem:local}(c) its orbit stays in $\cN$ and converges to the edge. By the choice of $\varepsilon_4$ in Step 1 of the proof of Lemma~\ref{lem:local}, no probability reaches $1-\delta$ in $\cN$.
\end{proof}

Near the edge the scale of Theorem~\ref{thm:separatrix} agrees with that of the closed form.

\begin{corollary}[The scale near the edge]\label{cor:scale}
In the setting of Proposition~\ref{prop:critical}, normalize $\ell$ by $\ell(0)=D_zF(z^*)$, with the normal coordinate $F$ of Lemma~\ref{lem:local}. Then
\begin{equation}\label{eq:scale}
\frac{2\sigma^*}{\ell(0)e}=\Sig^\circ+O(\rho_0)=\Sig_K+O(\rho_0),
\end{equation}
so the limit \eqref{eq:seplaw} equals $\Phic(\beta/\Sig_K)+O(\rho_0)$, uniformly in $\beta$.
\end{corollary}

\begin{proof}
The orbit $z(t)$ of $z^*$ stays in $\cN$ on the graph, so $\ell(t)=e^{-\cI_\alpha(t)}D_zF(z(t))$ for all $t\ge0$ by Corollary~\ref{cor:adjoint}, with $\cI_\alpha(t)=\int_0^t\alpha(z(t'))\,dt'$. Along the orbit $\rho(t)\le\rho_0$ and $|\zeta(t)|\le C\rho_0$. Since $D_zx_a=x_a(e_a-p)^\top$ and $|\partial_xh|\le C$,
\begin{equation}\label{eq:DFnear}
D_zF(z(t))=e^\top+O(\rho_0),\qquad \ell(0)\,e=2+O(\rho_0).
\end{equation}
The function $\alpha$ is continuously differentiable on the graph with $\alpha(0,0)=r^\circ$, so $|\cI_\alpha(t)-r^\circ t|\le C\rho_0t$. At the edge $\xi=\frac12d\,e$, so $\Cov_{p^\circ}(\xi)=\frac14v^\circ ee^\top$ and $\ehat^{\,\top}\Cov_{p^\circ}(\xi)\,\ehat=v^\circ$ because $\ehat^{\,\top}e=2$. Hence the integrand of $(\sigma^*)^2$ is $e^{-2\cI_\alpha(t)}\big(v^\circ+O(\rho_0)\big)$, and
\[
(\sigma^*)^2=\frac{v^\circ}{2r^\circ}+O(\rho_0).
\]
Together with \eqref{eq:DFnear} this gives the first equality of \eqref{eq:scale}, and \eqref{eq:sepedge} gives the second. Finally $s\mapsto\Phic(\beta/s)$ is Lipschitz near $\Sig^\circ>0$, uniformly in $\beta$.
\end{proof}

Far from the edge the hypothesis can fail, and then the two-leader law fails as well.

\begin{example}[No critical point]\label{ex:nocritical}
Let $n=3$, $G=3$, $\omega\equiv1$ and $K=H$. Then $\mu_a=\frac23p_a\big(p_a-\sum_bp_b^2\big)$, so the mean flow is a gradient flow of $\sum_ap_a^2$, and the set of policies that flow to $(\frac12,\frac12,0)$ is $\{p_1=p_2,\ p_3<\frac13\}$. On the line $\{z_3-\frac12(z_1+z_2)=y\}$, the equality $p_1=p_2$ holds only at $\zeta=0$, where $p_3=e^y/(2+e^y)$. For $y\ge0$ this is at least $\frac13$, so the line contains no critical point. For $y>0$, the starts $z+\frac12\beta\sqrt\eta e$ with $z$ on the line and $\zeta(z)=0$ have $\Pp(W=3)\to1$, because answer $3$ is the unique most likely answer by a fixed margin and Lemma~\ref{lem:noreturn} applies. So the two-leader probit law fails without a critical point. For $y<0$ the critical point is $\zeta=0$, in agreement with Proposition~\ref{prop:sepedge}.
\end{example}

\begin{example}[A trailing answer overtakes the leaders]\label{ex:overtake}
Let $n=4$, $G=8$ and the mean baseline, and let $K=H P^\top MP\,H$ with $P=\big(e_3,e_4,\tfrac12e\big)^\top$ and
\begin{equation}\label{eq:overtakeM}
M=\begin{pmatrix}1&-0.9&0\\-0.9&1&0\\0&0&2\end{pmatrix}.
\end{equation}
By \eqref{eq:decomp}, $M$ is the Gram matrix of $\tilde e_3$, $\tilde e_4$ and $e$ under $K$, so $\kappa=1$, $Q=\big(\begin{smallmatrix}1&-0.9\\-0.9&1\end{smallmatrix}\big)$ and $\chi=0$. The kernel is positive semidefinite with the eigenvalues $0.05$, $1$ and $1.9$ on $\one^\perp$, the matrix $S=Q$ is positive definite, and $Ke=\kappa e$, so there is no spillover. Start from $p^0=(0.30,0.30,0.25,0.15)$. Since $\chi=0$, the tied hyperplane is invariant under the mean flow, and the tied path is a mean orbit. Along it the logit of answer $3$ moves relative to the leaders at the rate
\begin{equation}\label{eq:overtakerate}
y_3'=\mu_3-0.9\,\mu_4,
\end{equation}
which is $+0.022$ at the start, because answer $4$ loses votes ($\mu_4=-0.033$) and the negative entry of $Q$ turns this loss into a gain of answer $3$. Answer $3$ overtakes the tied leaders at $t=6.1$ and wins. On a grid of step $0.01$, the mean flow from $z^0+\frac12\zeta_0e$ decides for answer $3$ for $|\zeta_0|\le0.08$. It decides for answer $1$ at every tested $\zeta_0\ge0.09$ and for answer $2$ at every tested $\zeta_0\le-0.09$, up to $|\zeta_0|=0.6$. So no start near the tie lies on a separatrix between the two leaders. With $\eta=0.01$ and $4000$ runs, the chain from $\zeta_0=0$ is won by answer $3$ in $0.901$ of the runs and by the two leaders in $0.048$ and $0.052$, and from $\zeta_0=0.1$ and $\zeta_0=0.2$ answer $3$ still wins $0.409$ and $0.029$ of the runs. A two-leader law gives answer $3$ no chance. The scope check of Section~\ref{sec:scope} flags this start, because the tied path of $K$ is overtaken. Near the edge the same kernel has a separatrix, by Proposition~\ref{prop:critical}.
\end{example}

\subsection{A Probit With an Omitted Shift}\label{app:omitted}

\begin{proof}[Proof of Proposition~\ref{prop:omitted}]
Let $Y_1$ be standard normal and independent of $B$. Then
\[
\Pp(W=2\mid\zeta_0)=\E\,\Phic\Big(\frac{\zeta_0+B}{\sigma}\Big)=\Pp\big(\sigma Y_1>\zeta_0+B\big)=\Pp\big(\sigma Y_1-B>\zeta_0\big),
\]
and $\sigma Y_1-B$ is normal with mean $-\mu_B$ and variance $\sigma^2+\sigma_B^2$. This gives \eqref{eq:omitted}. Writing $\sigma^2+\sigma_B^2=\lambda\,\eta\Sig^2$ gives $\lambda=1+\sigma_B^2/(\eta\Sig^2)$, and the mean $\mu_B$ appears as the intercept $\mu_B/\sqrt{\sigma^2+\sigma_B^2}$ of the probit.
\end{proof}

At a fixed kernel $B_K$ does not scale with $\eta$, so $\sigma_B^2$ is of order one and $\lambda-1$ grows like $1/\eta$. A set of runs selected near the separatrix correlates $\zeta_0$ with $B$, and then the multiplier is larger than this formula (Section~\ref{sec:lambda}).

\section{Questions That Share Parameters}\label{app:shared}

Let $\cS$ be a finite set of questions. Question $s$ has $n_s$ answers, $G_s\ge3$ votes, a positive $\omega_s$, logits $z_s\in\R^{n_s}$, the update $\xi_s$ and its mean $\mu_s$. The questions share the kernel with the blocks $K_{ss'}=K_{s's}^\top$, every block satisfies $K_{ss'}\one=0$, and all questions are updated at every step,
\begin{equation}\label{eq:sharedchain}
z_{s,k+1}=z_{s,k}+\lr\sum_{s'\in\cS}K_{ss'}\,\xi_{s',k},\qquad s\in\cS,
\end{equation}
where the votes of different questions are drawn independently given the current logits. For question $s$ let $e_s$, $\kappa_s=\frac12e_s^\top K_{ss}e_s$, $\eta_s=\lr\,\kappa_s$ and $\chi_s$ be the quantities of Section~\ref{sec:kernel} for the block $K_{ss}$.

\subsection{The Gap Step}\label{app:sharedstep}

\begin{proof}[Proof of Lemma~\ref{lem:sharedstep}]
By \eqref{eq:sharedchain}, one step changes the gap of question $s$ by
\[
\lr\,e_s^\top K_{ss}\xi_{s,k}+\sum_{s'\ne s}\lr\,e_s^\top K_{ss'}\xi_{s',k}.
\]
By Lemma~\ref{lem:gapstep} for the block $K_{ss}$, the first term is $\eta_s\big(d_{s,k}+\chi_s\cdot\xi_{s,k,\perp}\big)$. Since $K_{ss'}^\top=K_{s's}$,
\begin{equation}\label{eq:crossterm}
\lr\,e_s^\top K_{ss'}\xi_{s',k}=\lr\,\big(K_{s's}e_s\big)^\top\xi_{s',k}=\eta_s\,c_{ss'}^\top\xi_{s',k},
\end{equation}
which gives \eqref{eq:sharedstep}.
\end{proof}

The mean of the cross term is $\eta_s\,c_{ss'}^\top\mu_{s'}$. While question $s'$ is near a tie, its mean update splits into a symmetric part, in which both of its leaders gain and its trailing answers lose, and a part along its own gap, which is of the size of that gap. The symmetric part shifts the race of question $s$ in the same way as its own spillover. The gap part couples the gaps of the two questions. The next theorem shows that only the symmetric part survives when the cross blocks are of the order of the noise.

\subsection{Weakly Coupled Questions}\label{app:weakcoupled}

Let $A_{ss'}$ be fixed real $n_s\times n_{s'}$ matrices with $A_{s's}=A_{ss'}^\top$ and $A_{ss'}\one=0$, and consider the chain
\begin{equation}\label{eq:coupledchain}
z_{s,k+1}=z_{s,k}+\eta\big(H_s+\sqrt\eta\,A_{ss}\big)\xi_{s,k}+\eta\sqrt\eta\sum_{s'\ne s}A_{ss'}\,\xi_{s',k},\qquad s\in\cS,
\end{equation}
with $H_s=\Id-\one\one^\top/n_s$. It is \eqref{eq:sharedchain} with $\lr=\eta/\kappa$, $K_{ss}=\kappa(H_s+\sqrt\eta A_{ss})$ and $K_{ss'}=\kappa\sqrt\eta A_{ss'}$. Question $s$ starts from $z_{s,0}=z^0_s+\sqrt\eta\,\beta_s$ with $\beta_s\in\R^{n_s}$, where $z^0_s$ has $m_s\ge2$ tied leaders, numbered $1,\dots,m_s$, strictly above its other answers. Let $P_s$ be the projection $H_{m_s}$ of Lemma~\ref{lem:schur} for question $s$, let $\bar z_s$, $r_s$ and $\cI_s$ be its tabular tied path, rate and integrated rate, and let $v_s=\Var(\xi_{s,1}-\xi_{s,2})$ along $\bar z_s$. Put
\begin{equation}\label{eq:Fs}
\varphi_s(t)=A_{ss}\,\mu_s\big(\bar z_s(t)\big)+\sum_{s'\ne s}A_{ss'}\,\mu_{s'}\big(\bar z_{s'}(t)\big),\qquad b_s=\int_0^\infty e^{-\cI_s}P_s\varphi_s\,dt,\qquad \sigma_s^2=\frac12\int_0^\infty e^{-2\cI_s}v_s\,dt .
\end{equation}
The integrals converge because $\varphi_s$ is bounded and $\cI_s(t)\ge r_{\rm low}t$ by \eqref{eq:rlow}.

\begin{theorem}[Weakly coupled questions]\label{thm:coupled}
Fix thresholds $1-\delta_s>\frac12$ and let $W_s$ be the winner of question $s$. For every choice of $i_s\le m_s$,
\begin{equation}\label{eq:coupledlaw}
\lim_{\eta\to0}\Pp\big(W_s=i_s\ \text{for all}\ s\in\cS\big)=\prod_{s\in\cS}\Pp\big(Y_{s,i_s}>Y_{s,j}\ \text{for all}\ j\le m_s,\ j\ne i_s\big),
\end{equation}
where $Y_{s,j}=\beta_{s,j}+b_{s,j}+\sigma_sZ_{s,j}$ and the $Z_{s,j}$ are independent standard normal variables. In particular the winners are asymptotically independent, and the probability that some question is won by a trailing answer or never decides tends to zero.
\end{theorem}

\begin{proof}
We follow the proof of Theorem~\ref{thm:block} for every question and list the changes. Write $\tilde\xi_{s,k}=\xi_{s,k}-\mu_s(z_{s,k})$, and note that $H_s\xi_s=\xi_s$.

\emph{Step 1. The joint window.} For a fixed $T$, the deterministic corrections $\eta\sqrt\eta\big(A_{ss}\mu_s+\sum_{s'\ne s}A_{ss'}\mu_{s'}\big)$ accumulate to $O_T(\sqrt\eta)$, and the extra martingale with the increments $\eta\sqrt\eta\big(A_{ss}\tilde\xi_{s,k}+\sum_{s'\ne s}A_{ss'}\tilde\xi_{s',k}\big)$ has a maximal second moment of order $O_T(\eta^2)$. Doob's inequality and Gronwall's lemma give
\begin{equation}\label{eq:cotrack}
\E\max_{k\le T/\eta}\big|z_{s,k}-\bar z_s(k\eta)\big|^2\ \le\ C_T\,\eta\qquad\text{for every } s\in\cS .
\end{equation}
Put $U_{s,k}=P_sz_{s,k}/\sqrt\eta$. As in \eqref{eq:weakU},
\begin{equation}\label{eq:coU}
U_{s,k+1}=\big(1+\eta r_s(k\eta)\big)U_{s,k}+\eta\,P_s\varphi_s(k\eta)+\sqrt\eta\,P_s\tilde\xi_{s,k}+R_{s,k}.
\end{equation}
Besides the terms in the proof of Theorem~\ref{thm:block}, the remainder $R_{s,k}$ contains the difference $\eta P_s\big(A_{ss}\mu_s(z_{s,k})+\sum_{s'\ne s}A_{ss'}\mu_{s'}(z_{s',k})\big)-\eta P_s\varphi_s(k\eta)$, whose accumulated scaled $L^1$ norm is $O_T(\sqrt\eta)$ by \eqref{eq:cotrack} for all questions, and a martingale whose accumulated $L^2$ norm is $O_T(\sqrt\eta)$. The vector martingale with the increments $(\sqrt\eta\,P_s\tilde\xi_{s,k})_{s\in\cS}$ has increments of order $\sqrt\eta$, and since the votes of different questions are conditionally independent with mean zero, its conditional covariance is block diagonal,
\begin{equation}\label{eq:blockcov}
\E_k\Big[\sqrt\eta\,P_s\tilde\xi_{s,k}\big(\sqrt\eta\,P_{s'}\tilde\xi_{s',k}\big)^\top\Big]=\one\{s=s'\}\,\eta\,P_s\Cov_{z_{s,k}}(\xi_s)P_s .
\end{equation}
Variation of constants and the martingale central limit theorem, with $P_s\Cov_{\bar z_s}(\xi_s)P_s=\frac12v_sP_s$ from Lemma~\ref{lem:schur}, give the joint convergence
\begin{equation}\label{eq:jointwindow}
\big(e^{-\cI_s(T)}U_{s,\lfloor T/\eta\rfloor}\big)_{s\in\cS}\Longrightarrow\big(Y_{s,T}\big)_{s\in\cS},\qquad Y_{s,T}\sim\mathcal N\Big(P_s\beta_s+\int_0^Te^{-\cI_s}P_s\varphi_s\,dt,\ \Big(\frac12\int_0^Te^{-2\cI_s}v_s\,dt\Big)P_s\Big),
\end{equation}
with independent components. As $T\to\infty$, $Y_{s,T}\Rightarrow Y_s=P_s\beta_s+b_s+\sigma_sP_sZ_s$, independently over $s$.

\emph{Steps 2 and 3. Amplification and capture.} Steps 2 and 3 of the proof of Theorem~\ref{thm:block} use three facts about one question. The tabular drift satisfies Lemma~\ref{lem:leader}, the corrections change every mean drift of a log-odds by at most $C\sqrt\eta$ per unit of time, and every gap martingale has a conditional step variance of at most $C\eta^2$. In \eqref{eq:coupledchain} the correction to $z_s$ in one step is
\[
\eta\sqrt\eta\Big(A_{ss}\xi_{s,k}+\sum_{s'\ne s}A_{ss'}\xi_{s',k}\Big),
\]
which is bounded by $C'\eta\sqrt\eta$ whatever the states of the other questions are. So the three facts hold for every question with a common constant, conditionally on the past of all questions. Step 2 then holds for every question $s$, with its selected leader $i_s$ and the common horizon $t_\eta$, and the union over the finitely many questions costs a probability of at most $\sum_sC_{m_s}/c_1^2+o(1)$, where $c_1$ is the cutoff of that proof. Step 3 holds for every question, because the tabular flow of question $s$ depends only on $z_s$, and the chain of question $s$ tracks it with an extra drift of order $\sqrt\eta$.

\emph{Step 4. Removal of the cutoffs.} Let $\mathrm{gap}_s(y)$ be the largest block coordinate of $y$ minus the second largest. By Steps 2 and 3,
\begin{equation}\label{eq:comismatch}
\Big|\Pp\big(W_s=i_s\ \forall s\big)-\Pp\big(\argmax\nolimits_{j\le m_s}U_{s,k_T,j}=i_s\ \forall s\big)\Big|\ \le\ \sum_{s\in\cS}\Big(\Pp\big(e^{\cI_s(T)}\mathrm{gap}_s(Y_{s,T})\le c_1\big)+\frac{C_{m_s}}{c_1^2}\Big)+o(1).
\end{equation}
The limit $(Y_{s,T})_s$ has a nondegenerate normal law on the product of the spaces $E_{m_s}$, so ties have probability zero, and the joint law of the block maximizers of $(U_{s,k_T})_s$ converges to that of $(Y_{s,T})_s$, which is a product law. Let $\eta\to0$, then $T\to\infty$ and then $c_1\to\infty$. Since $\cI_s(T)\to\infty$ and $\mathrm{gap}_s(Y_s)>0$ almost surely, the right side of \eqref{eq:comismatch} vanishes. Subtracting the block average of $Z_s$ does not change the maximizer, which gives \eqref{eq:coupledlaw}.
\end{proof}

\begin{corollary}[Two leaders]\label{cor:coupled2}
Let $m_s=2$ and start question $s$ from $z^0_s+\frac12\beta_s\sqrt\eta\,e_s$. Then the winners are asymptotically independent, $\Pp(W_s\ge3)\to0$, and
\begin{equation}\label{eq:coupled2}
\Pp(W_s=2)\to\Phic\Big(\frac{\beta_s+\beta^{\rm f}_s}{\Sig_s}\Big),\qquad \beta^{\rm f}_s=\int_0^\infty\psi_s\,e^{-\cI_s}dt,\qquad \psi_s=\chih_s\cdot\mu_{s,\perp}(\bar z_s)+\sum_{s'\ne s}\big(A_{s's}e_s\big)^\top\mu_{s'}(\bar z_{s'}),
\end{equation}
with $\chih_{s,a}=\tilde e_a^\top A_{ss}e_s$ and $\Sig_s^2=\int_0^\infty v_se^{-2\cI_s}dt$.
\end{corollary}

\begin{proof}
For $m_s=2$, $P_s=\frac12e_se_s^\top$, and the vector $\beta_s$ of the theorem is $\frac12\beta_se_s$. The difference of the first two coordinates of $Y_s$ is $\beta_s+e_s^\top b_s+\sigma_s(Z_{s,1}-Z_{s,2})$. At a tie of question $s$, $\mu_s=\sum_a\mu_{s,a}\tilde e_a$ by \eqref{eq:decomp}, so $e_s^\top A_{ss}\mu_s=\chih_s\cdot\mu_{s,\perp}$, while $e_s^\top A_{ss'}\mu_{s'}=(A_{s's}e_s)^\top\mu_{s'}$. Hence $e_s^\top b_s=\beta^{\rm f}_s$. Finally $2\sigma_s^2=\Sig_s^2$.
\end{proof}

In the variables of Lemma~\ref{lem:sharedstep}, $\chi_s=\sqrt\eta\,\chih_s(1+O(\sqrt\eta))$ and $c_{ss'}=\sqrt\eta\,A_{s's}e_s(1+O(\sqrt\eta))$, so to leading order $\sqrt\eta\,\psi_s$ is the mean of the own spillover plus the mean of the cross term, both along the tied paths. This is the shift of the closed form for shared parameters in Section~\ref{sec:shared}. The gap parts of the other questions do not enter. The gap of question $s'$ is of order $\sqrt\eta$ in its window and enters the gap of question $s$ with a factor $\sqrt\eta$, so its contribution is of order $\eta$ per unit of time, which is $o(\sqrt\eta)$ over a window of fixed length.

With cross blocks of order one, the gap parts enter at order one and the theorem does not apply, as in our shared experiments. On the probes of Section~\ref{sec:shared}, the update of one other question moved a gap by $0.28$ to $0.29$ of what the question's own update did. The share of the own update in the total push on its gap was $0.80$, $0.58$, $0.24$ and $0.08$ for sets of $2$, $4$, $16$ and $64$ questions.

\subsection{A Slow Transient}\label{app:slow}

The proof of Theorem~\ref{thm:coupled} gives convergence and no rate. The following heuristic describes the leading correction when two questions $s$ and $s'$ with $m_s=m_{s'}=2$ have a cross block that couples their gaps directly, through a term $\alpha_3\,e_se_{s'}^\top$ in $A_{ss'}$. The gap of question $s$ then receives $2\alpha_3\eta\sqrt\eta\,d_{s',k}$ per step, that is $2\alpha_3\sqrt\eta\,\bar d_{s'}$ per unit of time, and near the tie $\bar d_{s'}\approx r\zeta_{s'}$. For two questions with the same rate $r$ the linearized gaps obey
\begin{equation}\label{eq:mixode}
\zeta_s'=r\zeta_s+2\alpha_3\sqrt\eta\,r\,\zeta_{s'},\qquad \zeta_{s'}'=r\zeta_{s'}+2\alpha_3\sqrt\eta\,r\,\zeta_s ,
\end{equation}
so the modes $\zeta_s\pm\zeta_{s'}$ grow at the rates $r(1\pm2\alpha_3\sqrt\eta)$. Up to the time $\log(1/\sqrt\eta)/r$ at which the gaps reach order one, this mixes the window variables into
\begin{equation}\label{eq:mixing}
\big(\cosh u\,Y_s+\sinh u\,Y_{s'},\ \sinh u\,Y_s+\cosh u\,Y_{s'}\big),\qquad u\approx2\alpha_3\sqrt\eta\,\log\big(1/\sqrt\eta\big).
\end{equation}
The heuristic therefore predicts a correlation of the winners of order $\alpha_3\sqrt\eta\log(1/\eta)$, which vanishes as $\eta\to0$, but slowly.

We checked this on two questions with $n_s=3$, $G=3$, the mean baseline, the threshold $0.9$, the same start $p^0=(0.4,0.4,0.2)$, $\beta_s=0$, $A_{ss}=0$ and
\begin{equation}\label{eq:A12}
A_{12}=\alpha_1\,e\,\tilde e_3^\top+\alpha_2\,\tilde e_3\,e^\top+\alpha_3\,e\,e^\top,\qquad A_{21}=A_{12}^\top,\qquad e=(1,-1,0)^\top,\quad \tilde e_3=(-\tfrac12,-\tfrac12,1)^\top .
\end{equation}
The term with $\alpha_1$ forces the gap of question $1$ through the mean update of question $2$, the term with $\alpha_2$ forces the gap of question $2$ through that of question $1$, and the term with $\alpha_3$ couples the two gaps. Since $\tilde e_3^\top\mu=\frac32\mu_3$ at a tie, Corollary~\ref{cor:coupled2} gives
\begin{equation}\label{eq:betaf12}
\beta^{\rm f}_1=3\alpha_1\bar\mu_3,\qquad \beta^{\rm f}_2=3\alpha_2\bar\mu_3,\qquad \bar\mu_3=\int_0^\infty\mu_3(\bar z)\,e^{-\cI}dt=-0.17622,
\end{equation}
and the gap coupling does not enter the limit. With $(\alpha_1,\alpha_2)=(1.5,-1)$ and $\Sig=0.66217$ the predicted probabilities are $\Pp(W_1=2)=0.8845$ and $\Pp(W_2=2)=0.2123$, with independent winners. Table~\ref{tab:coupled} compares them with $8000$ runs of \eqref{eq:coupledchain} for $\alpha_3=0$, $\frac14$ and $1$.

\begin{table}[ht]
\caption{Two weakly coupled questions, $8000$ runs per row. Standard errors are about $0.004$ for the marginals and $0.011$ for the correlation of the indicators $\one\{W_s=2\}$. The column ``product'' is the product of the two observed marginals. The last row is the limit of Corollary~\ref{cor:coupled2}, which does not depend on $\alpha_3$. No trailing answer won in any run.}
\label{tab:coupled}
\centering\small\setlength{\tabcolsep}{5pt}
\begin{tabular}{@{}lrrrrrr@{}}
\toprule
coupling & $\eta$ & $\Pp(W_1=2)$ & $\Pp(W_2=2)$ & $\Pp(W_1=W_2=2)$ & product & correlation\\
\midrule
$\alpha_3=0$ & $4\times10^{-3}$ & 0.8878 & 0.2090 & 0.1842 & 0.1855 & $-0.010$\\
$\alpha_3=0$ & $10^{-3}$ & 0.8845 & 0.2162 & 0.1891 & 0.1913 & $-0.016$\\
$\alpha_3=1/4$ & $4\times10^{-3}$ & 0.8746 & 0.2381 & 0.2213 & 0.2083 & $+0.092$\\
$\alpha_3=1/4$ & $10^{-3}$ & 0.8791 & 0.2291 & 0.2104 & 0.2014 & $+0.065$\\
$\alpha_3=1$ & $4\times10^{-3}$ & 0.8080 & 0.3483 & 0.3455 & 0.2814 & $+0.342$\\
$\alpha_3=1$ & $10^{-3}$ & 0.8399 & 0.2899 & 0.2801 & 0.2435 & $+0.220$\\
$\alpha_3=1$ & $2.5\times10^{-4}$ & 0.8649 & 0.2569 & 0.2394 & 0.2222 & $+0.115$\\
\midrule
limit & $\to0$ & 0.8845 & 0.2123 & 0.1878 & 0.1878 & $0$\\
\bottomrule
\end{tabular}
\end{table}

Without the gap coupling the marginals agree with the corollary within one or two standard errors, the joint frequency agrees with the product of the marginals, and the correlation of the winners is zero within two standard errors. With the gap coupling the chain shows the transient. The correlation is roughly proportional to $\alpha_3$, and for $\alpha_3=1$ it falls from $0.34$ to $0.22$ and $0.12$ as $\eta$ goes from $4\times10^{-3}$ to $10^{-3}$ and $2.5\times10^{-4}$, roughly in proportion to $\sqrt\eta\log(1/\sqrt\eta)$. The marginals move as the mixing \eqref{eq:mixing} predicts, since for $\alpha_3>0$ each question is pulled toward the outcome of the other. A least-squares fit of the single parameter $u$ to the three frequencies of a row with $\alpha_3=1$ reproduces them within $0.006$, with $u/(2\alpha_3\sqrt\eta)=2.78$, $3.32$ and $3.81$ at the three steps, against $\log(1/\sqrt\eta)=2.76$, $3.45$ and $4.15$.

At the step sizes of practice, the winners of questions whose gaps are coupled are therefore correlated. A closed form that adds only the mean shift of the other questions treats them as independent and is more confident than the coupled chain, which is what we observe in Section~\ref{sec:shared} and Appendix~\ref{app:addshared}.

\section{Numerical Checks of the Theory}\label{app:numerics}

All checks in this appendix use exact vote moments from \eqref{eq:moments}, and the simulated chains draw exact multinomial votes with the uniform tie break. Unless stated otherwise the baseline is the mean, the threshold is $1-\delta=0.95$, and standard errors are binomial. Critical gaps $\zeta^*$ are found by bisection on the winner of the mean flow, which we integrate with an explicit Runge Kutta method of order eight (DOP853) at the relative tolerance $10^{-10}$ or below. The offsets $B_K$, $\beta_A(\cdot)$ and the constants $\Sig$, $\Sig_K$ are integrals along tied paths, computed with the same integrator.

\subsection{The Kernel of Figure~\ref{fig:race}}\label{app:fig1}

Figure~\ref{fig:race} uses $n=3$, $G=8$ and the kernel
\begin{equation}\label{eq:fig1kernel}
K=u_1u_1^\top+0.3\big(u_1u_2^\top+u_2u_1^\top\big)+u_2u_2^\top,\qquad u_1=\frac{e}{\sqrt2},\qquad u_2=\frac{\tilde e_3}{|\tilde e_3|},
\end{equation}
which has $K\one=0$, $\kappa=1$, the eigenvalues $0$, $0.7$ and $1.3$, and the spillover weight $\chi_3=0.3\sqrt3=0.520$. The starts are $z^0=\log\big(\frac{1-\rho_0}2,\frac{1-\rho_0}2,\rho_0\big)$. Table~\ref{tab:fig1} compares the critical gap with the zero of the closed form. Their difference stays within about $1.3$ percent of $\zeta^*$ up to $\rho_0=0.3$, which is $0.02$ in probit units at $\eta=0.01$. For small $\rho_0$ the relative difference is close to the loop term of Theorem~\ref{thm:second}, $\frac{\gamma^\circ}{2r^\circ}\chi_3^2\rho_0=0.19\rho_0$. Panel (b) of the figure uses $\rho_0=0.2$, $\eta=0.01$ and $6000$ runs of the chain for each of $17$ gaps.

\begin{table}[ht]
\caption{The separatrix of the kernel \eqref{eq:fig1kernel} against the zero of the closed form. The last two columns are the noise constants of the closed form and of the scalar law.}
\label{tab:fig1}
\centering\small
\begin{tabular}{@{}rrrrrr@{}}
\toprule
$\rho_0$ & $\zeta^*$ & $-B_K$ & $(\zeta^*+B_K)/\zeta^*$ & $\Sig_K$ & $\Sig$\\
\midrule
0.02 & 0.01382 & 0.01378 & $+0.3\%$ & 0.4869 & 0.4867\\
0.04 & 0.02631 & 0.02616 & $+0.6\%$ & 0.4830 & 0.4825\\
0.06 & 0.03757 & 0.03729 & $+0.7\%$ & 0.4796 & 0.4788\\
0.10 & 0.05667 & 0.05612 & $+1.0\%$ & 0.4740 & 0.4728\\
0.15 & 0.07428 & 0.07344 & $+1.1\%$ & 0.4693 & 0.4669\\
0.20 & 0.08417 & 0.08324 & $+1.1\%$ & 0.4673 & 0.4627\\
0.25 & 0.08342 & 0.08281 & $+0.7\%$ & 0.4691 & 0.4610\\
0.30 & 0.06155 & 0.06237 & $-1.3\%$ & 0.4779 & 0.4640\\
\bottomrule
\end{tabular}
\end{table}

\subsection{Two Answers and the Error of Order $\eta$}\label{app:numeta}

For two answers with $G=8$, every value of $d$ is a multiple of $1/32$, so the chain lives on the lattice $\zeta_0+\frac\eta{32}\mathbb Z$ and $\Pp(W=2)$ can be computed exactly by solving the backward equation on this lattice. We solved $\mathcal T_0\varphi_1=-\mathcal T_1\varphi_0$ numerically with the rational coefficients of Appendix~\ref{app:eta}. Table~\ref{tab:eta2} shows that the error divided by $\eta$ approaches $\varphi_1(\beta)$ and that the remainder divided by $\eta^2$ is nearly constant in $\eta$, as Proposition~\ref{prop:eta2} states. Over $41$ values of $\beta/\Sig$ in $[-5,5]$, the largest value of $|\Pp(W=2)-\Phic(\beta/\Sig)-\eta \varphi_1(\beta)|/\eta^2$ is $0.0012$ at $\eta=0.03$ and at $\eta=0.01$. Even at $\eta=1$, the first-order formula is within $16$ percent of the exact error.

\begin{table}[ht]
\caption{Two answers, $G=8$. Exact $\Pp(W=2)$ from the lattice, its limit $\Phic(\beta/\Sig)$, the error divided by $\eta$, the coefficient $\varphi_1(\beta)$ of Proposition~\ref{prop:eta2} and the remainder $\big(\Pp(W=2)-\Phic(\beta/\Sig)-\eta \varphi_1(\beta)\big)/\eta^2$.}
\label{tab:eta2}
\centering\small
\begin{tabular}{@{}lrrrrrr@{}}
\toprule
$\eta$ & $\beta/\Sig$ & exact & limit & error$/\eta$ & $\varphi_1(\beta)$ & remainder$/\eta^2$\\
\midrule
1.0 & 0.5 & 0.296983 & 0.308538 & $-0.01155$ & $-0.01130$ & $-0.00026$\\
1.0 & 1.0 & 0.142548 & 0.158655 & $-0.01611$ & $-0.01618$ & $+0.00008$\\
1.0 & 2.0 & 0.015508 & 0.022750 & $-0.00724$ & $-0.00839$ & $+0.00115$\\
0.3 & 0.5 & 0.305131 & 0.308538 & $-0.01135$ & $-0.01130$ & $-0.00019$\\
0.3 & 1.0 & 0.153813 & 0.158655 & $-0.01614$ & $-0.01618$ & $+0.00015$\\
0.3 & 2.0 & 0.020340 & 0.022750 & $-0.00803$ & $-0.00839$ & $+0.00120$\\
0.1 & 0.5 & 0.307406 & 0.308538 & $-0.01131$ & $-0.01130$ & $-0.00018$\\
0.1 & 1.0 & 0.157039 & 0.158655 & $-0.01617$ & $-0.01618$ & $+0.00017$\\
0.1 & 2.0 & 0.021923 & 0.022750 & $-0.00827$ & $-0.00839$ & $+0.00121$\\
0.03 & 0.5 & 0.308198 & 0.308538 & $-0.01130$ & $-0.01130$ & $-0.00017$\\
0.03 & 1.0 & 0.158170 & 0.158655 & $-0.01618$ & $-0.01618$ & $+0.00018$\\
0.03 & 2.0 & 0.022499 & 0.022750 & $-0.00836$ & $-0.00839$ & $+0.00122$\\
0.01 & 0.5 & 0.308425 & 0.308538 & $-0.01130$ & $-0.01130$ & $-0.00017$\\
0.01 & 1.0 & 0.158493 & 0.158655 & $-0.01618$ & $-0.01618$ & $+0.00018$\\
0.01 & 2.0 & 0.022666 & 0.022750 & $-0.00838$ & $-0.00839$ & $+0.00122$\\
\bottomrule
\end{tabular}
\end{table}

With eight answers we ran a tabular study with the leader at $0.4$, the runner-up at $0.3$ or $0.35$ and six answers sharing the rest, both baselines and four group sizes, sixteen conditions per step size. The mean signed error of the limit law is $-0.0008\pm0.0005$ at $\eta=0.03$, $-0.0014\pm0.0008$ at $\eta=0.1$, $-0.0042\pm0.0010$ at $\eta=0.3$ and $-0.0135\pm0.0011$ at $\eta=1$, where the second number is the standard error. Only the growth from $\eta=0.3$ to $\eta=1$ is resolved. There the error grows by a factor of $3.2$ for a factor of $3.3$ in $\eta$, which fits an error of order $\eta$. An error of order $\sqrt\eta$ would grow by a factor of about $1.8$.

\subsection{Several Tied Answers}\label{app:numtie}

Table~\ref{tab:tie3} checks Corollary~\ref{cor:thurstone} with three answers tied at $0.25$ and five answers at $0.05$ ($n=8$, $G=8$, $\Sig=0.4437$). Over all rows the largest deviation from the law is $2.1$ standard errors, and no answer outside the tie won. A larger run with $120{,}000$ chains at $\eta=0.03$ and $\beta/\Sig=(0.5,0,-0.5)$ gave $0.6314$, $0.2709$ and $0.0977$ against $0.6303$, $0.2722$ and $0.0975$.

\begin{table}[ht]
\caption{Winning frequencies of three tied answers, with the Thurstone law in parentheses. The start is $z^0+\sqrt\eta\,\beta$, and $\beta$ is given in units of $\Sig$. Standard errors are about $0.0024$ near $0.3$ to $0.6$ and $0.0015$ near $0.1$ for $40{,}000$ runs.}
\label{tab:tie3}
\centering\small\setlength{\tabcolsep}{4pt}
\begin{tabular}{@{}lrlccc@{}}
\toprule
$\eta$ & runs & $\beta/\Sig$ & answer 1 & answer 2 & answer 3\\
\midrule
0.03 & 40{,}000 & $(0,0,0)$ & 0.338 (0.333) & 0.331 (0.333) & 0.331 (0.333)\\
0.03 & 40{,}000 & $(0.5,0,-0.5)$ & 0.629 (0.630) & 0.274 (0.272) & 0.097 (0.097)\\
0.03 & 40{,}000 & $(1.0,0.5,0)$ & 0.631 (0.630) & 0.274 (0.272) & 0.095 (0.097)\\
0.03 & 40{,}000 & $(0.7,0.7,0)$ & 0.442 (0.443) & 0.444 (0.443) & 0.114 (0.115)\\
0.01 & 40{,}000 & $(0,0,0)$ & 0.333 (0.333) & 0.332 (0.333) & 0.335 (0.333)\\
0.01 & 40{,}000 & $(0.5,0,-0.5)$ & 0.627 (0.630) & 0.277 (0.272) & 0.096 (0.097)\\
0.01 & 40{,}000 & $(1.0,0.5,0)$ & 0.632 (0.630) & 0.272 (0.272) & 0.096 (0.097)\\
0.01 & 40{,}000 & $(0.7,0.7,0)$ & 0.441 (0.443) & 0.445 (0.443) & 0.114 (0.115)\\
0.01 & 80{,}000 & $(0.5,0,-0.5)$ & 0.633 (0.630) & 0.270 (0.272) & 0.096 (0.097)\\
0.003 & 20{,}000 & $(0.5,0,-0.5)$ & 0.631 (0.630) & 0.276 (0.272) & 0.093 (0.097)\\
\bottomrule
\end{tabular}
\end{table}

With four answers tied at $0.2$ and four answers at $0.05$ ($G=8$, $\Sig=0.4467$), Table~\ref{tab:tie4} shows the same agreement with the Thurstone law. The largest deviation is a slight excess of the answer with the largest $\beta$ at $\eta=0.03$ and $\eta=0.01$, of at most $0.006$, which corresponds to a $\Sig$ smaller by about $2$ percent, and at $\eta=0.003$ and $\eta=0.001$ the frequencies agree with the law within one standard error.

\begin{table}[ht]
\caption{Winning frequencies of four tied answers, with the Thurstone law in parentheses.}
\label{tab:tie4}
\centering\small\setlength{\tabcolsep}{3pt}
\begin{tabular}{@{}lrlcccc@{}}
\toprule
$\eta$ & runs & $\beta/\Sig$ & answer 1 & answer 2 & answer 3 & answer 4\\
\midrule
0.03 & 40{,}000 & $(0.6,0.2,-0.2,-0.6)$ & 0.556 (0.552) & 0.277 (0.277) & 0.122 (0.123) & 0.045 (0.047)\\
0.03 & 40{,}000 & $(0.3,0.3,0,-0.6)$ & 0.374 (0.368) & 0.366 (0.368) & 0.208 (0.210) & 0.052 (0.054)\\
0.03 & 200{,}000 & $(0.6,0.2,-0.2,-0.6)$ & 0.556 (0.552) & 0.276 (0.277) & 0.121 (0.123) & 0.046 (0.047)\\
0.01 & 80{,}000 & $(0.6,0.2,-0.2,-0.6)$ & 0.558 (0.552) & 0.275 (0.277) & 0.121 (0.123) & 0.046 (0.047)\\
0.003 & 100{,}000 & $(0.6,0.2,-0.2,-0.6)$ & 0.553 (0.552) & 0.278 (0.277) & 0.122 (0.123) & 0.047 (0.047)\\
0.001 & 40{,}000 & $(0.6,0.2,-0.2,-0.6)$ & 0.555 (0.552) & 0.277 (0.277) & 0.121 (0.123) & 0.047 (0.047)\\
\bottomrule
\end{tabular}
\end{table}

The law also holds when the trailing answers carry most of the mass. With $p_1=p_2=0.2$ and six answers at $0.1$ ($n=8$, $G=8$, $\Sig=0.4230$), Table~\ref{tab:rho06} agrees with $\Phic(\beta/\Sig)$, and no other answer won.

\begin{table}[ht]
\caption{Probability that the runner-up wins from a start with trailing mass $0.6$, $40{,}000$ runs per row.}
\label{tab:rho06}
\centering\small
\begin{tabular}{@{}lrrrr@{}}
\toprule
$\eta$ & $\beta/\Sig$ & $\Pp(W=2)$ & $\Phic(\beta/\Sig)$ & standard error\\
\midrule
0.03 & 0.5 & 0.3078 & 0.3085 & 0.0023\\
0.03 & 1.0 & 0.1608 & 0.1587 & 0.0018\\
0.03 & 2.0 & 0.0223 & 0.0228 & 0.0007\\
0.01 & 0.5 & 0.3094 & 0.3085 & 0.0023\\
0.01 & 1.0 & 0.1545 & 0.1587 & 0.0018\\
0.01 & 2.0 & 0.0225 & 0.0228 & 0.0007\\
\bottomrule
\end{tabular}
\end{table}

\subsection{Three Leaders Under a Weak Anisotropy}\label{app:numblock}

Theorem~\ref{thm:block} predicts that a weak anisotropy can move all the winning probabilities of a tie. Take $G=3$, $n=4$, three leaders and the start $(q,q,q,\rho_0)$ with $\rho_0=0.15$ and $q=(1-\rho_0)/3$. Put
\begin{equation}\label{eq:threeA}
u_1=\tfrac1{\sqrt6}(2,-1,-1,0)^\top,\qquad u_2=e_4-\tfrac13(e_1+e_2+e_3),\qquad A=2\big(u_1u_2^\top+u_2u_1^\top\big).
\end{equation}
Along the tabular tied path $\mu=(\mu_1,\mu_1,\mu_1,-3\mu_1)$ with $\mu_1>0$, so $H_3A\mu=-8\mu_1u_1$ and
\begin{equation}\label{eq:threebA}
b_A=-8u_1\int_0^\infty\mu_1(t)\,e^{-\cI(t)}\,dt=(-0.4211,\,0.2106,\,0.2106,\,0),\qquad \sigma_3=0.5291 .
\end{equation}
At $\beta=0$ the limit \eqref{eq:blocklaw} gives the winning probabilities $(0.0867,0.4567,0.4567)$, far from the symmetric $\frac13$. With the threshold $0.9$ and $6000$ runs, the chain gave $(0.0863,0.4517,0.4620)$ at $\eta=2\times10^{-3}$ and $(0.0842,0.4527,0.4632)$ at $\eta=5\times10^{-4}$, with standard errors of about $0.0036$ for the first answer and $0.0064$ for the other two.

\subsection{The Separatrix Near the Edge and Under a Weak Anisotropy}\label{app:numsep}

For a small trailing mass, the kernel is $K=H(XX^\top+0.6\,\Id)H$ on five answers with a Gaussian $5\times5$ matrix $X$, and the trailing logits have a fixed random direction and a common shift that sets $\rho_0$. The threshold is $0.9$. Table~\ref{tab:mass} lists $\zeta^*$, $-B_K$, the ratio $(\zeta^*+B_K)/\rho_0^2$ next to the loop term $B_{\rm loop}=\frac{\gamma^\circ}{2r^\circ}\zeta^*\sum_a\chi_a^2p_a(z^0)$ of Theorem~\ref{thm:second} divided by $\rho_0^2$, the remainder $(\zeta^*+B_K-B_{\rm loop})/\rho_0^3$ and the ratio $(\zeta^*+B_{\rm edge})/\rho_0^2$.

\begin{table}[ht]
\caption{The separatrix near the edge for two kernels on five answers, $G=3$ with $\chi=(0.62,0.19,0.23)$ and $G=4$ with $\chi=(0.44,0.60,-0.74)$. The ratio $(\zeta^*+B_K)/\rho_0^2$ converges to the loop term, and the remainder is of order $\rho_0^3$.}
\label{tab:mass}
\centering\small\setlength{\tabcolsep}{4pt}
\begin{tabular}{@{}rrrrrrrr@{}}
\toprule
$G$ & $\rho_0$ & $\zeta^*$ & $-B_K$ & $\frac{\zeta^*+B_K}{\rho_0^2}$ & $\frac{B_{\rm loop}}{\rho_0^2}$ & $\frac{\zeta^*+B_K-B_{\rm loop}}{\rho_0^3}$ & $\frac{\zeta^*+B_{\rm edge}}{\rho_0^2}$\\
\midrule
3 & 0.1 & 0.0482229 & 0.0478617 & $+0.0361$ & $+0.0668$ & $-0.307$ & $-1.591$\\
3 & 0.05 & 0.0272549 & 0.0271258 & $+0.0517$ & $+0.0756$ & $-0.478$ & $-1.924$\\
3 & 0.025 & 0.0146765 & 0.0146356 & $+0.0654$ & $+0.0814$ & $-0.641$ & $-2.170$\\
3 & 0.0125 & 0.0076527 & 0.0076409 & $+0.0753$ & $+0.0849$ & $-0.765$ & $-2.326$\\
3 & 0.00625 & 0.0039137 & 0.0039105 & $+0.0815$ & $+0.0868$ & $-0.846$ & $-2.417$\\
3 & 0.003125 & 0.0019800 & 0.0019791 & $+0.0850$ & $+0.0878$ & $-0.893$ & $-2.466$\\
3 & 0.0015625 & 0.0009959 & 0.0009957 & $+0.0869$ & $+0.0883$ & $-0.918$ & $-2.492$\\
\midrule
4 & 0.1 & 0.0326129 & 0.0317843 & $+0.0829$ & $+0.1760$ & $-0.931$ & $-1.681$\\
4 & 0.05 & 0.0195711 & 0.0192377 & $+0.1334$ & $+0.2112$ & $-1.556$ & $-2.056$\\
4 & 0.025 & 0.0108940 & 0.0107811 & $+0.1807$ & $+0.2351$ & $-2.177$ & $-2.338$\\
4 & 0.0125 & 0.0057838 & 0.0057500 & $+0.2163$ & $+0.2497$ & $-2.665$ & $-2.520$\\
4 & 0.00625 & 0.0029862 & 0.0029769 & $+0.2391$ & $+0.2578$ & $-2.986$ & $-2.626$\\
4 & 0.003125 & 0.0015182 & 0.0015157 & $+0.2522$ & $+0.2621$ & $-3.173$ & $-2.683$\\
4 & 0.0015625 & 0.0007656 & 0.0007649 & $+0.2592$ & $+0.2644$ & $-3.275$ & $-2.713$\\
\bottomrule
\end{tabular}
\end{table}

Both ratios of second order converge, as Proposition~\ref{prop:sepedge} states. The ratio $(\zeta^*+B_K)/\rho_0^2$ approaches the loop term, and their difference falls like $\rho_0^3$, as Theorem~\ref{thm:second} states. The coefficient of $-B_K$ is ten to thirty times smaller than that of $-B_{\rm edge}$.

For a weak anisotropy we take $n=4$, $G=3$, $z^0=\log(0.35,0.35,0.18,0.12)$, the threshold $0.9$ and a random symmetric $A$ with $A\one=0$ and $e^\top Ae=0$, scaled to the spectral norm one. The offset $B_{K_\theta}/\tho$ equals $\beta_A(\te)$ to $4\times10^{-11}$ for $\tho=0.1$ and $0.3$ at $\te=0.25$, as Lemma~\ref{lem:parity}(iv) states, with $\beta_A(0)=0.185863$ and $\beta_A(0.25)=0.183004$. The critical gap $\zeta^*(0.25,0)$ is zero to $10^{-15}$, and $\zeta^*(\te,-\tho)=-\zeta^*(\te,\tho)$ holds to all printed digits. Table~\ref{tab:aniso} shows that the remainders after the linear term are of third order.

\begin{table}[ht]
\caption{The separatrix under a weak anisotropy, with $R_{\te}(s)=(\zeta^*(\te,s)+s\,\beta_A(\te))/s^3$ and $R_{\rm d}(s)=(\zeta^*(s,s)+B_{K_s})/s^3$. The remainders after the linear term $-\tho\,\beta_A(\te)$ are of third order in $\tho$ for a fixed $\te$, and along the diagonal the error of $-B_{K_s}$ is of third order in $s$.}
\label{tab:aniso}
\centering\small\setlength{\tabcolsep}{4pt}
\begin{tabular}{@{}rrrrrrrr@{}}
\toprule
$s$ & $\zeta^*(0,s)$ & $10^3R_0(s)$ & $\zeta^*(0.25,s)$ & $10^3R_{0.25}(s)$ & $\zeta^*(s,s)$ & $10^3R_{\rm d}(s)$ & $10^3R_{\rm d}(-s)$\\
\midrule
0.4 & $-0.074795857$ & $-7.042$ & $-0.073628291$ & $-6.667$ & $-0.072920344$ & $-6.437$ & $-7.542$\\
0.2 & $-0.037228368$ & $-6.972$ & $-0.036653628$ & $-6.603$ & $-0.036769398$ & $-6.678$ & $-7.243$\\
0.1 & $-0.018593250$ & $-6.955$ & $-0.018306988$ & $-6.588$ & $-0.018479451$ & $-6.810$ & $-7.094$\\
0.05 & $-0.009294016$ & $-6.951$ & $-0.009151023$ & $-6.584$ & $-0.009265669$ & $-6.879$ & $-7.021$\\
\bottomrule
\end{tabular}
\end{table}

\subsection{The Closed Form Against the Chain on Synthetic Kernels}\label{app:numsynth}

Before the comparison on the probed kernels we checked the closed form on synthetic kernels with $n=10$, $G=8$ and $2000$ samples of the chain per run, as in the experiments. The kernels are the centered versions of $D^{1/2}(XX^\top+c_X\Id)D^{1/2}$, with $X$ an $n\times r_X$ Gaussian matrix scaled by $r_X^{-1/2}$ and $D$ diagonal and log-normal with the standard deviation $s_D$. The level \emph{weak} has $(r_X,c_X,s_D)=(30,0.3,0)$. The level \emph{LM} has $(6,0.3,0.9)$, with a median self-kernel difference $|K_{11}-K_{22}|/(2\kappa)$ of $0.32$ and a median norm of the trailing part of $Ke$ of $0.75$ relative to its part on the leaders, close to the values $0.39$ and $0.79$ of the reference setting of Qwen-1.5B. The level \emph{strong} has $(3,0.1,0.6)$. Each level has $100$ kernels and starts with a trailing mass between $0.15$ and $0.6$. For every kernel and step, the initial gap was set to $\zeta_0=-B_K+\sqrt\eta\,\Sig_Ku$ with $u$ uniform on $[-2.5,2.5]$, clipped so that the two leaders stay on top. This choice uses the law but not the chain. We compare the chain's probability that the runner-up beats the leader, $\Pmc=\Pp(W=2)/(\Pp(W=1)+\Pp(W=2))$, with the laws, and leave out the kernels whose tied path lets a trailing answer overtake the leaders ($0$, $1$ and $3$ kernels at the three levels).

\begin{table}[ht]
\caption{The closed form against the frozen-kernel chain on synthetic kernels. Mean absolute difference from $\Pmc$ on the runs with $\Pmc\in(0.01,0.99)$, for $\Plin$, for the scalar law with the edge offset $B_{\rm edge}$ of Proposition~\ref{prop:sepedge}(iii) (edge) and for the scalar law (scalar). Probit fit of the chain on the law, on the runs with $\Pmc\in(0.02,0.98)$, with the slope for $\Plin$ and the standard deviation of the probit difference. The last two columns are the step multipliers that the chain implies for the scalar law and for $\Plin$.}
\label{tab:val}
\centering\small\setlength{\tabcolsep}{4pt}
\begin{tabular}{@{}lrrrrrrrrrr@{}}
\toprule
& & & \multicolumn{3}{c}{mean abs.\ difference} & \multicolumn{3}{c}{probit fit} & \multicolumn{2}{c}{$\lambda$ from the chain}\\
\cmidrule(lr){4-6}\cmidrule(lr){7-9}\cmidrule(l){10-11}
level & $\eta$ & runs & $\Plin$ & edge & scalar & slope & sd $\Plin$ & sd scalar & scalar & $\Plin$\\
\midrule
weak & 0.3 & 97 & 0.008 & 0.027 & 0.051 & 1.02 & 0.04 & 0.21 & 1.1 & 1.02\\
weak & 0.1 & 98 & 0.008 & 0.037 & 0.083 & 1.01 & 0.04 & 0.35 & 1.1 & 1.00\\
weak & 0.03 & 95 & 0.008 & 0.057 & 0.161 & 1.00 & 0.04 & 0.67 & 2.1 & 0.99\\
weak & 0.01 & 95 & 0.007 & 0.076 & 0.226 & 1.00 & 0.04 & 1.07 & 4.3 & 1.00\\
LM & 0.3 & 95 & 0.016 & 0.060 & 0.176 & 1.10 & 0.12 & 0.81 & 2.8 & 1.05\\
LM & 0.1 & 90 & 0.018 & 0.085 & 0.224 & 1.06 & 0.13 & 1.15 & 8.1 & 1.03\\
LM & 0.03 & 78 & 0.025 & 0.126 & 0.279 & 1.00 & 0.22 & 1.94 & 37 & 1.07\\
LM & 0.01 & 68 & 0.032 & 0.176 & 0.314 & 1.01 & 0.21 & 2.59 & 153 & 1.11\\
strong & 0.3 & 91 & 0.038 & 0.073 & 0.201 & 1.14 & 0.29 & 0.90 & 3.7 & 0.92\\
strong & 0.1 & 85 & 0.029 & 0.089 & 0.267 & 1.14 & 0.23 & 1.24 & 8.6 & 0.84\\
strong & 0.03 & 62 & 0.034 & 0.129 & 0.289 & 1.07 & 0.29 & 1.97 & 40 & 0.95\\
strong & 0.01 & 63 & 0.074 & 0.191 & 0.385 & 1.07 & 0.42 & 3.05 & $3\times10^3$ & 1.27\\
\bottomrule
\end{tabular}
\end{table}

Table~\ref{tab:val} shows the pattern of the experiments. The scalar law breaks down as the step shrinks at every level. At the level LM, its logistic recalibration slope with the chain's probability as the target is $0.54$, $0.20$, $0.09$ and $0.00$ for $\eta=0.3$, $0.1$, $0.03$ and $0.01$, against $1.10$, $1.07$, $0.99$ and $1.00$ for $\Plin$. $\Plin$ follows the chain with a mean absolute difference of $0.008$ at the level weak, which is the Monte Carlo error of the chain, and of $0.016$ to $0.032$ at the level LM. The chain implies a step multiplier near one for $\Plin$ and one that grows from $2.8$ to $153$ for the scalar law at the level LM. This growth is faster than $1+\sigma_B^2/(\eta\Sig^2)$ of Proposition~\ref{prop:omitted}, because the starts were chosen around the separatrix, where $\zeta_0$ and $B_K$ are negatively correlated.

What remains of $\Plin$ at the level LM is an offset of the point at which the chain is even. As $\eta$ shrinks it converges to $\zeta^*+B_K$, with $\zeta^*$ the critical gap of the mean flow. For the $99$ kernels of the level LM whose path is not overtaken, the correlation of the two offsets is $0.85$, $0.90$, $0.98$ and $0.98$ at the four steps, with slopes between $0.99$ and $1.14$. Across these kernels $\zeta^*+B_K$ is close to $-1.1B_K^3$, with $R^2=0.94$. It is below $0.01$ for $|B_K|<0.1$, and for $|B_K|>0.25$ the separatrix lies $7$ to $23$ percent farther out than $-B_K$. This is the coefficient $1.1$ that Section~\ref{sec:match} compares with the probed kernels. The edge offset captures the spillover kernel by kernel but overstates its size at these trailing masses. Its correlation with $B_K$ is $0.98$, $0.98$ and $0.97$ at the three levels, while the regression slope of $B_K$ on $B_{\rm edge}$ is $0.66$, $0.60$ and $0.55$, because the trailing answers lose mass along the path. On one kernel of the level LM, $B_K/B_{\rm edge}$ is $0.82$, $0.93$, $0.97$ and $0.99$ for $\rho_0=0.2$, $0.05$, $0.02$ and $0.005$, as Proposition~\ref{prop:sepedge} requires.

\section{Experimental Details}\label{app:exp}

\subsection{Questions and Prompts}\label{app:questions}

The questions come from the test split of MMLU-Pro \citep{wang2024mmlupro}, whose questions have ten options with the letters A to J. The prompt is the chat template of the model with the question and the lettered options, and the assistant turn ends right before the answer letter, so that the next token is the letter. Every letter is a single token. The letter mass is the total probability of the ten letter tokens at this position, and the policy of a question is the softmax of the ten letter logits. For Llama, whose chat template inserts the current date, we fixed the date to the default of the template.

For each model we selected $500$ questions before any training, among those with at most $400$ tokens, a letter mass of at least $0.8$ and a largest probability below $0.95$. The selection was stratified over cells of the gap between the two leading logits and of the largest probability, and it kept at least $50$ questions whose leader is wrong and whose runner-up is right. For Qwen2.5-1.5B, $113$ of the $500$ questions are of this kind.

\subsection{Models, Adapters and Training}\label{app:training}

We use Qwen2.5-1.5B-Instruct, Llama-3.2-1B-Instruct and Qwen2.5-7B-Instruct \citep{qwen25report,grattafiori2024llama3}. The adapters are LoRA of rank $32$ on all attention and MLP projections \citep{hu2022lora}, with $B=0$ at the start, so the start logits do not depend on the seed while the kernel does through the random matrices $A$ of the adapters. A run trains one question. At every step it samples $G$ letters from the letter policy, computes the majority update \eqref{eq:update}, and takes one optimizer step on the surrogate $\sum_a\xi_a z_a(\theta)$ with $\xi$ held fixed, so that SGD moves the parameters by $\lr\,J^\top\xi$. There is no clipping, no KL term and no entropy term. A run stops at the first step at which a letter reaches the probability $0.95$, or at a cap of three times the mean decision step that the chain predicts for the run, with a minimum of $50$ steps. Runs that reach the cap are scored by their most likely letter. The seed of a run sets both the initialization of the adapters and the stream of samples, so runs with the same question and seed in different settings are paired.

All computations are in float32 with TF32 disabled. On Qwen2.5-1.5B, the median relative error of one step with respect to the kernel prediction, $|\Delta z-\lr\,K\xi|/|\lr\,K\xi|$, was $0.066$, $0.010$, $0.0028$, $0.0049$ and $0.016$ at $\eta=0.003$, $0.01$, $0.03$, $0.1$ and $0.3$ in float32. In bfloat16 it was $120$, $37$, $14$, $6.6$ and $2.1$, because the letter logits then move only by rounding jumps larger than one step. One step of the 1.5B model took about $0.17$ seconds on our GPUs.

\subsection{Kernel Probes}\label{app:probes}

For every question and seed we computed the exact Jacobian $J$ of the ten letter logits with respect to the LoRA parameters at the start, with ten backward passes, and the kernel $K=JJ^\top$. For AdamW the kernel is $K=JDJ^\top$ with the static preconditioner $D=\mathrm{diag}\big(1/(\sqrt{\E g^2}+\varepsilon_{\rm A})\big)$ at the start, where the second moment of the gradient $g=J^\top\xi$ is computed from the vote moments as $\E g_i^2=\sum_{a,b}J_{ai}J_{bi}\E[\xi_a\xi_b]$, with $\E[\xi_a\xi_b]$ estimated from $20{,}000$ sampled groups. This kernel depends on $G$ and on the baseline, so the probes were repeated for every group size. Momentum is not part of the kernel, and the chain predictions of AdamW with $\beta_1=0$ and $\beta_1=0.9$ coincide.

For a shared set the kernel has the blocks $K_{ss'}=J_sJ_{s'}^\top$. The diagonal blocks are exact. The cross blocks were estimated with a CountSketch of the parameter space with $2^{18}$ buckets, which adds noise of the order of $|J_s||J_{s'}|/2^9$ to every entry. For AdamW on shared sets the preconditioner uses the exact second moment of the gradient of the set, $\E g^2=\sum_s\big(\sum_{a,b}C_{s,ab}J_{sa}J_{sb}-(J_s^\top m_s)^2\big)+\big(\sum_sJ_s^\top m_s\big)^2$ with $C_s=\E[\xi_s\xi_s^\top]$, $m_s=\E\xi_s$ and products taken entrywise, which holds because the votes of different members are independent.

\subsection{Settings and Runs}\label{app:settings}

For each model and optimizer the learning rate was set before training so that the median of $\eta=\lr\,\kappa$ over all questions and seeds equals $0.1$ at $G=8$. For Qwen2.5-1.5B this gives $\lr=3.52\times10^{-5}$ for SGD. The reference setting has this rate and $G=8$. The five unseen settings use a third or three times this rate at $G=8$, or the reference rate with $G=4$, $16$ or $32$. All six settings have three seeds. In the unseen settings we trained the runs whose prediction had $\eta\in[0.01,1]$ and a probability between $0.01$ and $0.99$ that the leader loses, and in the settings with another $G$ also every run on a question with a wrong leader and a right runner-up. This gives $4{,}765$ runs and $1.31$ million steps. Llama-3.2-1B has the reference setting and two unseen settings, a third of the rate and $G=4$, with $2{,}631$ runs. Qwen2.5-7B has one setting with $\lr=1.475\times10^{-5}$, $G=8$ and two seeds, with $1{,}000$ runs.

The three settings with other optimizers train all $500$ questions with three seeds on Qwen2.5-1.5B. SGD with the standard-deviation baseline divides the advantage by the unbiased standard deviation of the group rewards and uses the reference rate. AdamW uses $\beta_1\in\{0,0.9\}$, $\beta_2=0.999$, $\varepsilon_{\rm A}=10^{-8}$, no weight decay and the rate that puts the median $\eta$ of the static Adam kernel at $0.1$.

The shared sets use the reference rate of SGD and $G=8$. We trained $250$, $125$, $31$ and $14$ sets of $2$, $4$, $16$ and $64$ questions, every step updating all members, until every member had decided. The $14$ sets of $64$ were trained again with SGD and with AdamW ($\beta_1=0.9$) for $300$ steps, also after the members had decided. For AdamW the rate was set with the diagonal blocks of the shared Adam kernel, because the shared preconditioner divides by the gradient of the whole set.

\subsection{Predictions Fixed Before Training}\label{app:registration}

Before training the runs of each model and setting, we computed the frozen-kernel chain $\Pmc$ for every run, a Monte Carlo of \eqref{eq:chain} with the probed kernel, exact multinomial votes over the ten letters, the threshold $0.95$ and $2{,}000$ samples. The random numbers of each run are derived from its seed and its index, so a prediction does not depend on the order of the computation. $\Pmc$ is the fraction of samples whose winner is not the leader, and a sample that has not decided at the step cap of the run is scored by its most likely letter, as the runs are.

Together with the predictions, we fixed the tests below, their success criteria and the code, and saved SHA-256 hashes of all these files before training. After training we checked that none of them had changed. We will release the files and their hashes with the code.

The tests use the evaluated runs of each setting, those with $\eta\in[0.01,0.3]$ and a predicted probability between $0.01$ and $0.99$ that the leader loses, without the $64$ runs of the reference setting that we had used to measure the drift of the kernel. The upper limit $0.3$ was fixed before training, because the tabular law overstates the probability of losing by up to $0.035$ at $\eta=1$. The comparison model is a logistic regression of the event that the leader wins on three start features, the gap of the two leading logits, the logarithm of the largest probability and the entropy of the start policy, fitted on the outcomes of the tested runs with two-fold cross-fitting by question. A test of noninferiority asks whether the Brier score of $\Pmc$ exceeds that of this model by less than $0.003$. Calibration slopes are the slopes of a logistic recalibration of the outcomes on the logit of the prediction, and a test of equivalence asks whether the slope lies in $[0.8,1.25]$. Intervals come from bootstrap resampling of questions, or of sets for shared sets, and the tests are one-sided with Holm corrections within each group of tests. Table~\ref{tab:alltests} lists all tests.

For Qwen2.5-1.5B we first fixed only the tests of the scalar law. After the scalar law had failed in the reference setting, and before any unseen setting was trained, we added the tests of the chain, which therefore use only the five unseen settings.

\begin{table*}[t]
\caption{All tests fixed before training. The first group tests the scalar law and was fixed before the reference setting was trained. The two tests of the chain on Qwen-1.5B were added after the reference setting and before the unseen settings, and every later model or setting has its own group. Intervals are $95\%$ bootstrap intervals.}
\label{tab:alltests}
\centering\small\setlength{\tabcolsep}{4pt}
\begin{tabular}{@{}llllr@{}}
\toprule
model and runs & hypothesis & result & outcome\\
\midrule
Qwen-1.5B, unseen, two letters & slope of $\Ptwo$ in $[0.8,1.25]$ & $0.077$ $[0.014,0.157]$ & failed\\
Qwen-1.5B, unseen, three or more & Thurstone law beats softmax & $-0.052$ $[-0.141,0.045]$ & not established\\
Qwen-1.5B, unseen, two letters & scalar law with one fitted step, noninferior & $+0.0107$ & failed\\
shared sets of $4$ & winners of a set agree more than isolated runs & $0.032$ $[0.014,0.051]$ & passed\\
\midrule
Qwen-1.5B, unseen & slope of $\Pmc$ in $[0.8,1.25]$ & $0.84$ $[0.75,0.94]$ & not established\\
Qwen-1.5B, unseen & $\Pmc$ noninferior to the refitted logistic & $-0.075$ & passed\\
\midrule
Llama-1B, unseen & $\Pmc$ noninferior to the refitted logistic & $-0.084$ & passed\\
Llama-1B, unseen & slope of $\Pmc$ in $[0.8,1.25]$ & $0.80$ $[0.69,0.94]$ & not established\\
Qwen-7B & $\Pmc$ noninferior to the refitted logistic & $-0.066$ $[-0.087,-0.048]$ & passed\\
Qwen-7B & slope of $\Pmc$ in $[0.8,1.25]$ & $1.56$ $[1.23,2.05]$ & not established\\
\midrule
shared sets of $16$ and $64$ & coupled chain beats isolated chain & $-0.086$ & passed\\
\midrule
AdamW, $\beta_1=0.9$ & $\Pmc$ noninferior to the refitted logistic & $-0.057$ $[-0.077,-0.037]$ & passed\\
SGD, std baseline & $\Pmc$ noninferior to the refitted logistic & $-0.075$ $[-0.092,-0.057]$ & passed\\
shared sets of $64$, AdamW & coupled chain beats isolated chain & $-0.104$ $[-0.148,-0.063]$ & passed\\
shared sets of $64$, AdamW & winners of a set agree more than isolated runs & $0.060$ $[0.044,0.075]$ & passed\\
\bottomrule
\end{tabular}
\end{table*}

\subsection{The Closed Form and the Scope Check}\label{app:closedform}

For every run we integrated the tied path $\bar z'=(\Id-\frac12ee^\top)K\mu(\bar z)/\kappa$ from the start logits with the two leaders tied at their mean logit, and computed $r$, $v_K$, $\psi_K$ and the integrals \eqref{eq:BK} along it. The vote moments are exact for $G\le8$, by enumeration over the ten letters. For $G\ge16$ they are Monte Carlo moments with common random numbers along the path. The computation takes about a quarter of a second per run at $G=8$. The scalar law $\Ptwo$ uses the same moments along the tabular tied path. The code of the closed form was fixed before we compared it with the chain or the outcomes.

A run is in scope when the tied path of its kernel keeps the two leaders above every other letter. Definition~\ref{def:lin} covers only these runs. For the flagged runs we evaluated the same integrals on the finite horizon of the path integration, and we use the resulting numbers only to show where the excess error lies. The two-leader target is the chain's probability that the runner-up wins given that one of the two leaders wins, which leaves out the wins of a third letter. Probit fits use the runs with this probability in $(0.02,0.98)$. A step multiplier $\lambda$ for a law $P$ is the factor by which the step in $P$ must be multiplied to fit a target, by maximum likelihood on the outcomes or by least squares in probit units on the chain. Intervals of Brier differences come from $1{,}000$ bootstrap resamples of questions. In the cubic regression of Section~\ref{sec:match}, the weights are the inverse Monte Carlo variances of the chain's probit.

For shared sets, the closed form integrates the tied paths of all members jointly under the shared kernel, keeping the two leaders of every member tied, and adds the mean of the cross term of Lemma~\ref{lem:sharedstep} to the spillover of each member. The gap parts of the other members are left out, as in Corollary~\ref{cor:coupled2}. The coupled chain is the Monte Carlo of the shared frozen-kernel chain, computed before training.

\section{Additional Results}\label{app:addresults}

\subsection{Predictions by Setting}\label{app:bysetting}

Table~\ref{tab:bysetting} breaks the comparison of Table~\ref{tab:registered} down by setting. The chain beats the refitted logistic model in every setting, and its calibration slope ranges from $0.71$ to $1.56$. In the reference settings, which are not part of the confirmatory tests, the slopes are $0.91$ for Qwen-1.5B and $1.22$ for Llama-1B. On the unseen settings of Qwen-1.5B the scalar law has Brier scores of $0.241$ to $0.295$, against $0.143$ to $0.159$ for the chain. On the questions whose leader is wrong and whose runner-up is right, the runner-up won $0.192$, $0.150$, $0.147$ and $0.118$ of the runs of Qwen-1.5B at $G=4$, $8$, $16$ and $32$, and the chain predicted $0.185$, $0.153$, $0.132$ and $0.111$. To fit the outcomes, the scalar law needed step multipliers between $11.8$ at $G=4$ and $40.2$ at $G=32$.

\begin{table}[ht]
\caption{Predictions of the chain by setting, on the evaluated runs. The slope is the calibration slope with its $95\%$ interval.}
\label{tab:bysetting}
\centering\small\setlength{\tabcolsep}{4pt}
\begin{tabular}{@{}llrrl@{}}
\toprule
model & setting & runs & Brier & slope\\
\midrule
Qwen-1.5B & $G=4$ & 547 & 0.148 & 0.76 $[0.63,0.95]$\\
Qwen-1.5B & $G=16$ & 392 & 0.143 & 1.06 $[0.90,1.30]$\\
Qwen-1.5B & $G=32$ & 364 & 0.159 & 0.83 $[0.69,1.04]$\\
Qwen-1.5B & $3\,\lr$ & 383 & 0.143 & 0.84 $[0.68,1.10]$\\
Qwen-1.5B & $\lr/3$ & 316 & 0.154 & 0.86 $[0.69,1.09]$\\
\midrule
Llama-1B & $G=4$ & 578 & 0.142 & 0.82 $[0.69,0.99]$\\
Llama-1B & $\lr/3$ & 339 & 0.145 & 0.77 $[0.61,1.00]$\\
\midrule
Qwen-7B & reference & 418 & 0.117 & 1.56 $[1.23,2.05]$\\
\midrule
Qwen-1.5B & AdamW, $\beta_1=0$ & 605 & 0.126 & 0.95 $[0.81,1.15]$\\
Qwen-1.5B & AdamW, $\beta_1=0.9$ & 600 & 0.159 & 0.71 $[0.59,0.86]$\\
Qwen-1.5B & SGD, std baseline & 602 & 0.129 & 0.95 $[0.81,1.14]$\\
\bottomrule
\end{tabular}
\end{table}

\subsection{The Step Multiplier by Setting}\label{app:lambda}

Table~\ref{tab:lambda} lists, for the thirteen settings, the step multipliers on the evaluated runs that are in scope and have two close letters, with the two-leader target. The multiplier that the chain implies for $\Plin$ lies between $0.90$ and $1.03$. The multiplier that $\Plin$ implies for $\Ptwo$ uses neither the chain nor the outcomes, and it follows the multiplier that the outcomes imply, with the rank correlation $0.93$. The variance part of Proposition~\ref{prop:omitted}, $1+\Var(B_K)/\langle\eta\Sig^2\rangle$, is between $1.3$ and $2.5$ on these runs and misses most of the effect, because the evaluated runs are selected near the separatrix, where $\zeta_0$ and $B_K$ are negatively correlated in eleven of the thirteen settings. On all two-letter runs in scope, without this selection, the multiplier that $\Plin$ implies for $\Ptwo$ is smaller by a factor of $1.3$ to $3.1$ (column ``$\Plin$, all''), except at a third of the rate, where nearly all runs in scope are evaluated anyway. Outside this exception it lies between $0.78$ and $1.83$ times $1+\E[B_K^2/(\eta\Sig^2)]$, the size of Proposition~\ref{prop:omitted} when the mean of $B_K$ is counted in the omitted shift.

On all evaluated runs, with the runs outside the scope and those with three or more close letters, the outcomes imply the multipliers $23.0$, $19.3$, $2.25$ and $8.5$ for the scalar law in Qwen-1.5B, Llama-1B, Qwen-7B and the optimizer settings, with the probability that the leader loses as the target. The median of $|B_K|/(\sqrt\eta\,\Sig_K)$ over these runs is $0.73$, $1.12$, $0.22$ and $0.66$.

Section~\ref{sec:match} quotes the multipliers $15.2$ and $1.11$ from the chain and $18.5$ and $1.56$ from the outcomes for the reference setting of Qwen-1.5B. They are fitted to the probability that the leader loses on all two-letter evaluated runs, as in the tests fixed before training, and therefore differ from the entries of the table.

\begin{table*}[t]
\caption{Step multipliers by setting on the two-letter evaluated runs that are in scope, with the two-leader target. The first two columns of multipliers are those that the chain and the outcomes imply for $\Plin$, and the next three are those that $\Plin$, the chain and the outcomes imply for the scalar law. The column ``$\Plin$, all'' is the multiplier that $\Plin$ implies for the scalar law on all two-letter runs in scope, without the restriction to evaluated runs. The column ``Prop.~\ref{prop:omitted}'' is $1+\Var(B_K)/\langle\eta\Sig^2\rangle$, and the last column is the correlation of $\zeta_0$ and $B_K$.}
\label{tab:lambda}
\centering\footnotesize\setlength{\tabcolsep}{4.5pt}
\begin{tabular}{@{}llrrrrrrrrr@{}}
\toprule
& & & \multicolumn{2}{c}{$\lambda$ for $\Plin$ from} & \multicolumn{4}{c}{$\lambda$ for $\Ptwo$ from} & & \\
\cmidrule(lr){4-5}\cmidrule(lr){6-9}
model & setting & runs & chain & outcomes & $\Plin$ & chain & outcomes & $\Plin$, all & Prop.~\ref{prop:omitted} & corr$(\zeta_0,B_K)$\\
\midrule
Qwen-1.5B & $3\,\lr$ & 193 & 0.90 & 0.92 & 7.11 & 8.21 & 7.35 & 3.27 & 2.00 & $+0.03$\\
& reference & 320 & 0.94 & 1.34 & 9.43 & 11.46 & 15.15 & 3.81 & 1.74 & $-0.35$\\
& $\lr/3$ & 255 & 0.99 & 1.13 & 31.63 & 54.97 & 21.67 & 28.84 & 2.14 & $-0.59$\\
& $G=4$ & 337 & 0.96 & 1.51 & 7.53 & 9.81 & 5.66 & 4.61 & 1.76 & $-0.16$\\
& $G=16$ & 294 & 0.91 & 0.95 & 17.37 & 22.69 & 27.20 & 6.93 & 1.88 & $-0.29$\\
& $G=32$ & 296 & 0.92 & 1.35 & 18.04 & 22.96 & 38.40 & 7.49 & 1.95 & $-0.37$\\
\midrule
Qwen-7B & reference & 333 & 0.96 & 0.52 & 2.25 & 2.35 & 1.70 & 1.78 & 1.26 & $-0.36$\\
\midrule
Llama-1B & $G=4$ & 203 & 0.91 & 1.47 & 3.99 & 4.07 & 4.38 & 2.70 & 2.01 & $+0.30$\\
& reference & 227 & 0.97 & 0.70 & 10.80 & 12.91 & 9.36 & 3.52 & 2.16 & $-0.35$\\
& $\lr/3$ & 256 & 1.03 & 1.46 & 32.88 & 46.86 & 27.32 & 32.83 & 2.53 & $-0.39$\\
\midrule
Qwen-1.5B & AdamW, $\beta_1=0$ & 408 & 0.93 & 0.94 & 4.71 & 5.07 & 4.56 & 2.51 & 1.95 & $-0.48$\\
& AdamW, $\beta_1=0.9$ & 408 & 0.94 & 1.89 & 4.71 & 5.12 & 7.13 & 2.51 & 1.95 & $-0.49$\\
& SGD, std baseline & 364 & 0.92 & 0.91 & 5.06 & 5.74 & 4.86 & 2.46 & 1.47 & $-0.10$\\
\bottomrule
\end{tabular}
\end{table*}

By the step, on the two-letter evaluated runs of Qwen-1.5B, the outcome slope of our recomputed scalar law is $0.52$, $0.13$ and $0.05$ for $\eta$ in $[0.1,0.3)$, $[0.03,0.1)$ and $[0.01,0.03)$, close to the slopes $0.57$, $0.09$ and $0.04$ in the test of the scalar law fixed before training (Table~\ref{tab:alltests}) and to the pattern of Table~\ref{tab:val}. That of $\Plin$ is $0.89$, $0.75$ and $0.49$, against $0.87$, $0.77$ and $1.06$ for the chain. The shortfall below $\eta=0.03$ comes from the $18$ percent of runs there whose tied path is overtaken. On the runs in scope the slopes in that bin are $1.08$ for $\Plin$ and $1.03$ for the chain, and the difference of their Brier scores is $-0.002$ $[-0.010,0.006]$.

\subsection{Runs Outside the Scope}\label{app:addscope}

Table~\ref{tab:scope} splits the excess Brier score of $\Plin$ over the chain by the scope check and by the number of close letters. In scope, the two-letter runs show no excess in any group, and the runs with three or more close letters lose only the share of the third letter, which no two-leader law can produce. The flagged runs carry almost all of the excess. In them a third letter wins often, and the chain predicts how often.

\begin{table}[ht]
\caption{The Brier score of $\Plin$ minus that of the chain, by scope and by the number of close letters, with $95\%$ intervals where the runs suffice. The last column gives the frequency with which a letter other than the two leaders won, predicted by the chain and observed.}
\label{tab:scope}
\centering\small\setlength{\tabcolsep}{4pt}
\begin{tabular}{@{}lllrrlc@{}}
\toprule
group & scope & letters & runs & difference & interval & third letter wins\\
\midrule
Qwen-1.5B & in scope & two & 1{,}695 & $-0.0001$ & $[-0.0028,0.0025]$ & \\
& in scope & three or more & 561 & $+0.0079$ & $[0.0007,0.0168]$ & \\
& flagged & two & 47 & $+0.09$ & & $0.40$, $0.40$\\
& flagged & three or more & 153 & $+0.33$ & $[0.19,0.48]$ & $0.60$, $0.61$\\
\midrule
Qwen-7B & in scope & two & 333 & $+0.0012$ & $[-0.0005,0.0035]$ & \\
& in scope & three or more & 83 & $+0.0038$ & $[-0.0071,0.0197]$ & \\
& flagged & three or more & 2 & & & \\
\midrule
Llama-1B & in scope & two & 686 & $+0.0025$ & $[-0.0021,0.0075]$ & \\
& in scope & three or more & 478 & $+0.0012$ & $[-0.0047,0.0068]$ & \\
& flagged & two & 20 & $+0.19$ & & $0.27$, $0.30$\\
& flagged & three or more & 171 & $+0.18$ & $[0.07,0.29]$ & $0.34$, $0.31$\\
\midrule
Qwen-1.5B, opt. & in scope & two & 1{,}180 & $+0.0000$ & $[-0.0020,0.0018]$ & \\
& in scope & three or more & 494 & $+0.0069$ & $[0.0004,0.0149]$ & \\
& flagged & two & 13 & $+0.09$ & & $0.24$, $0.31$\\
& flagged & three or more & 120 & $+0.33$ & $[0.20,0.47]$ & $0.64$, $0.55$\\
\bottomrule
\end{tabular}
\end{table}

\subsection{The Cubic Offset}\label{app:cubic}

Table~\ref{tab:cubic} gives the regression of Section~\ref{sec:match}. The probit residual of $\Plin$ against the chain, on all two-letter runs in scope without the restriction to evaluated runs and with the chain's probability in $(0.02,0.98)$, is regressed on $u=B_K^3/(\sqrt\eta\,\Sig_K)$ and on the probit of $\Plin$. The coefficient of $u$ has the sign and the size of the synthetic value $1.1$ of Appendix~\ref{app:numsynth}, and it stays between $0.48$ and $1.49$ when $B_K/(\sqrt\eta\,\Sig_K)$ enters the regression as well. Moving the zero of the closed form to $-B_K-1.1B_K^3$ removes the intercept of about $-0.06$ and lowers the spread of the probit difference by $39$, $26$ and $27$ percent for Qwen-1.5B, Qwen-7B and the optimizer settings, while it barely changes Llama-1B, whose coefficient is $0.56$. The move does not change the Brier scores on the outcomes, since the chain itself does not predict them better than $\Plin$.

\begin{table}[ht]
\caption{The cubic offset on the probed kernels. Coefficient of $u=B_K^3/(\sqrt\eta\,\Sig_K)$ with its standard error, share $R^2$ of the residual variance, and the standard deviation and intercept of the probit difference to the chain before and after the zero is moved to $-B_K-1.1B_K^3$.}
\label{tab:cubic}
\centering\small\setlength{\tabcolsep}{4pt}
\begin{tabular}{@{}lrrrll@{}}
\toprule
group & runs & coefficient & $R^2$ & sd, before and after & intercept, before and after\\
\midrule
Qwen-1.5B & 1{,}810 & 0.87 (0.01) & 0.72 & 0.132, 0.081 & $-0.061$, $+0.002$\\
Qwen-7B & 351 & 1.62 (0.08) & 0.56 & 0.061, 0.045 & $-0.019$, $-0.006$\\
Llama-1B & 621 & 0.56 (0.04) & 0.30 & 0.159, 0.156 & $-0.049$, $+0.033$\\
Qwen-1.5B, opt. & 1{,}096 & 0.79 (0.02) & 0.69 & 0.104, 0.076 & $-0.057$, $-0.001$\\
\bottomrule
\end{tabular}
\end{table}

\subsection{Optimizers and Momentum}\label{app:momentum}

AdamW with $\beta_1=0$ and with $\beta_1=0.9$ share the static Adam kernel, so their laws coincide and their chains differ only by the Monte Carlo error. On the two-letter runs in scope, the outcome slopes are $0.97$ for the chain and $1.00$ for $\Plin$ without momentum, and $0.72$ and $0.71$ with momentum. On $1{,}500$ pairs of runs with the same question and seed, momentum lowered the frequency with which the leader wins by $0.023$ $[0.009,0.035]$, while the frozen-kernel model predicts no change. A chain that adds the bias-corrected first moment of Adam to the frozen kernel, which we built after seeing these outcomes, predicts a change of $0.006$ in this frequency and a slope of $0.87$ for the chain, against $0.023$ and $0.71$ observed, so it explains only part of the effect. Momentum therefore lies outside the model, and the closed form adds no error to the chain. The standard-deviation baseline enters the law only through $\omega$. Replacing the mean baseline by it lowered the frequency with which the leader wins by $0.008$ $[-0.003,0.019]$ on paired runs, against the predicted $0.013$.

\subsection{Shared Sets}\label{app:addshared}

Table~\ref{tab:shared} compares the closed form with the shared shift to the coupled chain on the members whose tied path keeps their two leaders on top. The shift of a member is the sum of its own offset $B_K$ and of the cross offset $B_{\rm cross}$, the integral of the mean cross term of Lemma~\ref{lem:sharedstep} against $e^{-\cI}$ along the tied paths. The isolated law, computed from the diagonal blocks of the shared probes, agrees with the isolated law of the single-question runs to $3\times10^{-7}$ and follows the isolated chain of the same members with the probit slopes $1.05$ and $1.06$. The overconfidence of the shared closed form therefore comes from sharing. Its residual has no cubic part and is a pure scale, as for an omitted random shift, and at size $64$ it grows with the cross forcing, with the probit slopes $0.58$, $0.30$ and $0.28$ in the terciles of $|B_{\rm cross}|$, whose medians are $0.15$, $0.57$ and $1.45$. The median of $|B_{\rm cross}|$ grows from $0.021$ at size $2$ to $0.574$ at size $64$, while that of the own offset falls from $0.108$ to $0.050$. On the sets of $64$ retrained with SGD and with AdamW, the closed form closes $91$ and $92$ percent of the Brier gap between the isolated law and the coupled chain. In the sets of $64$ retrained with SGD, the leader won in $0.682$ of the members, against $0.667$ predicted by the coupled chain, and in $0.834$ of the isolated runs of the same members, against $0.827$ predicted by the isolated chain. The shifts of the letter frequencies that sharing caused correlated at $0.97$ with the predicted shifts. With AdamW the five numbers are $0.690$, $0.673$, $0.836$, $0.842$ and $0.86$.

\begin{table*}[t]
\caption{Shared sets. Probit slopes of the chain on the law for the isolated law against the isolated chain and for the shared closed form against the coupled chain, with the two-leader target, and the step multiplier of the shared closed form. The change that sharing causes in the probability that the leader loses is compared between the coupled chain and the closed form by the Pearson correlation and the slope. The Brier scores are for the event that the leader wins, and the last column is the share of the Brier gap between the isolated law and the coupled chain that the shared closed form closes.}
\label{tab:shared}
\centering\small\setlength{\tabcolsep}{4pt}
\begin{tabular}{@{}rrrrrrrrrrrr@{}}
\toprule
& & \multicolumn{2}{c}{probit slope} & & \multicolumn{2}{c}{shift by sharing} & \multicolumn{4}{c}{Brier score} & \\
\cmidrule(lr){3-4}\cmidrule(lr){6-7}\cmidrule(lr){8-11}
size & members & isolated & shared & $\lambda$ & Pearson & slope & coupled chain & shared $\Plin$ & isolated chain & isolated law & closed\\
\midrule
2 & 476 & 1.05 & 0.67 & 1.39 & 0.53 & 0.68 & 0.0487 & 0.0564 & 0.0579 & 0.0575 & 13\%\\
4 & 466 & 1.06 & 0.54 & 1.82 & 0.56 & 0.77 & 0.0660 & 0.0862 & 0.0913 & 0.0932 & 26\%\\
16 & 440 & 1.06 & 0.36 & 3.06 & 0.73 & 0.89 & 0.0824 & 0.0825 & 0.1070 & 0.1088 & 100\%\\
64 & 719 & 1.06 & 0.33 & 3.59 & 0.87 & 0.96 & 0.0805 & 0.0884 & 0.1498 & 0.1500 & 89\%\\
\bottomrule
\end{tabular}
\end{table*}

\subsection{The Kernel During Training}\label{app:growth}

The frozen kernel predicts the winners well but the decisions come earlier than it predicts. We measured, on the stored trajectories, the ratio $D$ of the observed mean change of the gap between the leader and the runner-up to the change that the frozen-kernel chain predicts from the same state, over blocks of eight steps before the decision. Table~\ref{tab:drift} shows that $D$ is close to one early in training and grows late, and that the order of this growth across five conditions is the order in which the decisions come earlier than the chain predicts. In shared sets the kernel grows with the size of the set. In sets of $64$, one step was about six times larger than the frozen prediction after $20$ to $40$ steps, while isolated runs and sets of two hardly changed. Most winners settle before this growth. Until $90$ percent of the winners had settled, the kernel had grown by a factor of at most $1.17$ in sets of up to $16$ questions and $1.41$ in sets of $64$. The calibration slopes of the chain, in contrast, were not explained by the drift and the noise of the gap measured in the same way, so we report them as a limitation.

\begin{table}[ht]
\caption{Growth of the kernel during training. $D$ is the ratio of the observed drift of the gap to the drift of the frozen-kernel chain from the same state, on the evaluated runs, early and late in training, and the last column is the ratio of the observed decision step to the one the chain predicts.}
\label{tab:drift}
\centering\small\setlength{\tabcolsep}{4pt}
\begin{tabular}{@{}lrllr@{}}
\toprule
condition & $D$ & early & late & decision\\
\midrule
Llama-1B & 1.94 & $1.01$ to $1.03$ & $1.80$, then $3.16$ & 0.49\\
AdamW, $\beta_1=0$ & 1.45 & about $1$ & $1.75$, then $2.82$ & 0.60\\
Qwen-1.5B & 1.28 & $0.90$ to $1.09$ & $1.30$, then $2.60$ & 0.65\\
SGD, std baseline & 1.22 & $0.87$ to $1.04$ & $1.70$, then $2.62$ & 0.67\\
Qwen-7B & 1.01 & $0.94$ to $1.06$ & $1.18$ to $1.24$ & 0.91\\
\bottomrule
\end{tabular}
\end{table}

\end{document}